\documentclass{article}

\usepackage[final]{neurips_2021}

\usepackage[utf8]{inputenc} 
\usepackage[T1]{fontenc}    
\usepackage[colorlinks=true,allcolors=blue]{hyperref}       
\usepackage{url}            
\usepackage{booktabs}       
\usepackage{amsfonts}       
\usepackage{nicefrac}       
\usepackage{microtype}      
\usepackage[dvipsnames]{xcolor}         

\usepackage{amsthm}
\newtheorem{definition}[]{Definition}
\newtheorem{assumption}[]{Assumption}
\newtheorem{theorem}{Theorem}[]
\newtheorem{lemma}[theorem]{Lemma}

\newtheorem{proposition}[theorem]{Proposition}

\newcommand{\mA}{\mathcal{A}}

\newcommand{\mN}{\mathcal{N}}

\newcommand{\pknn}{p_{k\mathrm{NN}}}
\newcommand{\pkde}{p_{\mathrm{KDE}}}
\newcommand{\pwsgmm}{p_{\mathrm{WS\text{-}GMM}}}

\newcommand{\platent}{P_{\mathrm{latent}}}
\newcommand{\dlatent}{d_{\mathrm{latent}}}

\newcommand{\mO}[1]{\mathcal{O}\left(#1\right)}
\newcommand{\OTheta}[1]{\Theta\left(#1\right)}
\newcommand{\KL}[2]{\mathrm{KL}\left(#1\|#2\right)}
\newcommand{\prob}[1]{\mathrm{Prob}\left(#1\right)}
\usepackage{amsmath}
\usepackage{amssymb}
\usepackage{graphicx}
\usepackage{multirow}
\usepackage{caption}
\usepackage{subfig}
\usepackage{tikz}
\usepackage[percent]{overpic}
\usepackage{pifont}
\usepackage{arydshln}

\title{Understanding Instance-based Interpretability of Variational Auto-Encoders}

\author{%
    Zhifeng Kong \\
    Computer Science and Engineering\\
    University of California San Diego\\
    La Jolla, CA 92093 \\
    \texttt{z4kong@eng.ucsd.edu} \\
    \And
    Kamalika Chaudhuri \\
    Computer Science and Engineering\\
    University of California San Diego\\
    La Jolla, CA 92093 \\
    \texttt{kamalika@cs.ucsd.edu} \\
}

\begin{document}

\maketitle

\begin{abstract}
Instance-based interpretation methods have been widely studied for supervised learning methods as they help explain how black box neural networks predict. However, instance-based interpretations remain ill-understood in the context of unsupervised learning. In this paper, we investigate influence functions \citep{koh2017understanding}, a popular instance-based interpretation method, for a class of deep generative models called variational auto-encoders (VAE). We formally frame the counter-factual question answered by influence functions in this setting, and through theoretical analysis, examine what they reveal about the impact of training samples on classical unsupervised learning methods. We then introduce VAE-TracIn, a computationally efficient and theoretically sound solution based on \citet{pruthi2020estimating}, for VAEs. Finally, we evaluate VAE-TracIn on several real world datasets with extensive quantitative and qualitative analysis. 
\end{abstract}

\section{Introduction}

Instance-based interpretation methods have been popular for supervised learning as they help explain why a model makes a certain prediction and hence have many applications \citep{barshan2020relatif,basu2020second,chen2020multi,ghorbani2019data,hara2019data,harutyunyan2021estimating,koh2017understanding,koh2019accuracy,pruthi2020estimating,yeh2018representer,yoon2020data}. For a classifier and a test sample $z$, an instance-based interpretation ranks all training samples $x$ according to an interpretability score between $x$ and $z$. Samples with high (low) scores are considered positively (negatively) important for the prediction of $z$. 

However, in the literature of unsupervised learning especially generative models, instance-based interpretations are much less understood. 
In this work, we investigate instance-based interpretation methods for unsupervised learning based on influence functions \citep{cook1980characterizations,koh2017understanding}. In particular, we theoretically analyze certain classical non-parametric and parametric methods. Then, we look at a canonical deep generative model, variational auto-encoders (VAE, \citep{higgins2016beta,kingma2013auto}), and explore some of the applications. 

The first challenge is framing the counter-factual question for unsupervised learning. For instance-based interpretability in supervised learning, the counter-factual question is "which training samples are most responsible for the prediction of a test sample?" -- which heavily relies on the label information. However, there is no label in unsupervised learning. In this work, we frame the counter-factual question for unsupervised learning as "which training samples are most responsible for increasing the likelihood (or reducing the loss when likelihood is not available) of a test sample?" We show that influence functions can answer this counter-factual question. Then, we examine influence functions for several classical unsupervised learning methods. We present theory and intuitions on how influence functions relate to likelihood and pairwise distances.

The second challenge is how to compute influence functions in VAE. The first difficulty here is that the VAE loss of a test sample involves an expectation over the encoder, so the actual influence function cannot be precisely computed. To deal with this problem, we use the empirical average of influence functions, and prove a concentration bound of the empirical average under mild conditions. Another difficulty is computation. The influence function involves inverting the Hessian of the loss with respect to all parameters, which involves massive computation for big neural networks with millions of parameters. To deal with this problem, we adapt a first-order estimate of the influence function called TracIn \citep{pruthi2020estimating} to VAE. We call our method VAE-TracIn. It is fast because ($i$) it does not involve the Hessian, and ($ii$) it can accelerate computation with only a few checkpoints. 

We begin with a sanity check that examines whether training samples have the highest influences over themselves, and show VAE-TracIn passes it. We then evaluate VAE-TracIn on several real world datasets. We find high (low) self influence training samples have large (small) losses. Intuitively, high self influence samples are hard to recognize or visually high-contrast, while low self influence ones share similar shapes or background. These findings lead to an application on unsupervised data cleaning, as high self influence samples are likely to be outside the data manifold. We then provide quantitative and visual analysis on influences over test data. We call high and low influence samples \textit{proponents} and \textit{opponents}, respectively. \footnote{There are different names in the literature, such as helpful/harmful samples \citep{koh2017understanding}, excitatory/inhibitory points \citep{yeh2018representer}, and proponents/opponents \citep{pruthi2020estimating}.} We find in certain cases both proponents and opponents are similar samples from the same class, while in other cases proponents have large norms. 

{We consider VAE-TracIn as a general-purpose tool that can potentially help understand many aspects in the unsupervised setting, including ($i$) detecting underlying memorization, bias or bugs \citep{feldman2020neural} in unsupervised learning, ($ii$) performing data deletion \citep{asokan2020teaching, izzo2021approximate} in generative models, and ($iii$) examining training data without label information.}

We make the following contributions in this paper.
\begin{itemize}
	\item We formally frame instance-based interpretations for unsupervised learning.
	\item We examine influence functions for several classical unsupervised learning methods.
	\item We present VAE-TracIn, an instance-based interpretation method for VAE. We provide both theoretical and empirical justification to VAE-TracIn. 
	\item We evaluate VAE-TracIn on several real world datasets. We provide extensive quantitative analysis and visualization, as well as an application on unsupervised data cleaning.
\end{itemize}

\subsection{Related Work}

There are two lines of research on instance-based interpretation methods for supervised learning.

The first line of research frames the following counter-factual question: which training samples are most responsible for the prediction of a particular test sample $z$? This is answered by designing an interpretability score that measures the importance of training samples over $z$ and selecting those with the highest scores. Many scores and their approximations have been proposed \citep{barshan2020relatif,basu2020second,chen2020multi,hara2019data,koh2017understanding,koh2019accuracy,pruthi2020estimating,yeh2018representer}. Specifically, \citet{koh2017understanding} introduce the influence function (IF) based on the terminology in robust statistics \citep{cook1980characterizations}. The intuition is removing an important training sample of $z$ should result in a huge increase of its test loss. Because the IF is hard to compute, \citet{pruthi2020estimating} propose TracIn, a fast first-order approximation to IF. 

Our paper extends the counter-factual question to unsupervised learning where there is no label. We ask: which training samples are most responsible for increasing the likelihood (or reducing the loss) of a test sample? In this paper, we propose VAE-TracIn, an instance-based interpretation method for VAE \citep{higgins2016beta,kingma2013auto} based on TracIn and IF. 

The second line of research considers a different counter-factual question: which training samples are most responsible for the overall performance of the model (e.g. accuracy)? This is answered by designing an interpretability score for each training sample. Again many scores have been proposed \citep{ghorbani2019data,harutyunyan2021estimating,yoon2020data}. \citet{terashita2021influence} extend this framework to a specific unsupervised model called generative adversarial networks \citep{goodfellow2014generative}. They measure influences of samples on several evaluation metrics, and discard samples that harm these metrics. Our paper is orthogonal to these works.

The instance-based interpretation methods lead to many applications in various areas including adversarial learning, data cleaning, prototype selection, data summarization, and outlier detection \citep{barshan2020relatif,basu2020second,chen2020multi,feldman2020neural,ghorbani2019data,hara2019data,harutyunyan2021estimating,khanna2019interpreting,koh2017understanding,pruthi2020estimating,suzuki2021data,ting2018optimal,ye2021out,yeh2018representer,yoon2020data}. In this paper, we apply the proposed VAE-TracIn to an unsupervised data cleaning task.

Prior works on interpreting generative models analyze their latent space via measuring disentanglement, explaining and visualizing representations, or analysis in an interactive interface \citep{alvarez2018towards,bengio2013representation,chen2016infogan,desjardins2012disentangling,kim2018disentangling,olah2017feature,olah2018the, ross2021evaluating}. These latent space analysis are complementary to the instance-based interpretation methods in this paper.

\section{Instance-based Interpretations}
Let $X=\{x_i\}_{i=1}^N\in\mathbb{R}^d$ be the training set. Let $\mA$ be an algorithm that takes $X$ as input and outputs a model that describes the distribution of $X$. $\mA$ can be a density estimator or a generative model. 
Let $L(X; \mA)=\frac1N\sum_{i=1}^N\ell(x_i; \mA(X))$ be a loss function. Then, the influence function of a training data $x_i$ over a test data $z\in\mathbb{R}^d$ is the loss of $z$ computed from the model trained without $x_i$ minus that computed from the model trained with $x_i$. If the difference is large, then $x_i$ should be very influential for $z$. Formally, the influence function is defined below. 

\begin{definition}[Influence functions \citep{koh2017understanding}]\label{def: influence}
Let $X_{-i}=X\setminus\{x_i\}$. Then, the influence of $x_i$ over $z$ is defined as
$
    \mathrm{IF}_{X,\mA}(x_i,z)=\ell(z;\mA(X_{-i}))-\ell(z;\mA(X)).
$
If $\mathrm{IF}_{X,\mA}(x_i,z)>0$, we say $x_i$ is a proponent of $z$; otherwise, we say $x_i$ is an opponent of $z$. 
\end{definition}

For big models $\mA$ such as deep neural networks, doing retraining and obtaining $\mA(X_{-i})$ can be expensive. The following TracIn score is a fast approximation to IF. 

\begin{definition}[TracIn scores \citep{pruthi2020estimating}]\label{def: tracin}
Suppose $\mA(X)$ is obtained by minimizing $L(X;\mA)$ via stochastic gradient descent. Let $\{\theta_{[c]}\}_{c=1}^C$ be $C$ checkpoints during the training procedure. Then, the estimated influence of $x_i$ over $z$ is defined as
$
    \mathrm{TracIn}_{X,\mA}(x_i,z)=\sum_{c=1}^C \nabla\ell(x_i;\theta_{[c]})^{\top} \nabla\ell(z;\theta_{[c]}).
$
\end{definition}

We are also interested in the influence of a training sample over itself. Formally, we define this quantity as the self influence of $x$, or $\mathrm{Self\text{-}IF}_{X,\mA}(x) = \mathrm{IF}_{X,\mA}(x,x)$. In supervised learning, self influences provide rich information about memorization properties of training samples. Intuitively, high self influence samples are atypical, ambiguous or mislabeled, while low self influence samples are typical \citep{feldman2020neural}. 

\section{Influence Functions for Classical Unsupervised Learning}\label{sec: classic}

In this section, we analyze influence functions for unsupervised learning. The goal is to provide intuition on what influence functions should tell us in the unsupervised setting. Specifically, we look at three classical methods: the non-parametric $k$-nearest-neighbor ($k$-NN) density estimator, the non-parametric kernel density estimator (KDE), and the parametric Gaussian mixture models (GMM). We let the loss function $\ell$ to be the negative log-likelihood: $\ell(z)=-\log p(z)$. 

\textbf{The $k$-Nearest-Neighbor ($k$-NN) density estimator.} The $k$-NN density estimator is defined as $\pknn(x;X)=k/(N V_d R_k(x;X)^d)$, where $R_k(x;X)$ is the distance between $x$ and its $k$-th nearest neighbor in $X$ and $V_d$ is the volume of the unit ball in $\mathbb{R}^d$. Then, we have the following influence function for the $k$-NN density estimator:

\begin{equation}\label{eq: knn influence}
    \mathrm{IF}_{X,k\mathrm{NN}}(x_i,z) = \log\frac{N-1}{N} + \left\{
    \begin{array}{cl}
        d\log\frac{R_{k+1}(z;X)}{R_k(z;X)} & \|x_i-z\|\leq R_k(z;X) \\
        0 & \text{otherwise}
    \end{array}
    \right. .
\end{equation}

See Appendix \ref{appendix: knn} for proof. 
Note, when $z$ is fixed, there are only two possible values for training data influences: $\log\frac{N-1}{N}$ and $\log\frac{N-1}{N} + d\log\frac{R_{k+1}(z;X)}{R_k(z;X)}$.
As for $\mathrm{Self\text{-}IF}_{X,k\mathrm{NN}}(x_i)$, samples with the largest self influences are those with the largest $\frac{R_{k+1}(x_i;X)}{R_k(x_i;X)}$. Intuitively, these samples belong to a cluster of size exactly $k$, and the cluster is far away from other samples. 

\textbf{Kernel Density Estimators (KDE).} The KDE is defined as $\pkde(x;X)=\frac1N\sum_{i=1}^N K_{\sigma}(x-x_i)$, where $K_{\sigma}$ is the Gaussian $\mN(0,\sigma^2I)$. Then, we have the following influence function for KDE:

\begin{equation}\label{eq: kde influence}
    \mathrm{IF}_{X,\mathrm{KDE}}(x_i,z) = \log\frac{N-1}{N} + \log\left(1+\frac{\frac1N K_{\sigma}(z-x_i)}{p_{\mathrm{KDE}}(z;X)-\frac1N K_{\sigma}(z-x_i)}\right) .
\end{equation}

See Appendix \ref{appendix: kde} for proof. For a fixed $z$, an $x_i$ with larger $\|z-x_i\|$ has a higher influence over $z$. Therefore, the strongest proponents of $z$ are those closest to $z$ in the $\ell_2$ distance, and the strongest opponents are the farthest. 
As for $\mathrm{Self\text{-}IF}_{X,\mathrm{KDE}}(x_i)$, samples with the largest self influences are those with the least likelihood $\pkde(x_i;X)$. Intuitively, these samples locate at very sparse regions and have few nearby samples. On the other hand, samples with the largest likelihood $\pkde(x_i;X)$, or those in the high density area, have the least self influences.

\textbf{Gaussian Mixture Models (GMM).} As there is no closed-form expression for general GMM, we make the following well-separation assumption to simplify the problem. 

\begin{assumption}\label{assump: well separation}
    $X=\bigcup_{k=0}^{K-1}X_k$ where each $X_k$ is a cluster. We assume these clusters are well-separated: $\min\{\|x-x'\|: x\in X_k,x'\in X_{k'}\} \gg \max\{\|x-x'\|: x,x'\in X_k\}$.
\end{assumption}

Let $|X_k|=N_k$ and $N=\sum_{k=0}^{K-1}N_k$. For $x\in\mathbb{R}^d$, let $k=\arg\min_i d(x,X_i)$. Then, we define the well-separated spherical GMM (WS-GMM) of $K$ mixtures as $\pwsgmm(x) = \frac{N_k}{N}\mN(x;\mu_k, \sigma_k^2I)$, where the parameters are given by the maximum likelihood estimates
\begin{equation}\label{eq: wsgmm parameters}
    \mu_k=\frac{1}{N_k}\sum_{x\in X_k}x,\  \sigma_k^2=\frac{1}{N_kd}\sum_{x\in X_k}\|x-\mu_k\|^2=\frac{1}{N_kd}\sum_{x\in X_k}x^{\top}x - \frac1d\mu_k^{\top}\mu_k.
\end{equation}

For conciseness, we let test sample $z$ from cluster zero: $z\in\mathrm{conv}(X_0)$. Then, we have the following influence function for WS-GMM. If $x_i\notin X_0$, $\mathrm{IF}_{X,\mathrm{WS\text{-}GMM}}(x_i,z) = -\frac1N + \mO{N^{-2}}$. Otherwise,
\begin{equation}\label{eq: wsgmm influence}
    \mathrm{IF}_{X,\mathrm{WS\text{-}GMM}}(x_i,z) = \frac{d+2}{2N_0} + \frac{1}{2N_0\sigma_0^2}\left(\frac{\|z-\mu_0\|^2}{\sigma_0^2}-\|z-x_i\|^2\right) -\frac{1}{N} + \mO{N_0^{-2}}.
\end{equation}

See Appendix \ref{appendix: ws-gmm} for proof. 
A surprising finding is that some $x_i\in X_0$ may have very negative influences over $z$ (i.e. strong opponents of $z$ are from the same class). This happens with high probability if $\|z-x_i\|^2\gtrapprox (1+\sigma_0^2)d+2\sigma_0^2$ for large dimension $d$. 
Next, we compute the self influence of an $x_i\in X_k$. According to \eqref{eq: wsgmm influence},
\begin{equation}\label{eq: wsgmm self influence}
\mathrm{Self\text{-}IF}_{X,\mathrm{WS\text{-}GMM}}(x_i) = \frac{d+2}{2N_k} + \frac{\|x_i-\mu_k\|^2}{2N_k\sigma_k^4} -\frac{1}{N} + \mO{N_k^{-2}}.
\end{equation}

Within each cluster $X_k$, samples far away to the cluster center $\mu_k$ have large self influences and vice versa. Across the entire dataset, samples in cluster $X_k$ whose $N_k$ or $\sigma_k$ is small tend to have large self influences, which is very different from $k$-NN or KDE.

\subsection{Summary}\label{sec: classic visualization}

We summarize the intuitions of influence functions in classical unsupervised learning in Table \ref{tab: intuition}. Among these methods, the strong proponents are all nearest samples, but self influences and strong opponents are quite different.
We then visualize an example of six clusters of 2D points in Figure \ref{fig: classic data appendix} in Appendix \ref{appendix: classic figures}. In Figure \ref{fig: classic self inf appendix}, We plot the self influences of these data points under different density estimators. For a test data point $z$, we plot influences of all data points over $z$ in Figure \ref{fig: classic test inf appendix}.

\section{Instance-based Interpretations for Variational Auto-encoders}\label{sec: VAE}

In this section, we show how to compute influence functions for a class of deep generative models called variational auto-encoders (VAE). Specifically, we look at $\beta$-VAE \citep{higgins2016beta} defined below, which generalizes the original VAE by \citet{kingma2013auto}. 

\begin{definition}[$\beta$-VAE \citep{higgins2016beta}]\label{def: beta vae}
Let $\dlatent$ be the latent dimension. Let $P_{\phi}:\mathbb{R}^{\dlatent}\rightarrow\mathbb{R}^+$ be the decoder and $Q_{\psi}:\mathbb{R}^{d}\rightarrow\mathbb{R}^+$ be the encoder, where $\phi$ and $\psi$ are the parameters of the networks. Let $\theta=[\phi,\psi]$. Let the latent distribution $\platent$ be $\mN(0,I)$. For $\beta>0$, the $\beta$-VAE model minimizes the following loss:
\begin{equation}\label{eq: beta vae loss}
L_{\beta}(X;\theta) = \mathbb{E}_{x\sim X} \ell_{\beta}(x;\theta) = \beta\cdot\mathbb{E}_{x\sim X}\KL{Q_{\psi}(\cdot|x)}{\platent} - \mathbb{E}_{x\sim X}\mathbb{E}_{\xi\sim Q_{\psi}(\cdot|x)} \log P_{\phi}(x|\xi).
\end{equation}
\end{definition}

In practice, the encoder $Q=Q_{\psi}$ outputs two vectors, $\mu_Q$ and $\sigma_Q$, so that $Q(\cdot|x)=\mN(\mu_Q(x),\mathrm{diag}(\sigma_Q(x))^2I)$. The decoder $P=P_{\phi}$ outputs a vector $\mu_P$ so that $\log P(x|\xi)$ is a constant times $\|\mu_P(\xi)-x\|^2$ plus a constant.

Let $\mA$ be the $\beta$-VAE that returns $\mA(X)=\arg\min_{\theta}L_{\beta}(X;\theta)$. Let $\theta^*=\mA(X)$ and $\theta_{-i}^*=\mA(X_{-i})$. Then, the influence function of $x_i$ over a test point $z$ is $\ell_{\beta}(z;\theta_{-i}^*) - \ell_{\beta}(z;\theta^*)$, which equals to
\begin{equation}\label{eq: beta vae inf def}
\begin{array}{rl}
    \mathrm{IF}_{X,\mathrm{VAE}}(x_i,z) & = \beta\left(\KL{Q_{\psi_{-i}^*}(\cdot|z)}{\platent}-\KL{Q_{\psi^*}(\cdot|z)}{\platent}\right) \\
    & ~~ - \left(\mathbb{E}_{\xi\sim Q_{\psi_{-i}^*}(\cdot|z)} \log P_{\phi_{-i}^*}(z|\xi) - \mathbb{E}_{\xi\sim Q_{\psi^*}(\cdot|z)} \log P_{\phi^*}(z|\xi)\right).
\end{array}
\end{equation}

\textbf{Challenge.}
The first challenge is that IF in \eqref{eq: beta vae inf def} involves an expectation over the encoder, so it cannot be precisely computed. To solve the problem, we compute the empirical average of the influence function over $m$ samples. In \textbf{Theorem} \ref{thm: error bound beta vae informal}, we theoretically prove that the empirical influence function is close to the actual influence function with high probability when $m$ is properly selected. The second challenge is that IF is hard to compute. To solve this problem, in Section \ref{sec: VAE-TracIn}, we propose VAE-TracIn, a computationally efficient solution to VAE.

\textbf{A probabilistic bound on influence estimates.}
Let $\hat{\mathrm{IF}}_{X,\mathrm{VAE}}^{(m)}$ be the empirical average of the influence function over $m$ i.i.d. samples. We have the following result.

\begin{theorem}[Error bounds on influence estimates (informal, see formal statement in \textbf{Theorem} \ref{thm: error bound beta vae})]\label{thm: error bound beta vae informal} 
Under mild conditions, for any small $\epsilon>0$ and $\delta>0$, there exists an $m=\OTheta{\frac{1}{\epsilon^2\delta}}$ such that
\begin{equation}\label{eq: inf error bound}
    \prob{\left|\mathrm{IF}_{X,\mathrm{VAE}}(x_i,z)-\hat{\mathrm{IF}}_{X,\mathrm{VAE}}^{(m)}(x_i,z)\right|\geq \epsilon} \leq \delta.
\end{equation}
\end{theorem}

Formal statements and proofs are in Appendix \ref{sec: prob bound VAE appendix}.

\subsection{VAE-TracIn} \label{sec: VAE-TracIn}
In this section, we introduce VAE-TracIn, a computationally efficient interpretation method for VAE. VAE-TracIn is built based on TracIn (\textbf{Definition} \ref{def: tracin}). According to \eqref{eq: beta vae loss}, the gradient of the loss $\ell_{\beta}(x;\theta)$ can be written as $\nabla_{\theta}\ell_{\beta}(x;\theta)=[\nabla_{\phi}\ell_{\beta}(x;\theta)^{\top}, \nabla_{\psi}\ell_{\beta}(x;\theta)^{\top}]^{\top}$, where
\begin{equation}\label{eq: beta vae loss gradient}
\begin{array}{rl}
	\nabla_{\phi}\ell_{\beta}(x;\theta) 
	& \displaystyle = \mathbb{E}_{\xi\sim Q_{\psi}(\cdot|x)} \left(-\nabla_{\phi}\log P_{\phi}(x|\xi)\right) =: \mathbb{E}_{\xi\sim Q_{\psi}(\cdot|x)} U(x,\xi,\phi,\psi),\text{ and }\\
	\nabla_{\psi}\ell_{\beta}(x;\theta) 
	& \displaystyle = \mathbb{E}_{\xi\sim Q_{\psi}(\cdot|x)} \nabla_{\psi}\log Q_{\psi}(\xi|x) \left(\beta\log\frac{Q_{\psi}(\xi|x)}{P_{\mathrm{latent}}(\xi)}-\log P_{\phi}(x|\xi)\right) \\
	& \displaystyle =: \mathbb{E}_{\xi\sim Q_{\psi}(\cdot|x)} V(x,\xi,\phi,\psi).
\end{array}
\end{equation}

The derivations are based on the Stochastic Gradient Variational Bayes estimator \citep{kingma2013auto}, which offers low variance \citep{rezende2014stochastic}. See Appendix \ref{appendix: gradient loss} for full details of the derivation. We estimate the expectation $\mathbb{E}_{\xi}$ by averaging over $m$ i.i.d. samples. Then, for a training data $x$ and test data $z$, the VAE-TracIn score of $x$ over $z$ is computed as 
\begin{equation}\label{eq: VAE-TracIn}
\begin{array}{rl}
	\mathrm{VAE\text{-}TracIn}(x,z) & \displaystyle = \sum_{c=1}^C \left(\frac1m\sum_{j=1}^m U(x,\xi_{j,[c]},\phi_{[c]},\psi_{[c]})\right)^{\top}\left(\frac1m\sum_{j=1}^m U(z,\xi'_{j,[c]},\phi_{[c]},\psi_{[c]})\right) \\
	& \displaystyle + \sum_{c=1}^C \left(\frac1m\sum_{j=1}^m V(x,\xi_{j,[c]},\phi_{[c]},\psi_{[c]})\right)^{\top}\left(\frac1m\sum_{j=1}^m V(z,\xi'_{j,[c]},\phi_{[c]},\psi_{[c]})\right),
\end{array}
\end{equation}
where the notations $U,V$ are from \eqref{eq: beta vae loss gradient}, $\theta_{[c]}=[\phi_{[c]},\psi_{[c]}]$ is the $c$-th checkpoint, $\{\xi_{j,[c]}\}_{j=1}^m$ are i.i.d. samples from $Q_{\psi_{[c]}}(\cdot|x)$, and $\{\xi'_{j,[c]}\}_{j=1}^m$ are i.i.d. samples from $Q_{\psi_{[c]}}(\cdot|z)$.

{
\paragraph{Connections between VAE-TracIn and influence functions \citep{koh2017understanding}.} \citet{koh2017understanding} use the second-order (Hessian-based) approximation to the change of loss under the assumption that the loss function is convex. The TracIn algorithm \citep{pruthi2020estimating} uses the first-order (gradient-based) approximation to the change of loss during the training process under the assumption that (stochastic) gradient descent is the optimizer. We expect these methods to give similar results in the ideal situation. However, we implemented the method by \citet{koh2017understanding} and found it to be inaccurate for VAE. A possible reason is that the Hessian vector product used to approximate the second order term is unstable.
}

{
\paragraph{Complexity of VAE-TracIn.} The run-time complexity of VAE-TracIn is linear in the number of samples ($N$), checkpoints ($C$), and network parameters ($|\theta|$).
}

\section{Experiments}\label{sec: experiments}

In this section, we aim to answer the following questions. 
\begin{itemize}
    \item Does VAE-TracIn pass a sanity check for instance-based interpretations?
    \item Which training samples have the highest and lowest self influences, respectively?
    \item Which training samples have the highest influences over (i.e. are strong proponents of) a test sample? Which have the lowest influences over it (i.e. are its strong opponents)?
\end{itemize}

These questions are examined in experiments on the MNIST \citep{lecun2010mnist} and CIFAR-10 \citep{krizhevsky2009learning} datasets.

\subsection{Sanity Check}\label{sec: sanity check}
\textbf{Question.} Does VAE-TracIn find the most influential training samples? In a perfect instance-based interpretation for a good model, training samples should have large influences over themselves. As a sanity check, we examine if training samples are the strongest proponents over themselves. This is analogous to the identical subclass test by \citet{hanawa2020evaluation}.

\textbf{Methodology.} We train separate VAE models on MNIST, CIFAR, and each CIFAR subclass (the set of five thousand CIFAR samples in each class). For each model, we examine the frequency that a training sample is the most influential one among all samples over itself. Due to computational limits we examine the first 128 samples. The results for MNIST, CIFAR, and the averaged result for CIFAR subclasses are reported in Table \ref{tab: validity}. Detailed results for CIFAR subclasses are in Appendix \ref{appendix: validity}.

\begin{table}[!t]
    \caption{Sanity check on the frequency that a training sample is the most influential one over itself. Results on MNIST, CIFAR, and the averaged result on CIFAR subclasses are reported.}
    \vspace{0.5em}
    \label{tab: validity}
    \centering
    \begin{tabular}{cc|cc|c}
    \hline 
    	\multicolumn{2}{c|}{MNIST} &  \multicolumn{2}{|c|}{CIFAR} &  Averaged CIFAR subclass \\
        $\dlatent=64$ & $\dlatent=128$ & $\dlatent=64$ & $\dlatent=128$ & $\dlatent=64$ \\ \hline
        0.992 & 1.000 & 0.609 & 0.602 & 0.998 \\ 
    \hline
    \end{tabular}
\end{table}

\textbf{Results.} The results indicate that VAE-TracIn can find the most influential training samples in MNIST and CIFAR subclasses. This is achieved even under the challenge that many training samples are very similar to each other. The results for CIFAR is possibly due to underfitting as it is challenging to train a good VAE on this dataset. Note, the same VAE architecture is trained on CIFAR subclasses. 

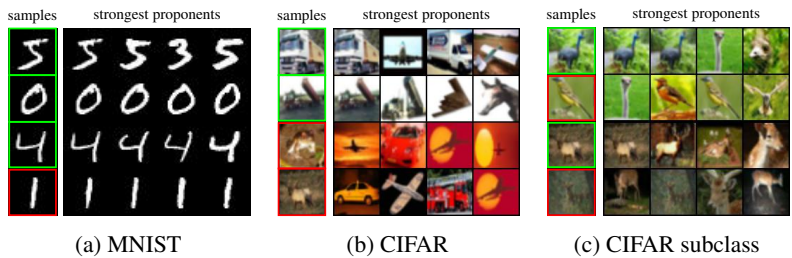
\begin{figure}[!t]
    \centering
    \subfloat[][MNIST]{ 
    \resizebox{0.25\textwidth}{!}{%
	\begin{tikzpicture}
	    \node (image) at (0,0) {
            \includegraphics[height=3cm]{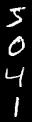}};
            \node (image) at (2,0) {
            \includegraphics[height=3cm]{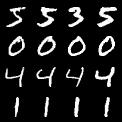}};
            \draw[red, thick] (-0.375,-1.5) rectangle (0.375,-0.77);
            \draw[green, thick] (-0.375,-0.73) rectangle (0.375,0);
            \draw[green, thick] (-0.375,0) rectangle (0.375,0.75);
            \draw[green, thick] (-0.375,0.75) rectangle (0.375,1.5);
            \node[] at (0,1.7)  {\scriptsize{samples}};
            \node[] at (2,1.7)  {\scriptsize{strongest proponents}};
        \end{tikzpicture}
	}%
    }
    \subfloat[][CIFAR]{ 
    \resizebox{0.25\textwidth}{!}{%
        \begin{tikzpicture}
	    \node (image) at (0,0) {
            \includegraphics[height=3cm]{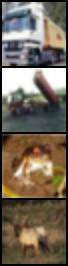}};
            \node (image) at (2,0) {
            \includegraphics[height=3cm]{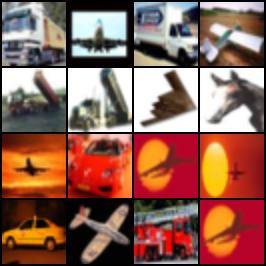}};
            \draw[red, thick] (-0.375,-1.5) rectangle (0.375,-0.75);
            \draw[red, thick] (-0.375,-0.75) rectangle (0.375,-0.02);
            \draw[green, thick] (-0.375,0.02) rectangle (0.375,0.75);
            \draw[green, thick] (-0.375,0.75) rectangle (0.375,1.5);
            \node[] at (0,1.7)  {\scriptsize{samples}};
            \node[] at (2,1.7)  {\scriptsize{strongest proponents}};
        \end{tikzpicture}
	}%
    }
    \subfloat[][CIFAR subclass]{ 
    \resizebox{0.25\textwidth}{!}{%
        \begin{tikzpicture}
	    \node (image) at (0,0) {
            \includegraphics[height=3cm]{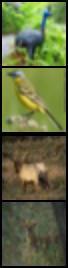}};
            \node (image) at (2,0) {
            \includegraphics[height=3cm]{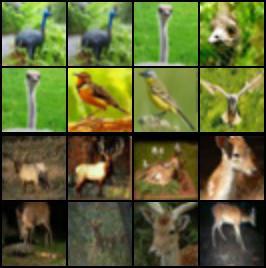}};
            \draw[red, thick] (-0.375,-1.5) rectangle (0.375,-0.77);
            \draw[green, thick] (-0.375,-0.73) rectangle (0.375,-0.02);
            \draw[red, thick] (-0.375,0.02) rectangle (0.375,0.73);
            \draw[green, thick] (-0.375,0.77) rectangle (0.375,1.5);
            \node[] at (0,1.7)  {\scriptsize{samples}};
            \node[] at (2,1.7)  {\scriptsize{strongest proponents}};
        \end{tikzpicture}
	}%
    }
    \vspace{-0.25em}
    \caption{Certain training samples and their strongest proponents in the training set (sorted from left to right). A sample $x_i$ is marked in green box if it is more influential than other samples over itself (i.e. it is the strongest proponent of itself) and otherwise in red box. }
    \label{fig: validity}
\end{figure}

\textbf{Visualization.} We visualize some correctly and incorrectly identified strongest proponents in Figure \ref{fig: validity}. On MNIST or CIFAR subclasses, even if a training sample is not exactly the strongest proponent of itself, it still ranks very high in the order of influences.

\subsection{Self Influences}\label{sec: self inf}
\textbf{Question.} Which training samples have the highest and lowest self influences, respectively? Self influences provide rich information about properties of training samples such as memorization. In supervised learning, high self influence samples can be atypical, ambiguous or mislabeled, while low self influence samples are typical \citep{feldman2020neural}. We examine what self influences reveal in VAE. 

\begin{figure}[!t]
    \centering
     \subfloat[][MNIST]{ 
    	\includegraphics[trim=10 12 5 0, clip, width=0.32\textwidth]{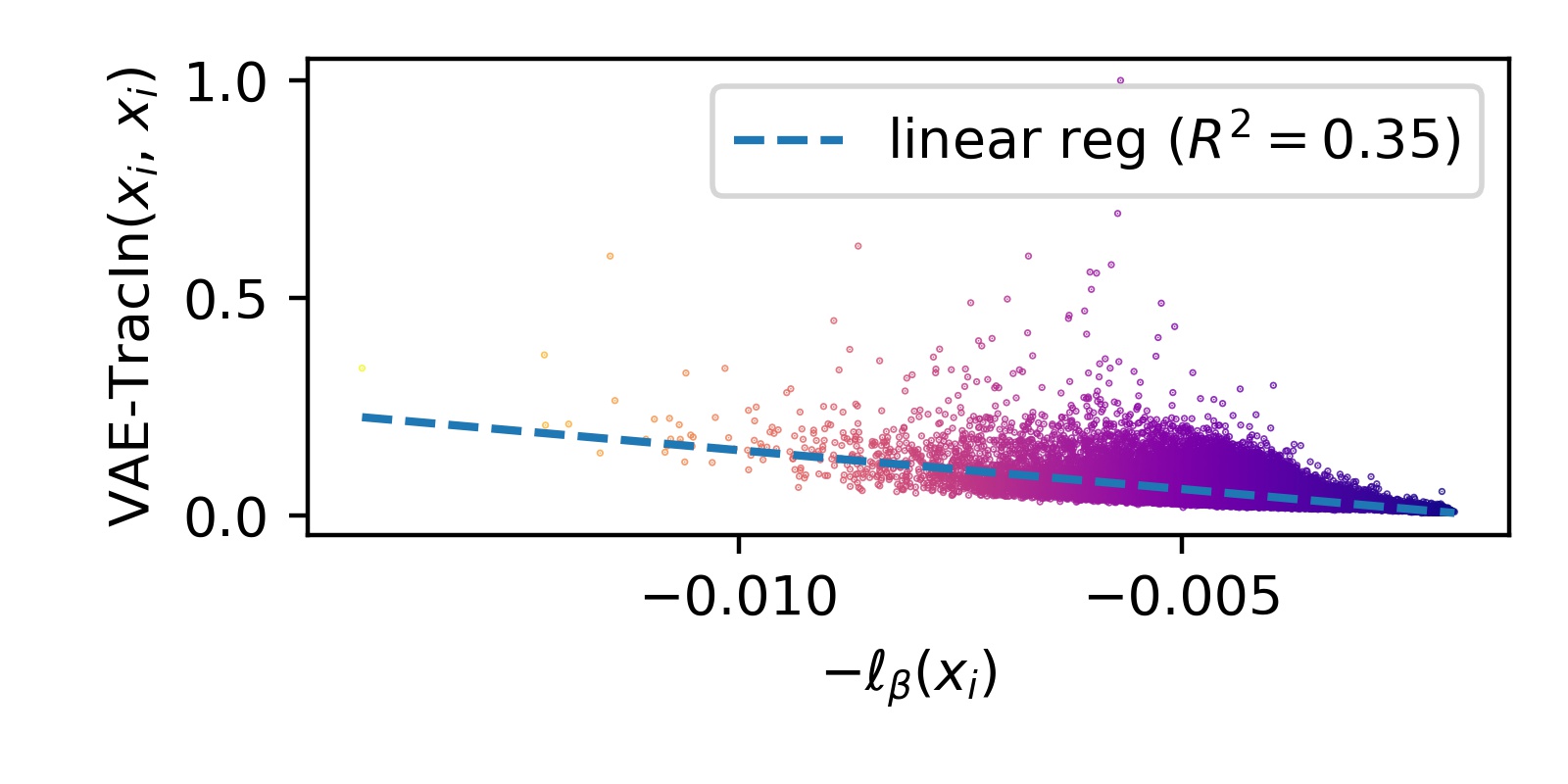}
    	\label{fig: self inf vs loss mnist}
	}
     \subfloat[][CIFAR]{ 
    	\includegraphics[trim=10 12 5 0, clip, width=0.32\textwidth]{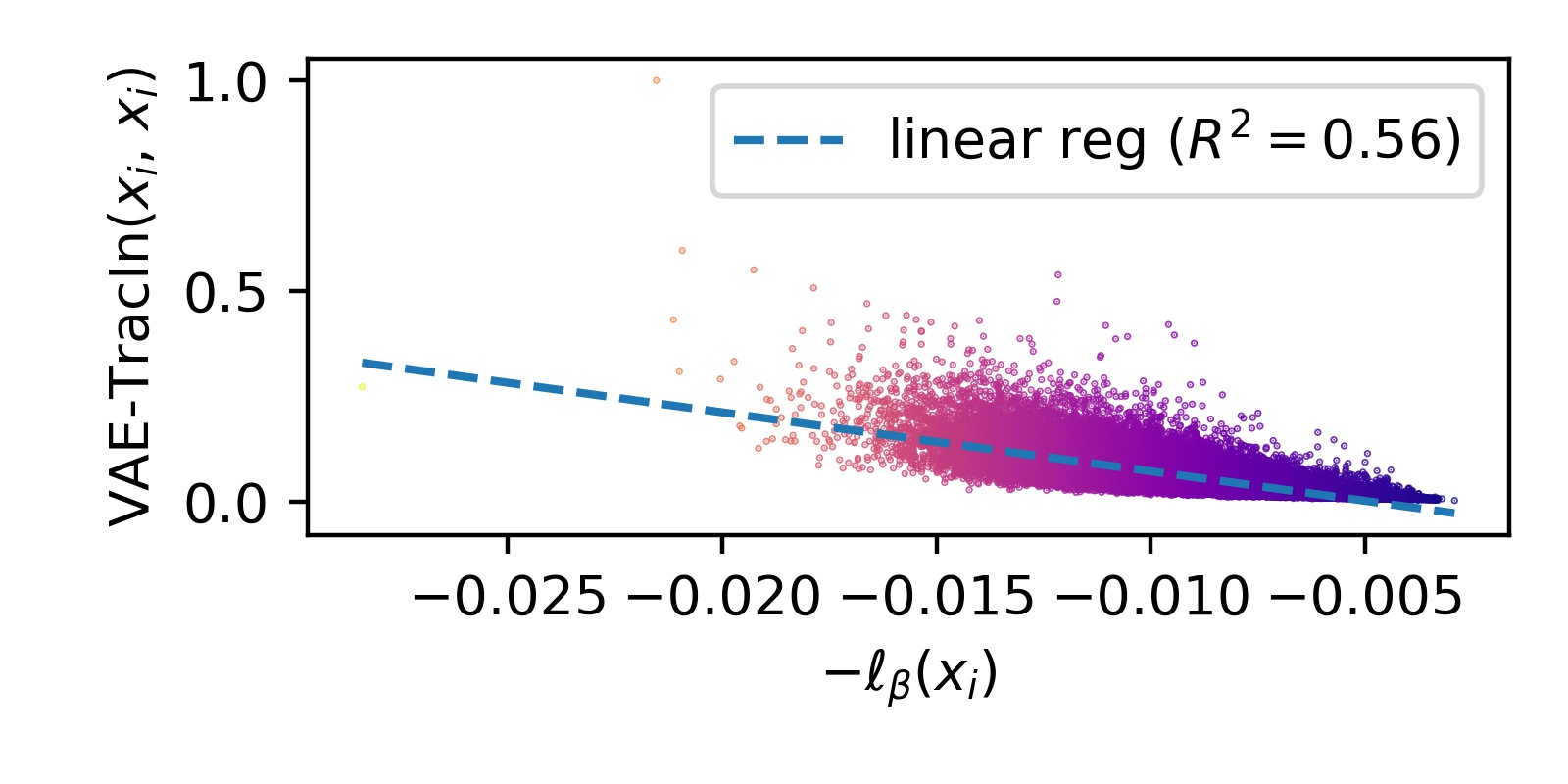}
    	\label{fig: self inf vs loss cifar}
	}
     \subfloat[][CIFAR-Airplane]{ 
    	\includegraphics[trim=10 12 5 0, clip, width=0.32\textwidth]{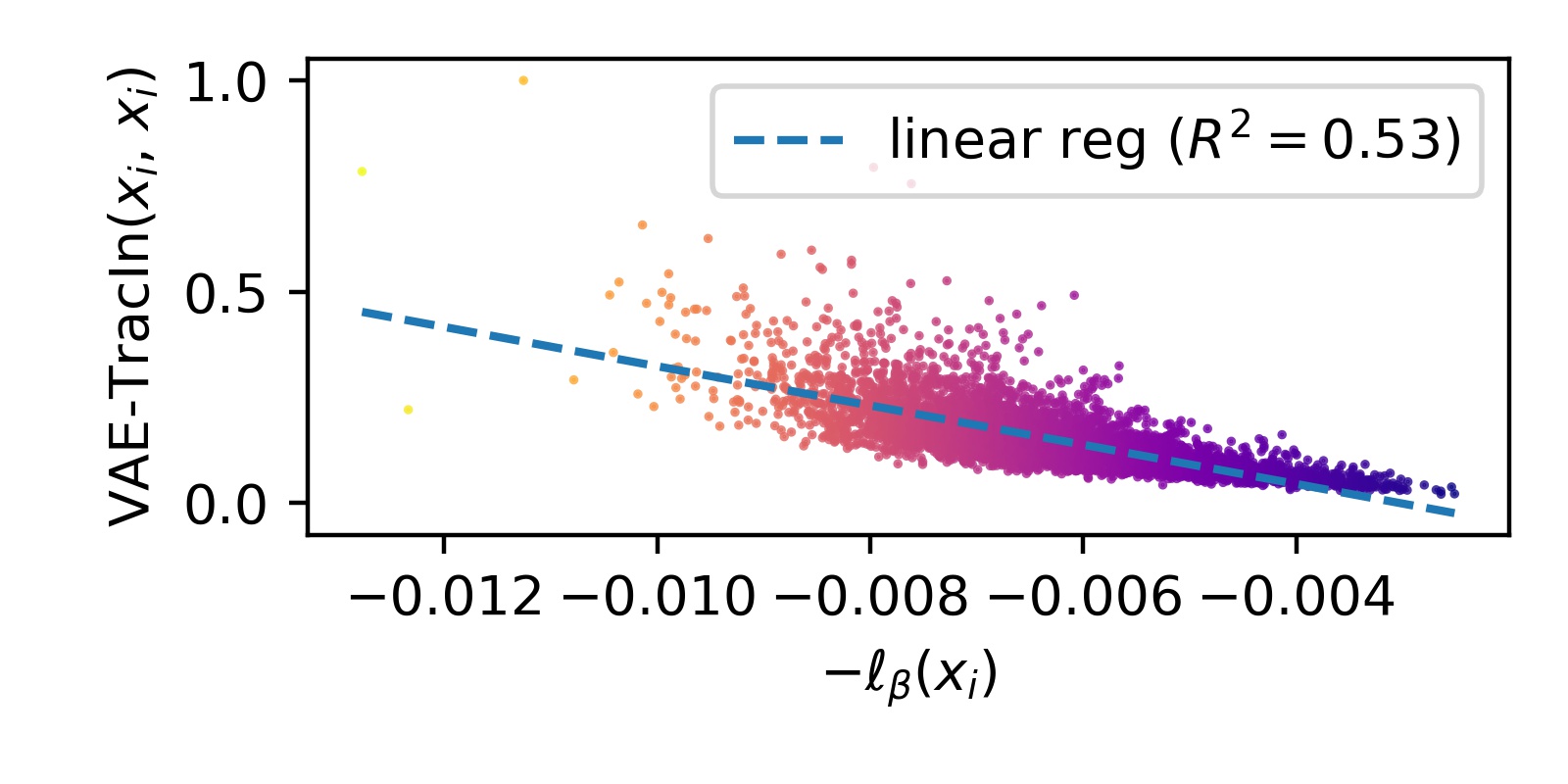}
    	\label{fig: self inf vs loss cifar-0}
	}
    \vspace{-0.25em}
    \caption{Scatter plots of self influences versus negative losses of all training samples in several datasets. The linear regressors show that high self influence samples have large losses.}
    \label{fig: self inf vs loss}
\end{figure}

\textbf{Methodology.} We train separate VAE models on MNIST, CIFAR, and each CIFAR subclass. We then compute the self influences and losses of each training sample. We show the scatter plots of self influences versus negative losses in Figure \ref{fig: self inf vs loss}. \footnote{We use the negative loss because it relates to the log-likelihood of $x_i$: when $\beta=1$, $-\ell_{\beta}(x)\leq\log p(x)$.} We fit linear regression models to these points and report the $R^2$ scores of the regressors. More comparisons including the marginal distributions and the joint distributions can be found in Appendix \ref{appendix: self-inf mnist} and Appendix \ref{appendix: self-inf cifar}.

\textbf{Results.} We find the self influence of a training sample $x_i$ tends to be large when its loss $\ell_{\beta}(x_i)$ is large. This finding in VAE is consistent with KDE and GMM (see Figure \ref{fig: classic self inf appendix}). In supervised learning, \citet{pruthi2020estimating} find high self influence samples come from densely populated areas while low self influence samples come from sparsely populated areas. Our findings indicate significant difference between supervised and unsupervised learning in terms of self influences under certain scenarios.

\textbf{Visualization.} We visualize high and low self influence samples in Figure \ref{fig: self inf visualization} (more visualizations in Appendix \ref{appendix: self-inf cifar}). High self influence samples are either hard to recognize or visually high-contrast, while low self influence samples share similar shapes or background. These visualizations are consistent with the memorization analysis by \citet{feldman2020neural} in the supervised setting.  We also notice that there is a concurrent work connecting self influences on log-likelihood and memorization properties in VAE through cross validation and retraining \citep{van2021memorization}. Our quantitative and qualitative results are consistent with their results.

\begin{figure}[!t]
    \centering
     \subfloat[][MNIST (highest self-inf)]{ 
    	\includegraphics[width=0.32\textwidth]{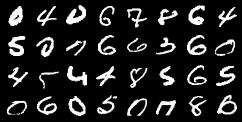}
	}
     \subfloat[][CIFAR (highest self-inf)]{ 
    	\includegraphics[width=0.32\textwidth]{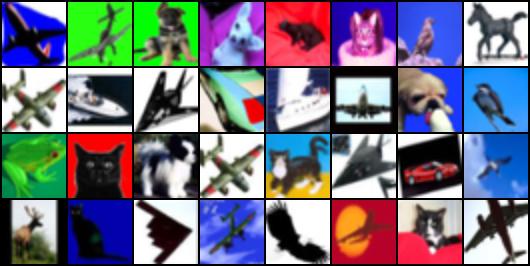}
	}
     \subfloat[][CIFAR-Airplane (highest self-inf)]{ 
    	\includegraphics[width=0.32\textwidth]{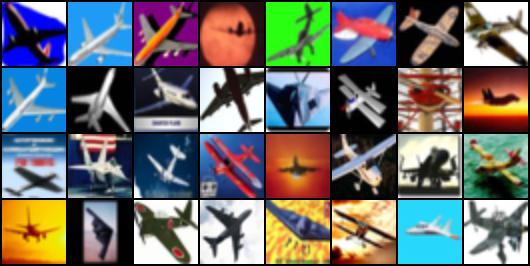}
	}\\ \vspace{-0.5em}
   \subfloat[][MNIST (lowest self-inf)]{ 
    	\includegraphics[width=0.32\textwidth]{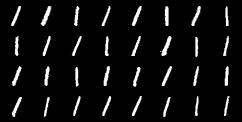}
	}
     \subfloat[][CIFAR (lowest self-inf)]{ 
    	\includegraphics[width=0.32\textwidth]{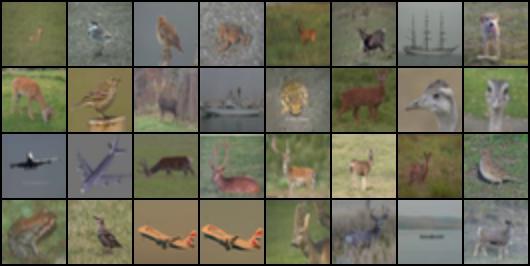}
	}
     \subfloat[][CIFAR-Airplane (lowest self-inf)]{ 
    	\includegraphics[width=0.32\textwidth]{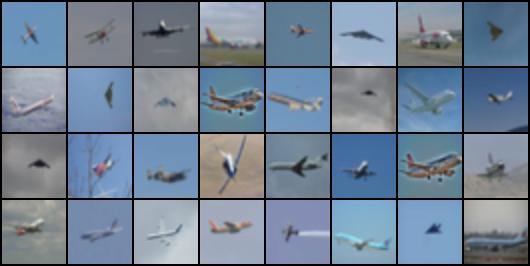}
	}
    \vspace{-0.25em}
    \caption{High and low self influence samples from several datasets. High self influence samples are hard to recognize or high-contrast. Low self influence samples share similar shapes or background.}
    \label{fig: self inf visualization}
\end{figure}

\textbf{Application on unsupervised data cleaning.} 
A potential application on unsupervised data cleaning is to use self influences to detect unlikely samples and let a human expert decide whether to discard them before training. The unlikely samples may include noisy samples, contaminated samples, or incorrectly collected samples due to bugs in the data collection process. For example, they could be unrecognizable handwritten digits in MNIST or objects in CIFAR. Similar approaches in supervised learning use self influences to detect mislabeled data \citep{koh2017understanding,pruthi2020estimating,yeh2018representer} or memorized samples \citep{feldman2020neural}. We extend the application of self influences to scenarios where there are no labels.

To test this application, we design an experiment to see if self influences can find a small amount of extra samples added to the original dataset. The extra samples are from other datasets: 1000 EMNIST \citep{cohen2017emnist} samples for MNIST, and 1000 CelebA \citep{liu2015faceattributes} samples for CIFAR, respectively. In Figure \ref{fig: rare auc}, we plot the detection curves to show fraction of extra samples found when all samples are sorted in the self influence order. The area under these detection curves (AUC) are 0.887 in the MNIST experiment and 0.760 in the CIFAR experiment. \footnote{AUC $\approx 1$ means the detection is near perfect, and AUC $\approx 0.5$ means the detection is near random.} Full results and more comparisons can be found in Appendix \ref{appendix: data cleaning}. The results indicate that extra samples generally have higher self influences than original samples, so it has much potential to apply VAE-TracIn to unsupervised data cleaning.

\subsection{Influences over Test Data}\label{sec: test inf}

\textbf{Question.} Which training samples are strong proponents or opponents of a test sample, respectively? Influences over a test sample $z$ provide rich information about the relationship between training data and $z$. In supervised learning, strong proponents help the model correctly predict the label of $z$ while strong opponents harm it. Empirically, strong proponents are visually similar samples from the same class, while strong opponents tend to confuse the model \citep{pruthi2020estimating}. In unsupervised learning, we expect that strong proponents increase the likelihood of $z$ and strong opponents reduce it. We examine which samples are strong proponents or opponents in VAE. 

\textbf{Methodology.} We train separate VAE models on MNIST, CIFAR, and each CIFAR subclass. We then compute VAE-TracIn scores of all training samples over 128 test samples. 

In MNIST experiments, we plot the distributions of influences according to whether training and test samples belong to the same class (See results on label zero in Figure \ref{fig: mnist test inf by class} and full results in Figure \ref{fig: mnist test inf by class appendix}). We then compare the influences of training over test samples to their distances in the latent space in Figure \ref{fig: mnist test inf by latent dist}. Quantitatively, we define samples that have the $0.1\%$ highest/lowest influences as the strongest proponents/opponents. Then, we report the fraction of the strongest proponents/opponents that belong to the same class as the test sample and the statistics of pairwise distances in Table \ref{tab: mnist test inf stat}. Additional comparisons can be found in Appendix \ref{appendix: test inf mnist},

In CIFAR and CIFAR subclass experiments, we compare influences of training over test samples to the norms of training samples in the latent space in Figure \ref{fig: cifar test-inf by latent norm appendix} and Figure \ref{fig: cifar subset test-inf by latent norm appendix}. Quantitatively, we report the statistics of the norms in Table \ref{tab: cifar test inf stat}. Additional comparisons can be found in Appendix \ref{appendix: test inf cifar}.

\textbf{Results.}
In MNIST experiments, we find many strong proponents and opponents of a test sample are its similar samples from the same class. In terms of class information, many ($\sim80\%$) strongest proponents and many ($\sim40\%$) strongest opponents have the same label as test samples. In terms of distances in the latent space, it is shown that the strongest proponents and opponents are close (thus similar) samples, while far away samples have small absolute influences. These findings are similar to GMM discussed in Section \ref{sec: classic}, where the strongest opponents may come from the same class (see Figure \ref{fig: classic test inf appendix}). The findings are also related to the supervised setting in the sense that dissimilar samples from a different class have small influences. 

Results in CIFAR and CIFAR subclass experiments indicate strong proponents have large norms in the latent space. \footnote{Large norm samples can be outliers, high-contrast samples, or very bright samples.} This observation also happens to many instance-based interpretations in the supervised setting including classification methods \citep{hanawa2020evaluation} and logistic regression \citep{barshan2020relatif}, where large norm samples can impact a large region in the data space, so they are influential to many test samples. 

\textbf{Visualization.} We visualize the strongest proponents and opponents in Figure \ref{fig: test inf visualization}. More visualizations can be found in Appendix \ref{appendix: test inf mnist} and Appendix \ref{appendix: test inf cifar}. In the MNIST experiment, the strongest proponents look very similar to test samples. The strongest opponents are often the same but visually different digits. For example, the opponents of the test "two" have very different thickness and styles. In CIFAR and CIFAR subclass experiments, we find strong proponents seem to match the color of the images -- including the background and the object -- and they tend to have the same but brighter colors. Nevertheless, many proponents are from the same class. Strong opponents, on the other hand, tend to have very different colors as the test samples.

\begin{figure}[!t]
    \centering
    
     \begin{minipage}[t]{0.6\textwidth}
      \captionsetup{type=table} 
             \caption{Statistics of influences, class information, and distances of train-test sample pairs in MNIST. "$+$" means top-$0.1\%$ strong proponents, "$-$" means top-$0.1\%$ strong opponents, and "all" means the train set. The first two rows are fractions of samples that belong to the same class as the test sample. The bottom three rows are means $\pm$ standard errors of latent space distances between train-test sample pairs.}
        \label{tab: mnist test inf stat}
        \centering
        \fontsize{8.5}{10}\selectfont
        \begin{tabular}{l|ccc}
           \hline
           $\dlatent$ & 64 & 96 & 128 \\ \hline
           same class rate ($+$) & $81.9\%$ & $84.0\%$ & $82.1\%$ \\
           same class rate ($-$) & $37.3\%$ & $43.3\%$ & $40.3\%$ \\ \hline
           distances ($+$) & $0.94\pm0.53$ & $0.94\pm0.55$ & $0.76\pm0.51$ \\
           distances ($-$) & $1.78\pm0.75$ & $1.84\pm0.78$ & $1.29\pm0.67$ \\
           distances (all) & $2.54\pm0.90$ & $2.57\pm0.91$ & $2.23\pm0.92$ \\ 
        \hline
        \end{tabular}
        \vspace{-0.15em}
    \end{minipage}
     \begin{minipage}[t]{0.36\textwidth}
     \captionsetup{margin=0.3cm}
     \captionsetup{type=table} 
             \caption{The means $\pm$ standard errors of latent space norms of training samples in CIFAR and CIFAR-Airplane. Notations follow Table \ref{tab: mnist test inf stat}. It is shown that strong proponents tend to have very large norms.}
        \label{tab: cifar test inf stat}
        \centering
        \fontsize{8.5}{10}\selectfont
        \begin{tabular}{cc|c}
           \hline
           \multirow{3}{*}{CIFAR} & ($+$) & $7.42\pm1.10$ \\
           & ($-$) & $3.89\pm1.26$ \\
           & (all) & $5.06\pm1.18$ \\ \hline
           \multirow{3}{*}{\shortstack[l]{CIFAR-\\Airplane}} & ($+$) & $4.73\pm0.78$ \\
           & ($-$) & $4.26\pm0.91$ \\
           & (all) & $4.07\pm0.83$ \\ \hline
        \end{tabular}
        \vspace{-0.1em}
    \end{minipage}
\end{figure}

\begin{figure}[!t]
    \centering
    \begin{tikzpicture}
	    \node (image) at (0,5.95) {
            \includegraphics[height=2.5cm]{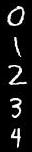}};
            \node (image) at (3,5.95) {
            \includegraphics[height=2.5cm]{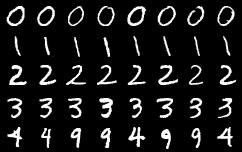}};
            \node (image) at (7.3,5.95) {
            \includegraphics[height=2.5cm]{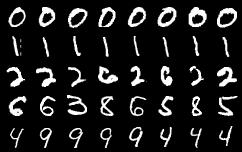}};
            \node (image) at (0,3.1) {
            \includegraphics[height=3cm]{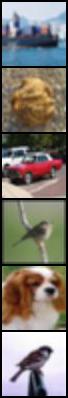}};
            \node (image) at (3,3.1) {
            \includegraphics[height=3cm]{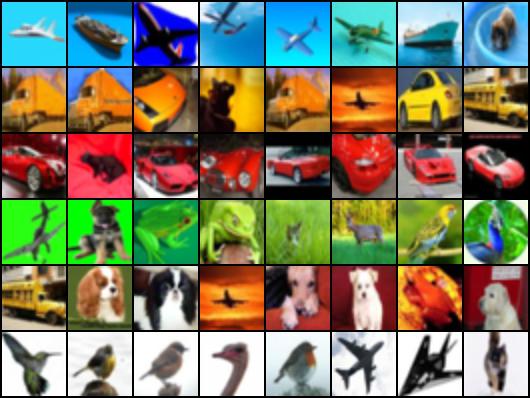}};
            \node (image) at (7.3,3.1) {
            \includegraphics[height=3cm]{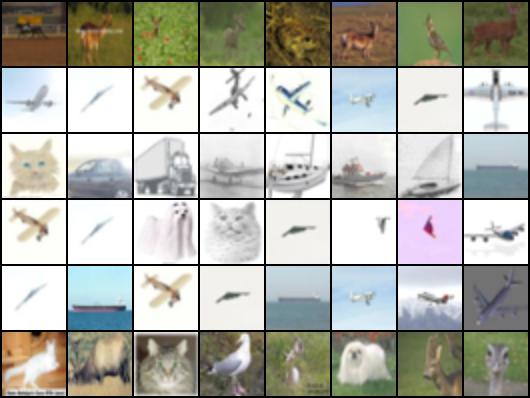}};
            \node (image) at (0,0) {
            \includegraphics[height=3cm]{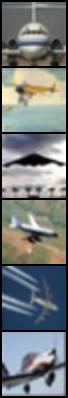}};
            \node (image) at (3,0) {
            \includegraphics[height=3cm]{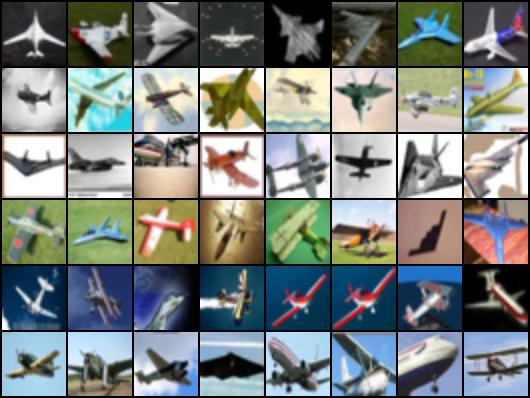}};
            \node (image) at (7.3,0) {
            \includegraphics[height=3cm]{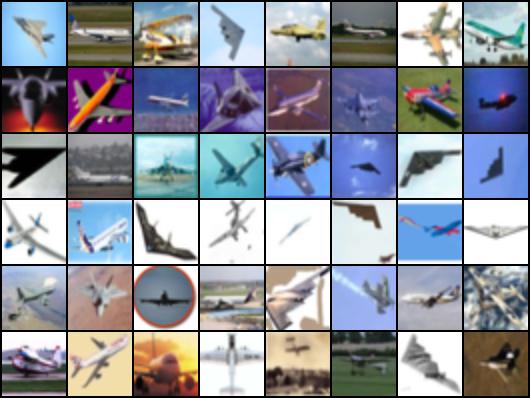}};
            \node[] at (0,7.35)  {\tiny{test samples}};
            \node[] at (3,7.35)  {\tiny{strongest proponents}};
            \node[] at (7.3,7.35)  {\tiny{strongest opponents}};
            \node[] at (-1.2,5.95)  {\tiny{MNIST}};
            \node[] at (-1.2,3.1)  {\tiny{CIFAR}};
            \node[] at (-1.2,0)  {\tiny{CIFAR-Airplane}};
    \end{tikzpicture}
    \vspace{-0.25em}
    \caption{Test samples from several datasets, their strongest proponents, and strongest opponents. In MNIST the strongest proponents are visually similar while the strongest opponents are often the same digit but are visually different. In CIFAR and CIFAR-Airplane the strongest proponents seem to match the colors and are often very bright or high-contrast.}
    \label{fig: test inf visualization}
\end{figure}

\subsection{Discussion}

\begin{table}[!t]
	\caption{High level summary of influence functions in classical unsupervised learning methods ($k$-NN, KDE and GMM) and VAE. In terms of self influences, VAE is similar to KDE, a non-parametric method. In terms of proponents and opponents, VAE trained on MNIST is similar to GMM, a parametric method. In addition, VAE trained on CIFAR is similar to supervised methods \citep{hanawa2020evaluation, barshan2020relatif}.}
	\vspace{0.5em}
	\label{tab: intuition}
        \centering
        \begin{tabular}{c|c|c}
           \hline
           \textbf{Method} & \textbf{high self influence samples} & \textbf{low self influence samples} \\ \hline
           $k$-NN & in a cluster of size exactly $k$ & -- \\
           KDE & in low density (sparse) region & in high density region \\
           GMM & far away to cluster centers & near cluster centers \\ \hdashline
           \multirow{2}{*}{VAE}& large loss & small loss \\
           & visually complicated or high-contrast & simple shapes or simple background \\
           \hline\hline
           
           \textbf{Method} & \textbf{strong proponents} & \textbf{strong opponents} \\ \hline
           $k$-NN & $k$ nearest neighbours & other than $k$ nearest neighbours \\
           KDE & nearest neighbours & farthest samples \\
           GMM & nearest neighbours & possibly far away samples in the same class \\ \hdashline
           {VAE$_{\rm{(MNIST)}}$} & nearest neighbors in the same class & far away samples in the same class \\
           {VAE$_{\rm{(CIFAR)}}$} & large norms and similar colors & different colors \\
           \hline 
        \end{tabular}
\end{table}

VAE-TracIn provides rich information about instance-level interpretability in VAE. In terms of self influences, there is correlation between self influences and VAE losses. Visually, high self influence samples are ambiguous or high-contrast while low self influence samples are similar in shape or background. In terms of influences over test samples, for VAE trained on MNIST, many proponents and opponents are similar samples in the same class, and for VAE trained on CIFAR, proponents have large norms in the latent space. {We summarize these high level intuitions of influence functions in VAE in Table \ref{tab: intuition}.} We observe there are strong connections between these findings and influence functions in KDE, GMM, classification and simple regression models.

\section{Conclusion}
Influence functions in unsupervised learning can reveal the most responsible training samples that increase the likelihood (or reduce the loss) of a particular test sample. In this paper, we investigate influence functions for several classical unsupervised learning methods and one deep generative model with extensive theoretical and empirical analysis. We present VAE-TracIn, a theoretical sound and computationally efficient algorithm that estimates influence functions for VAE, and evaluate it on real world datasets. 

One limitation of our work is that it is still challenging to apply VAE-TracIn to modern, huge models trained on a large amount of data, which is an important future direction. 
{There are several potential ways to scale up VAE-TracIn for large networks and datasets. First, we observe both positively and negatively influential samples (i.e. strong proponents and opponents) are similar to the test sample. Therefore, we could train an embedding space or a tree structure (such as the kd-tree) and then only compute VAE-TracIn values for similar samples. Second, because training at earlier epochs may be more effective than later epochs (as optimization is near convergence then), we could select a smaller but optimal subset of checkpoints to compute VAE-TracIn. Finally, we could use gradients of certain layers (e.g. the last fully-connected layer of the network as in \citet{pruthi2020estimating}).}

{Another important future direction is to investigate down-stream applications of VAE-TracIn such as detecting memorization or bias and performing data deletion or debugging.}

\newpage
\section*{Acknowledgements}
We thank NSF under IIS 1719133 and CNS 1804829 for research support. We thank Casey Meehan, Yao-Yuan Yang, and Mary Anne Smart for helpful feedback.

\bibliographystyle{plainnat}
\bibliography{main}

\newpage
\appendix
\addcontentsline{toc}{section}{Appendix}
\tableofcontents

\section{Omitted Proofs}\label{appendix: proofs}

\subsection{Omitted Proofs in Section \ref{sec: classic}}
\subsubsection{Proof of \eqref{eq: knn influence}}\label{appendix: knn}
\begin{proof}
By definition, we have
\[\pknn(z;X)=\frac{k}{N V_d R_k(z;X)^d}\]
and
\[\pknn(z;X_{-i})=\frac{k}{(N-1) V_d R_k(z;X_{-i})^d}.\]
If $x_i$ belongs to $k$-NN of $z$, then $R_k(z;X_{-i})$ is $R_{k+1}(z;X)$; otherwise, $R_k(z;X_{-i})$ is $R_k(z;X)$. The result follows by subtracting the logarithm of these two densities.
\end{proof}

\subsubsection{Proof of \eqref{eq: kde influence}}\label{appendix: kde}
\begin{proof}
By definition, we have
\[\pkde(z;X)=\frac1N\sum_{j=1}^N K_{\sigma}(z-x_j)\]
and
\[\pkde(z;X_{-i})=\frac{1}{N-1}\sum_{j\neq i}^N K_{\sigma}(z-x_j).\]
The result follows by subtracting the logarithm of these two densities. It is interesting to notice when $\max_iK_{\sigma}(z-x_i) \ll N\cdot\pkde(z;X)$, we have 
\[\sum_{i=1}^N \mathrm{IF}_{X,\mathrm{KDE}}(x_i,z) \approx \frac1N\sum_{i=1}^N\left(\frac{K_{\sigma}(z-x_i)}{\pkde(z;X)}-1\right) = 0.\] 
\end{proof}

\subsubsection{Proof of \eqref{eq: wsgmm influence}}\label{appendix: ws-gmm}
\begin{proof}
By definition, we have
\[\pwsgmm(z;X)=\frac{N_0}{N}\mN(z;\mu_0,\sigma_0^2I).\]
If $x_i\notin X_0$, then
\[\pwsgmm(z;X_{-i})=\frac{N_0}{N-1}\mN(z;\mu_0,\sigma_0^2I).\]
In this case, we have $\mathrm{IF}_{X,\mathrm{WS\text{-}GMM}}(x_i,z)=-\log(1+1/N)$. 

If $x_i\in X_0$, then parameters $\mu_0$ and $\sigma_0$ are to be modified to maximize likelihood estimates over $X_0\setminus\{x_i\}$. Denote the modified parameters as $\mu'_0$ and $\sigma'_0$. Then, we have
\[\pwsgmm(z;X_{-i})=\frac{N_0-1}{N-1}\mN(z;\mu'_0,(\sigma'_0)^2I).\]
Next, we express $\mu'_0$ and $\sigma'_0$ in terms of known variables. For conciseness, we let $v=z-\mu_0$ and $u=x_i-\mu_0$.
\[\begin{array}{rl}
    \mu'_0 
    & \displaystyle = \frac{1}{N_0-1}\sum_{x\in X_0\setminus\{x_i\}}x \\
    & \displaystyle = \frac{N_0\mu_0-x_i}{N_0-1} \\
    & \displaystyle = \mu_0 - \frac{u}{N_0-1}.
\end{array}\]
\[\begin{array}{rl}
    (\sigma'_0 )^2
    & \displaystyle = \frac{1}{(N_0-1)d}\sum_{x\in X_0\setminus\{x_i\}}x^{\top}x-\frac1d(\mu'_0)^{\top}\mu'_0 \\
    & \displaystyle = \frac{1}{(N_0-1)d}\left(\sum_{x\in X_0}x^{\top}x-x_i^{\top}x_i-(N_0-1)\left(\mu_0^{\top}\mu_0-\frac{2u^{\top}\mu_0}{N_0-1}+\frac{u^{\top}u}{(N_0-1)^2}\right)\right) \\
    & \displaystyle = \frac{1}{(N_0-1)d}\left(N_0d\sigma_0^2-x_i^{\top}x_i+\mu_0^{\top}\mu_0+2u^{\top}\mu_0-\frac{u^{\top}u}{N_0-1}\right) \\
    & \displaystyle = \frac{1}{(N_0-1)d}\left(N_0d\sigma_0^2-x_i^{\top}x_i-\frac{N_0u^{\top}u}{N_0-1}\right) \\
    & \displaystyle = \frac{N_0}{N_0-1}\sigma_0^2-\frac{N_0 u^{\top}u}{(N_0-1)^2d}.
\end{array}\]

Then, we have 
\[\begin{array}{rl}
    \log \pwsgmm(z;X) & \displaystyle = \log\frac{N_0}{N} -\frac d2\log2\pi -\frac d2 \log\sigma_0^2-\frac{1}{2\sigma_0^2}(z-\mu_0)^{\top}(z-\mu_0) \\
    & \displaystyle = \log\frac{N_0}{N} -\frac d2\log2\pi -\frac d2 \log\sigma_0^2-\frac{v^{\top}v}{2\sigma_0^2},
\end{array}\]
and
\[\begin{array}{rl}
    \log \pwsgmm(z;X_{-i})
    & \displaystyle = \log\frac{N_0-1}{N-1} -\frac d2\log2\pi -\frac d2 \log(\sigma'_0)^2-\frac{1}{2{\sigma'_0}^2}(z-\mu'_0)^{\top}(z-\mu'_0) \\
    & \displaystyle = \log\frac{N_0-1}{N-1} -\frac d2\log2\pi -\frac d2 \log\frac{N_0}{N_0-1} - \frac d2 \log\sigma_0^2  \\
    & \displaystyle ~~~~ - \frac d2 \log\left(1-\frac{u^{\top}u}{(N_0-1)d\sigma_0^2}\right) \\
    \left(z-\mu'_0=v+\frac{u}{N_0-1}\right)
    & \displaystyle ~~~~ - \frac{(N_0-1)^2v^{\top}v+2(N_0-1)u^{\top}v+u^{\top}u}{2N_0\left((N_0-1)\sigma_0^2-\frac1du^{\top}u\right)} \\
    & \displaystyle = \log\frac{N_0-1}{N-1} -\frac d2\log2\pi -\frac{d}{2N_0} - \frac d2 \log\sigma_0^2 + \frac{u^{\top}u}{2N_0\sigma_0^2} \\
    & \displaystyle ~~~~ - \frac{v^{\top}v}{2\sigma_0^2} - \frac{1}{2N_0\sigma_0^2}\left(2u^{\top}v-v^{\top}v+\frac{v^{\top}v}{\sigma_0^2}\right) + \mO{N_0^{-2}}. \\
\end{array}\]
Subtracting the above two equations, we have
\[\begin{array}{rl}
    \mathrm{IF}_{X,\mathrm{WS\text{-}GMM}}(x_i,z) 
    & \displaystyle = \frac{1}{N_0}-\frac{1}{N} +\frac{d}{2N_0} + \frac{1}{2N_0\sigma_0^2}\left(2u^{\top}v-v^{\top}v-u^{\top}u+\frac{v^{\top}v}{\sigma_0^2}\right) + \mO{N_0^{-2}} \\
    & \displaystyle = \frac{d+2}{2N_0} + \frac{1}{2N_0\sigma_0^2}\left(\frac{\|z-\mu_0\|^2}{\sigma_0^2}-\|z-x_i\|^2\right) -\frac{1}{N} + \mO{N_0^{-2}}.
\end{array}\]

\end{proof}

\subsection{Probabilistic Bound on Influence Estimates}\label{sec: prob bound VAE appendix}

Let $\{\xi_j\}_{j=1}^m$ be $m$ i.i.d. samples drawn from $Q_{\psi^*}(\cdot|z)$ and $\{\xi'_j\}_{j=1}^m$ be $m$ i.i.d. samples drawn from $Q_{\psi_{-i}^*}(\cdot|z)$. We can use the empirical influence $\hat{\mathrm{IF}}_{X,\mathrm{VAE}}^{(m)}(x_i,z)$ to estimate the true influence in \eqref{eq: beta vae inf def}, which is defined below:

\begin{equation}\label{eq: beta vae inf empirical}
\begin{array}{rl}
    \hat{\mathrm{IF}}_{X,\mathrm{VAE}}^{(m)}(x_i,z) & = \beta\left(\KL{Q_{\psi_{-i}^*}(\cdot|z)}{\platent}-\KL{Q_{\psi^*}(\cdot|z)}{\platent}\right) \\
    & ~~ - \frac1m\sum_{j=1}^m\left(\log P_{\phi_{-i}^*}(z|\xi'_j) - \log P_{\phi^*}(z|\xi_j)\right).
\end{array}
\end{equation}

The questions is, when can we guarantee the empirical influence score $\hat{\mathrm{IF}}_{X,\mathrm{VAE}}^{(m)}(x_i,z)$ is close to the true influence score $\mathrm{IF}_{X,\mathrm{VAE}}(x_i,z)$? We answer this question via an $(\epsilon,\delta)$-probabilistic bound: as long as $m$ is larger than a function of $\epsilon$ and $\delta$, then with probability at least $1-\delta$, the difference between the empirical and true influence scores is no more than $\epsilon$. To introduce the theory, we first provide the following definition. 

\begin{definition}[Polynomially-bounded functions]\label{def: poly bound}
Let $f:\mathbb{R}^d\rightarrow\mathbb{R}^{d'}$. We say $f$ is polynomially bounded by $\{a_c\}_{c=1}^C$ if for any $x\in\mathbb{R}^d$, we have
\begin{equation}\label{eq: poly bound def}
    \|f(x)\| \leq \sum_{c=1}^C a_c\|x\|^c.
\end{equation}
\end{definition}

We provide a useful lemma on polynomially-bounded functions below.

\begin{lemma}\label{lemma: composition poly bound}
The composition of polynomially bounded functions is polynomially bounded. 
\end{lemma}

Next, we show common neural networks are polynomially bounded in the following proposition.

\begin{proposition}\label{prop: neural net poly bound}
Let $f$ be a neural network taking the following form:
\begin{equation}\label{eq: neural net def}
    f(x)=\sigma_l(W_l\sigma_{l-1}(W_{l-1}\cdots\sigma_1(W_1x)\cdots)).
\end{equation}
If every activation function $\sigma_j$ is polynomially bounded, then $f$ is polynomially bounded. 
\end{proposition}

With the above result, we state the $(\epsilon,\delta)$-probabilistic bound on influence estimates below. 

\begin{theorem}[Error bounds on influence estimates]\label{thm: error bound beta vae} 
Let $P$ and $Q$ be two polynomially bounded networks. For any small $\epsilon>0$ and $\delta>0$, there exists an $m=\OTheta{\frac{1}{\epsilon^2\delta}}$ such that
\begin{equation}
    \prob{\left|\mathrm{IF}_{X,\mathrm{VAE}}(x_i,z)-\hat{\mathrm{IF}}_{X,\mathrm{VAE}}^{(m)}(x_i,z)\right|\geq \epsilon} \leq \delta,
\end{equation}
where the randomness is over all $\xi_j$ and $\xi'_j$.
\end{theorem}

\subsubsection{Proof of \textbf{Lemma} \ref{lemma: composition poly bound}}\label{appendix: lemma polynomial}
\begin{proof}
Let $f:\mathbb{R}^{m_0}\rightarrow\mathbb{R}^{m_1}$ be polynomially bounded by $\{a_c\}_{c=1}^{C_f}$ and $g:\mathbb{R}^{m_1}\rightarrow\mathbb{R}^{m_2}$ be polynomially bounded by $\{b_c\}_{c=1}^{C_g}$. Then, for any $x\in\mathbb{R}^{m_0}$,
\[\|f(x)\|\leq\sum_{c=1}^{C_f}a_c\|x\|^c,\]
and
\[\|g\circ f(x)\|\leq\sum_{c=1}^{C_g}b_c\|f(x)\|^c.\]
Therefore, we have
\[\|g\circ f(x)\|\leq\sum_{c=1}^{C_g}b_c\left(\sum_{c'=1}^{C_f}a_{c'}\|x\|^{c'}\right)^c.\]
This indicates that $g\circ f$ is polynomially bounded.
\end{proof}

\subsubsection{Proof of \textbf{Proposition} \ref{prop: neural net poly bound}}\label{appendix: prop nn polynomial}
\begin{proof}
First, an affine transformation $Wx$ is polynomially bounded because $\|Wx\|\leq \|W\|\cdot\|x\|$. Then, we show an element-wise transformation $\sigma(x)$ is polynomially bounded. Let $\sigma$ be polynomially bounded by $\{a_c\}_{c=1}^{C}$. Then,
\[\|\sigma(x)\| \leq \frac{1}{\sqrt{d}}\left(\sum_{i=1}^d |\sigma(x_i)|\right)^2 \leq \frac{1}{\sqrt{d}}\left(\sum_{i=1}^d \left|\sum_{c=1}^C a_c|x_i|^c\right|\right)^2 \leq \frac{1}{\sqrt{d}}\left(\sum_{i=1}^d \left|\sum_{c=1}^C a_c\|x\|^c\right|\right)^2. \]
By \textbf{Lemma} \ref{lemma: composition poly bound}, since $f$ is a composition of polynomially bounded functions, we have $f$ is polynomially bounded.
\end{proof}

\subsubsection{Proof of \textbf{Theorem} \ref{thm: error bound beta vae}}\label{appendix: vae error bound}
\begin{proof}
According to \eqref{eq: beta vae inf def} and \eqref{eq: beta vae inf empirical},
\[\begin{array}{rl}
    \displaystyle \mathrm{IF}_{X,\mathrm{VAE}}(x_i,z)-\hat{\mathrm{IF}}_{X,\mathrm{VAE}}^{(m)}(x_i,z) 
    & \displaystyle = \frac1m\sum_{j=1}^m \log P_{\phi_{-i}^*}(z|\xi'_j) - \mathbb{E}_{\xi\sim Q_{\psi_{-i}^*}(\cdot|z)} \log P_{\phi_{-i}^*}(z|\xi) \\
    & \displaystyle ~~ - \frac1m\sum_{j=1}^m \log P_{\phi^*}(z|\xi_j) + \mathbb{E}_{\xi\sim Q_{\psi^*}(\cdot|z)} \log P_{\phi^*}(z|\xi).
\end{array}\]
If we have
\[\left|\mathbb{E}_{\xi\sim Q_{\psi^*}(\cdot|z)} \log P_{\phi^*}(z|\xi) - \frac1m\sum_{j=1}^m \log P_{\phi^*}(z|\xi_j)\right| \leq \frac{\epsilon}{2}\]
and
\[\left|\mathbb{E}_{\xi\sim Q_{\psi_{-i}^*}(\cdot|z)} \log P_{\phi_{-i}^*}(z|\xi) - \frac1m\sum_{j=1}^m \log P_{\phi_{-i}^*}(z|\xi'_j)\right| \leq \frac{\epsilon}{2},\]
then $\left|\mathrm{IF}_{X,\mathrm{VAE}}(x_i,z)-\hat{\mathrm{IF}}_{X,\mathrm{VAE}}^{(m)}(x_i,z) \right|\leq\epsilon$. Let $\zeta$ and $\zeta_j$ be i.i.d. standard Gaussian random variables for $j=1,2,\cdots,m$. First, we provide the probabilistic bound for the first inequality. Let $P=P_{\phi^*}$ and $Q=Q_{\psi^*}$. Then, we can reparameterize $\xi = \mu_Q(z)+\sigma_Q(z)\odot\zeta$ and $\xi_j = \mu_Q(z)+\sigma_Q(z)\odot\zeta_j$. Let 
\[f(\zeta) = \left\|z-\mu_P(\mu_Q(z)+\sigma_Q(z)\odot\zeta)\right\|_2^2.\]
Since $\log P(z|\xi)$ is a constant $\alpha$ times $\|z-\mu_P(\xi)\|^2$ plus another constant, we have 
\[\mathbb{E}_{\xi\sim Q(\cdot|z)} \log P(z|\xi) - \frac1m\sum_{j=1}^m \log P(z|\xi_j) = \mathbb{E}_{\zeta}f(\zeta)-\frac1m\sum_{j=1}^m f(\zeta_j).\]
By Chebyshev’s inequality,
\[\prob{\left|\mathbb{E}_{\zeta}f(\zeta)-\frac1m\sum_{j=1}^m f(\zeta_j)\right|\geq\frac{\epsilon}{2}} \leq \frac{4\mathrm{Var}_{\zeta}f(\zeta)}{m\epsilon^2} \leq \frac{4\mathbb{E}_{\zeta}f(\zeta)^2}{m\epsilon^2}.\]
 
By \textbf{Lemma} \ref{lemma: composition poly bound}, if $P$ and $Q$ are polynomially bounded, then $f$ is polynomially bounded and so is $f^2$. Let 
\[f(\zeta)^2\leq\sum_{c=1}^Ca_c\|\zeta\|^c.\]
Then, 
\[\begin{array}{rl}
    \mathbb{E}_{\zeta}f(\zeta)^2
    & \displaystyle = \int_{\mathbb{R}^d} \mathcal{N}(\zeta;0,I) f(\zeta)^2 d\zeta \\
    & \displaystyle \leq \frac{1}{(2\pi)^{\frac d2}}\sum_{c=1}^C a_c \int_{\mathbb{R}^d} \|\zeta\|^c e^{-\frac{\|\zeta\|^2}{2}} d\zeta \\
    (\text{use polar coordinate}) 
    & \displaystyle = \frac{\pi^{\frac{d-1}{2}}}{(2\pi)^{\frac d2}\Gamma\left(\frac{d+1}{2}\right)} \sum_{c=1}^C a_c \int_{0}^{+\infty} r^{c+d-1} e^{-\frac{r^2}{2}} dr \\
    & \displaystyle = \frac{\pi^{\frac{d-1}{2}}}{(2\pi)^{\frac d2}\Gamma\left(\frac{d+1}{2}\right)} \sum_{c=1}^C a_c \int_{0}^{+\infty} (2r)^{\frac{c+d}{2}-1} e^{-r} dr \\
    & \displaystyle = \frac{1}{2\sqrt{\pi}\Gamma\left(\frac{d+1}{2}\right)}\sum_{c=1}^C 2^{\frac c2}a_c \Gamma\left(\frac{c+d}{2}\right),
\end{array}\]
which is a constant. Therefore, there exists an $M_1=\OTheta{\frac{1}{\epsilon^2\delta}}$ such that when $m\geq M_1$,
\[\prob{\left|\mathbb{E}_{\zeta}f(\zeta)-\frac1m\sum_{j=1}^m f(\zeta_j)\right|\geq\frac{\epsilon}{2}} \leq \frac{\delta}{2},\]
or
\[\prob{\left|\mathbb{E}_{\xi\sim Q_{\psi^*}(\cdot|z)} \log P_{\phi^*}(z|\xi) - \frac1m\sum_{j=1}^m \log P_{\phi^*}(z|\xi_j)\right| \geq \frac{\epsilon}{2}} \leq \frac{\delta}{2}.\]
Similarly, when $P=P_{\phi_{-i}^*}$ and $Q=Q_{\psi_{-i}^*}$, there exists an $M_2=\OTheta{\frac{1}{\epsilon^2\delta}}$ such that when $m\geq M_2$,
\[\prob{\left|\mathbb{E}_{\xi\sim Q_{\psi_{-i}^*}(\cdot|z)} \log P_{\phi_{-i}^*}(z|\xi) - \frac1m\sum_{j=1}^m \log P_{\phi_{-i}^*}(z|\xi'_j)\right| \geq \frac{\epsilon}{2}} \leq \frac{\delta}{2}.\]
Taking $m=\max(M_1, M_2)=\OTheta{\frac{1}{\epsilon^2\delta}}$, we have
\[\prob{\left|\mathrm{IF}_{X,\mathrm{VAE}}(x_i,z)-\hat{\mathrm{IF}}_{X,\mathrm{VAE}}^{(m)}(x_i,z)\right|\geq\epsilon} \leq\delta.\]
\end{proof}

\subsection{Derivation of \eqref{eq: beta vae loss gradient}}\label{appendix: gradient loss}
\[\begin{array}{rl}
	\nabla_{\phi}\ell_{\beta}(x;\theta) 
	& \displaystyle = -\nabla_{\phi} \mathbb{E}_{\xi\sim Q_{\psi}(\cdot|x)} \log P_{\phi}(x|\xi) \\
	& \displaystyle = -\mathbb{E}_{\xi\sim Q_{\psi}(\cdot|x)} \nabla_{\phi}\log P_{\phi}(x|\xi) ;\\
	\nabla_{\psi}\ell_{\beta}(x;\theta) 
	& \displaystyle = \nabla_{\psi} \int_{\xi} Q_{\psi}(\xi|x)\left(\beta\log\frac{Q_{\psi}(\xi|x)}{P_{\mathrm{latent}}(\xi)}-\log P_{\phi}(x|\xi)\right) d\xi \\
	& \displaystyle = \int_{\xi} \left[\nabla_{\psi}Q_{\psi}(\xi|x)\left(\beta\log\frac{Q_{\psi}(\xi|x)}{P_{\mathrm{latent}}(\xi)}-\log P_{\phi}(x|\xi)\right) + \beta Q_{\psi}(\xi|x)\cdot\frac{\nabla_{\psi}Q_{\psi}(\xi|x)}{Q_{\psi}(\xi|x)}\right] d\xi \\
    	& \displaystyle = \int_{\xi} \nabla_{\psi}Q_{\psi}(\xi|x)\left(\beta+\beta\log\frac{Q_{\psi}(\xi|x)}{P_{\mathrm{latent}}(\xi)}-\log P_{\phi}(x|\xi)\right) d\xi \\
    	& \displaystyle = \int_{\xi} \nabla_{\psi}Q_{\psi}(\xi|x)\left(\beta\log\frac{Q_{\psi}(\xi|x)}{P_{\mathrm{latent}}(\xi)}-\log P_{\phi}(x|\xi)\right) d\xi \\
    	& \displaystyle = \mathbb{E}_{\xi\sim Q_{\psi}(\cdot|x)} \nabla_{\psi}\log Q_{\psi}(\xi|x) \left(\beta\log\frac{Q_{\psi}(\xi|x)}{P_{\mathrm{latent}}(\xi)}-\log P_{\phi}(x|\xi)\right).
\end{array}\]

\newpage
\section{Additional Experiments and Details}\label{appendix: additional experiments}
\subsection{Additional Results on Density Estimators in Section \ref{sec: classic}}\label{appendix: classic figures}

\begin{figure}[!h]
    \centering
    \includegraphics[width=0.5\textwidth]{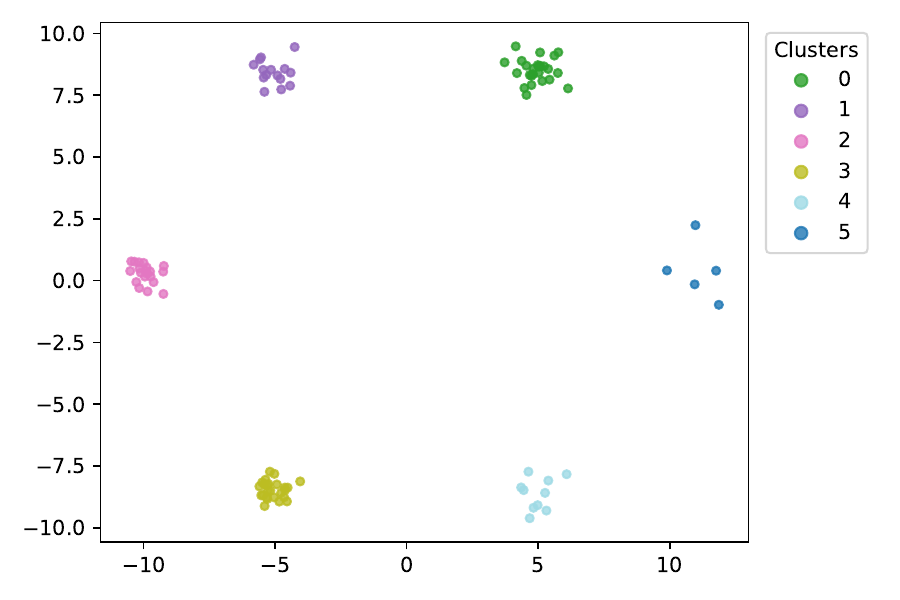}
    \caption{Synthetic data of six clusters in Section \ref{sec: classic visualization}.}
    \label{fig: classic data appendix}
\end{figure}

The synthetic data of six clusters are illustrated in Figure \ref{fig: classic data appendix}. The sizes for cluster zero through five are 25, 15, 20, 25, 10, 5, respectively. Each cluster is drawn from a spherical Gaussian distribution. The standard errors are 0.5, 0.5, 0.4, 0.4, 0.5, 1.0, respectively. 

\begin{figure}[!h]
    \centering
    \subfloat[][Self influence scores of training samples in different methods. The high self influence samples in $k$-NN are from a cluster with exactly $k$ samples; those in KDE are from the cluster with the smallest size; and those in GMM are far away to the center of the corresponding cluster.]{ 
    	 \includegraphics[width=0.99\textwidth]{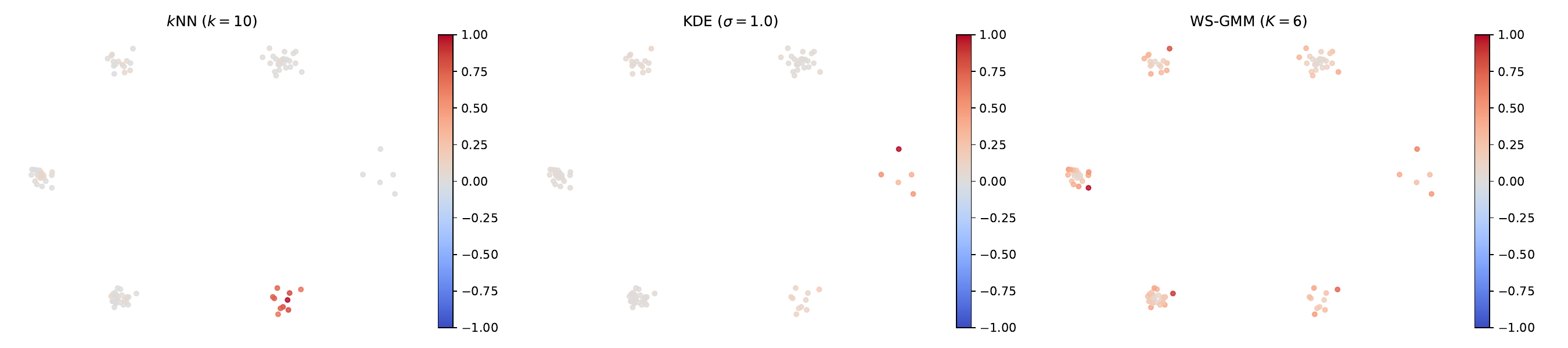}
    }\\
    \subfloat[][Self influences versus log-likelihood. In $k$-NN only samples from cluster 4 (which has exatly $k$ points) have large self influences. In KDE and GMM the self influences tend to decrease as the likelihood increases.]{ 
    	 \includegraphics[width=0.99\textwidth]{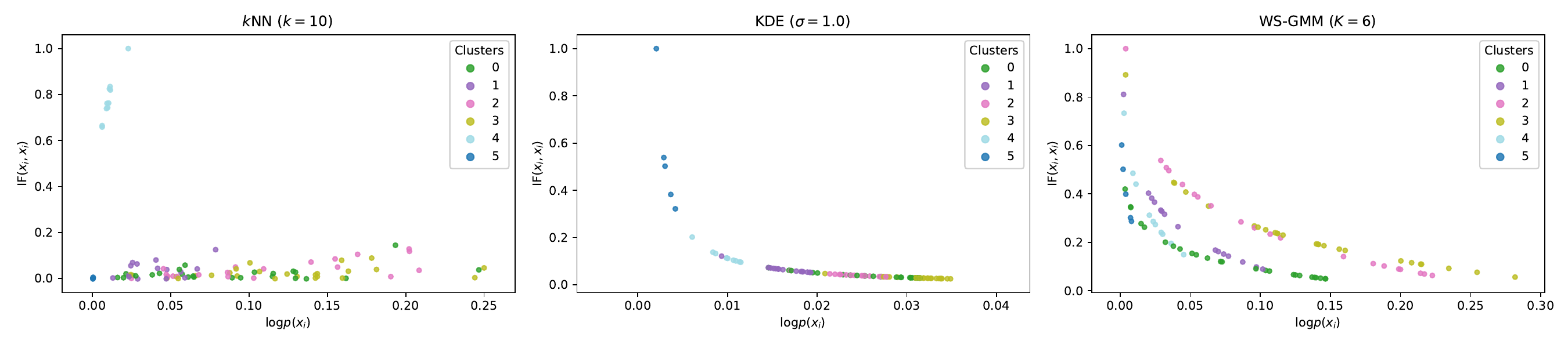}
	 \label{fig: classic self inf vs log p appendix}
    }
    \caption{Self influences in different density estimators.}
    \label{fig: classic self inf appendix}
\end{figure}

We visualize the self influences of all data samples, and compare them to the log likelihood in Figure \ref{fig: classic self inf appendix}. For $k$-NN samples from cluster 4 have the highest self influences because the size of this cluster is exactly $k=10$. For KDE samples in cluster 5 (which is the smallest cluster in terms of number of samples) have the highest self influences. The self influences strictly obey the reverse order of likelihood, which can be derived from \eqref{eq: kde influence}. For GMM samples far away to cluster centers have high self influences, which can be derived from \eqref{eq: wsgmm self influence}. Samples from clusters 2 and 3 generally have higher self influences because $\sigma_2$ and $\sigma_3$ are smaller than others.

\begin{figure}[!h]
    \centering
    \subfloat[][Influences of training samples over a test sample $z$ (shown as \textcolor{ForestGreen}{\ding{54}}) in different methods. In all cases the strongest proponents are nearest samples. In GMM, surprisingly, samples from the same cluster can be strong opponents.]{ 
        \begin{overpic}[width=0.99\textwidth]{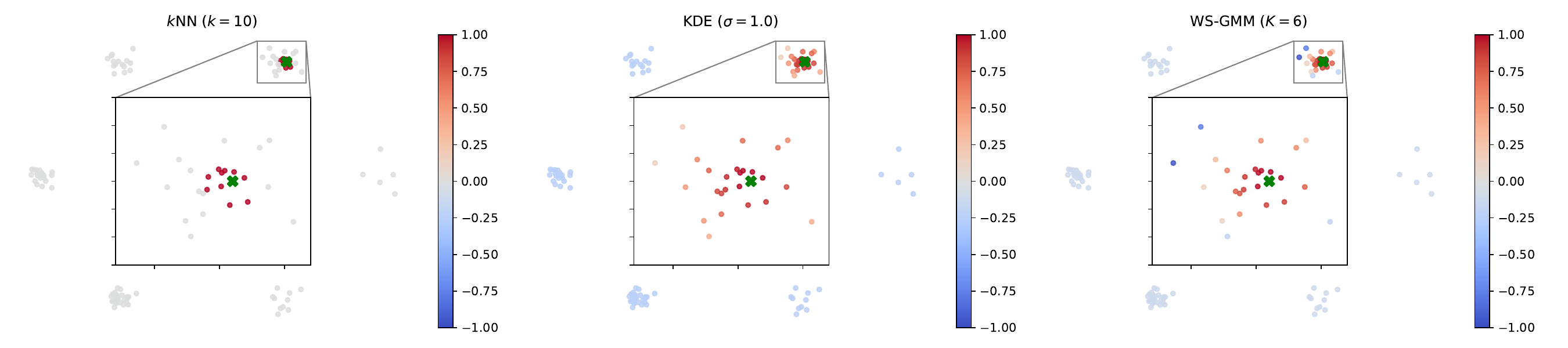}
            \linethickness{2pt}
            \put(70,17){\color{red}\vector(2,-1){5.6}}
            \put(70,17){\color{red}\vector(1,-1){4.5}}
            \put(65,16.5){\textcolor{red}{\tiny{strong}}}
            \put(65,15){\textcolor{red}{\tiny{opponents}}}
            \put(80.2,4){\color{red}\vector(-1,2){1.3}}
            \put(80.2,4){\color{red}\vector(1,1){3.6}}
            \put(76,2.5){\textcolor{red}{\tiny{opponents}}}
        \end{overpic}
        \label{fig: classic test inf samples appendix}
    }\\
    \subfloat[][Influences of training samples over $z$ versus distances to $z$.]{
    	\includegraphics[width=0.99\textwidth]{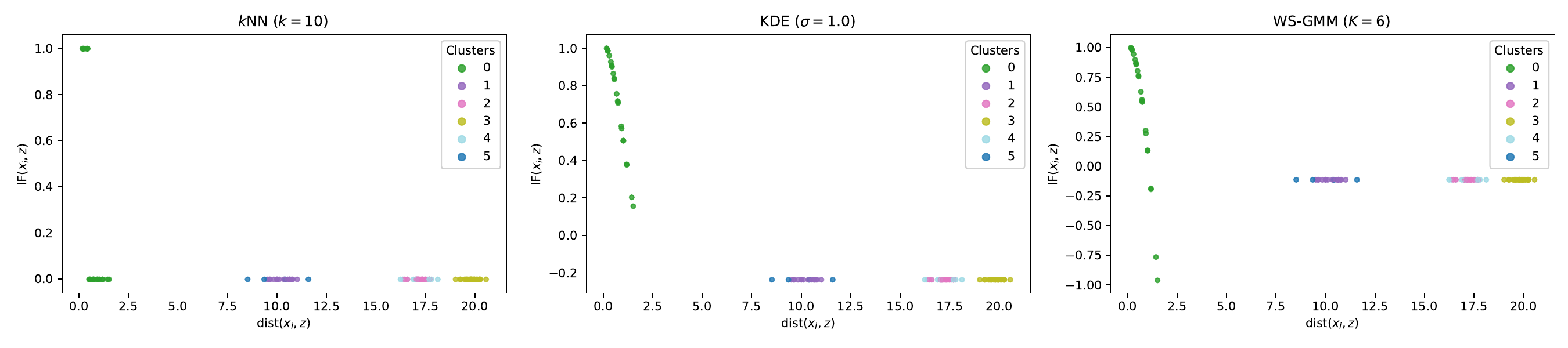}
        \label{fig: classic test inf vs dist appendix}
    } 
    \caption{Influences of training samples over a test sample $z$ in different methods. In all methods the strongest proponents are nearest samples. Surprisingly, in GMM strong opponents are also nearby samples.}
    \label{fig: classic test inf appendix}
\end{figure}

\newpage
We let $z$ be a data point near the center of cluster 0. We then visualize influences of all data samples over $z$, and compare these influences to the distances between the samples and $z$ in Figure \ref{fig: classic test inf appendix}. For $k$-NN the $k$ nearest samples are strong proponents of $z$, and the rest have little influences over $z$. For KDE proponents of $z$ are all samples from cluster 0, and the rest have slightly negative influences over $z$. The influences strictly obey the reverse order of distances to $z$, which can be derived from \eqref{eq: kde influence}. For GMM, it is surprising that samples in the same cluster as $z$ can be (even strong) opponents of $z$. This observation can be mathematically derived from \eqref{eq: wsgmm influence}. When $\|z-x_i\|^2 > (d+2)\sigma_0^2+\|z-\mu_0\|^2/\sigma_0^2$, we have $\mathrm{IF}_{X,\mathrm{WS\text{-}GMM}}(x_i,z)\lessapprox0$. When $d$ is large and $z$ is sampled from the mixture $\mN(\mu_0,\sigma_0^2I)$, then with high probability, $\|z-\mu_0\|^2\approx d\sigma_0^2$. Therefore, the influence of $x_i$ over $z$ is negative with high probability when $\|z-x_i\|^2\gtrapprox (1+\sigma_0^2)d+2\sigma_0^2$.

~~\newpage
\subsection{Details of Experiments in Section \ref{sec: experiments}}\label{appendix: details}

\paragraph{Datasets.} We conduct experiments on MNIST and CIFAR-10 (shortened as CIFAR). Because it is challenging to train a successful VAE model on the entire CIFAR dataset, we also train VAE models on each subclass of CIFAR. There are ten subclasses in total, which we name CIFAR$_0$ though CIFAR$_9$, and each subclass contains 5k training samples. In the main text, CIFAR-Airplane is CIFAR$_0$. All CIFAR images are resized to $64\times64$.

In Section \ref{sec: sanity check}, we examine influences of all training samples over the first 128 training samples in the trainset. In the unsupervised data cleaning application in Section \ref{sec: self inf}, the extra samples are the first 1k samples from EMNIST and CelebA, respectively.  In Section \ref{sec: test inf}, we randomly select 128 samples from the test set and compute influences of all training samples over these test samples. 

\paragraph{Models and hyperparameters.} For MNIST, our VAE models are composed of multilayer perceptrons as described by \citet{meehan2020non}. In these experiments we let $\beta=4$ and $\dlatent=128$ unless clearly specified. 

For CIFAR and CIFAR subclasses, our VAE models are composed of convolution networks as described by \citet{higgins2016beta}. We let $\beta=2,\dlatent=128$ for CIFAR and $\beta=2,\dlatent=64$ for CIFAR subclass unless clearly specified. 

We use stochastic gradient descent to train these VAE models based on a public implementation. \footnote{\url{https://github.com/1Konny/Beta-VAE} (MIT License)} In all experiments, we set the batch size to be 64 and train for 1.5M iterations. The learning rates are $1\times10^{-4}$ in MNIST experiments and $3\times10^{-4}$ in CIFAR experiments.

\paragraph{VAE-TracIn settings.} In all experiments, we average the loss for $m=16$ times when computing VAE-TracIn according to \eqref{eq: VAE-TracIn}. We use $C=30$ (evenly distributed) checkpoints to compute influences in Section \ref{sec: sanity check} and Section \ref{sec: test inf}. We use the last checkpoint to compute self influences in Section \ref{sec: self inf}. For visualization purpose, all self influences are normalized to $[0,1]$, all influences over test data are normalized to $[-1,1]$, and all distributions are normalized to densities.

~~\newpage
\subsection{Sanity Checks for VAE-TracIn}\label{appendix: validity}

As a sanity check for VAE-TracIn, we examine the frequency that a training sample is the most influential one among all training samples over itself, or formally, the frequency that $i=\arg\max_{1\leq i'\leq N} \mathrm{VAE\text{-}TracIn}(x_{i'},x_i)$. Due to computational limits we examine the first 128 training samples. The results for MNIST, CIFAR, and CIFAR subclasses are reported in Table \ref{tab: validity} and Table \ref{tab: cifar k validity appendix}. These results indicate that VAE-TracIn can find the most influential training samples in MNIST and CIFAR subclasses.

\begin{table}[!h]
    \caption{Sanity check on the frequency of a training sample being more influential than other samples over itself. Results on CIFAR subclasses (CIFAR$_i$, $0\leq i\leq 9$) are reported.}
    \vspace{0.5em}
    \label{tab: cifar k validity appendix}
    \centering
    \begin{tabular}{c|cccccccccc}
    \hline
        $i$ & 0 & 1 & 2 & 3 & 4 & 5 & 6 & 7 & 8 & 9 \\ \hline
        Top-1 scores & 1.000 & 1.000 & 0.992 & 1.000 & 0.984 & 1.000 & 1.000 & 1.000 & 1.000 & 1.000 \\ \hline
    \end{tabular}
\end{table}

\begin{figure}[!h]
    \centering
    \includegraphics[width=0.7\textwidth]{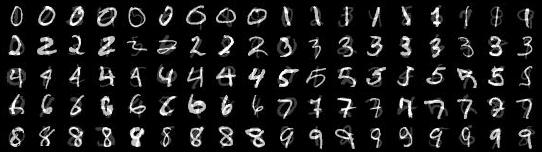}
    \caption{Synthesized samples $\hat{x}=\alpha x_{\mathrm{major}} + (1-\alpha)x_{\mathrm{minor}}$, where $\alpha=0.75$. }
    \label{fig: mnist 75-25 appendix}
\end{figure}

We next conduct an additional sanity check for VAE-TracIn on MNIST. For two training samples $x_{\mathrm{major}}=x_i$ and $x_{\mathrm{minor}}=x_j$, we synthesize a new sample $\hat{x}=\alpha x_{\mathrm{major}} + (1-\alpha)x_{\mathrm{minor}}$, where $\alpha=0.75$. Then, $\hat{x}$ is very similar to $x_{\mathrm{major}}$ but the minor component $x_{\mathrm{minor}}$ can also be visually recognized. For each pair of different labels, we obtain $x_{\mathrm{major}}$ and $x_{\mathrm{minor}}$ by randomly picking one sample within each class. The entire 90 samples are shown in Figure \ref{fig: mnist 75-25 appendix}. We expect a perfect instance-based interpretation should indicate $x_i$ and $x_j$ have very high influences over $\hat{x}$. We report quantiles of the 90 ranks of $x_{\mathrm{major}}$ and $x_{\mathrm{minor}}$ sorted by influences over $\hat{x}$ in Table \ref{tab: mnist 75-25 appendix}. We then compute the frequency that $x_{\mathrm{major}}$ is exactly the strongest proponent of $\hat{x}$, namely the top-1 score of the major component. We compare the results to a baseline model that finds nearest neighbours in a perceptual autoencoder latent space (PAE-NN, \citep{meehan2020non,zhang2018unreasonable}). Although VAE-TracIn does not detect $x_{\mathrm{major}}$ as well as PAE-NN, it still has reasonable results, and performs much better in detecting $x_{\mathrm{minor}}$. The results indicate that VAE-TracIn can capture potentially influential components. 

\vspace{-1.5em}

\begin{table}[!h]
    \caption{Quantiles of ranks of $x_{\mathrm{major}}$ and $x_{\mathrm{minor}}$ sorted by influences over $\hat{x}$, and top-1 scores of the major components. We let 0 to be the highest rank.}
    \vspace{0.5em}
    \label{tab: mnist 75-25 appendix}
    \centering
    \begin{tabular}{l|cccc|ccc}
        \hline
        \multirow{2}{*}{Method} & \multicolumn{3}{c}{rank($x_{\mathrm{major}}$) quantiles} & Top-1 & \multicolumn{3}{c}{rank($x_{\mathrm{minor}}$) quantiles} \\
        & 25\% & 50\% & 75\% & scores & 25\% & 50\% & 75\% \\ \hline
        PAE-NN  & 0 & 0 & 1   & 0.633 & 6943 & 13405 & 29993 \\ \hline
        VAE-TracIn ($\dlatent=64$)  & 0 & 2 & 146 & 0.422 & 1097 & 5206 & 10220 \\
        VAE-TracIn ($\dlatent=96$)  & 0 & 1 & 44  & 0.456 & 1372 & 4283 & 15319 \\
        VAE-TracIn ($\dlatent=128$) & 0 & 1 & 18  & 0.467 & 1203 & 6043 & 13873 \\ \hline
    \end{tabular}
\end{table}

We then evaluate the approximation accuracy of VAE-TracIn. We randomly select 128 test samples for each CIFAR-subclass and save 1000 checkpoints at the first 1000 iterations. We compute the total Pearson correlation coefficients between (1) the VAE-TracIn scores and (2) the loss change of all training samples on the selected test samples between consecutive checkpoints, similar to Appendix G by \citet{pruthi2020estimating}. The results are reported in Table \ref{tab: approximation appendix}, which indicate high correlations.

\vspace{-0.5em}
\begin{table}[!h]
    \centering
    \caption{Pearson correlation coefficients $\rho$ between VAE-TracIn scores and loss change on each CIFAR-subclass (CIFAR$_i$, $0\leq i\leq 9$). A coefficient $= 1$ means perfect correlation.}
    \vspace{0.5em}
    \begin{tabular}{r|cccccccccc}
    \hline
        $i$ & 0 & 1 & 2 & 3 & 4 & 5 & 6 & 7 & 8 & 9 \\ \hline
        $\rho$ & 0.882 & 0.910 & 0.895 & 0.934 & 0.828 & 0.865 & 0.908 & 0.852 & 0.861 & 0.613 \\
    \hline
    \end{tabular}
    \label{tab: approximation appendix}
\end{table}

\newpage
\subsection{Self Influences (MNIST)}\label{appendix: self-inf mnist}

\begin{figure}[!h]
    \centering
    \subfloat[][$\beta=1$, $\dlatent=128$]{ 
    	\includegraphics[trim=10 0 30 30, clip, width=0.32\textwidth]{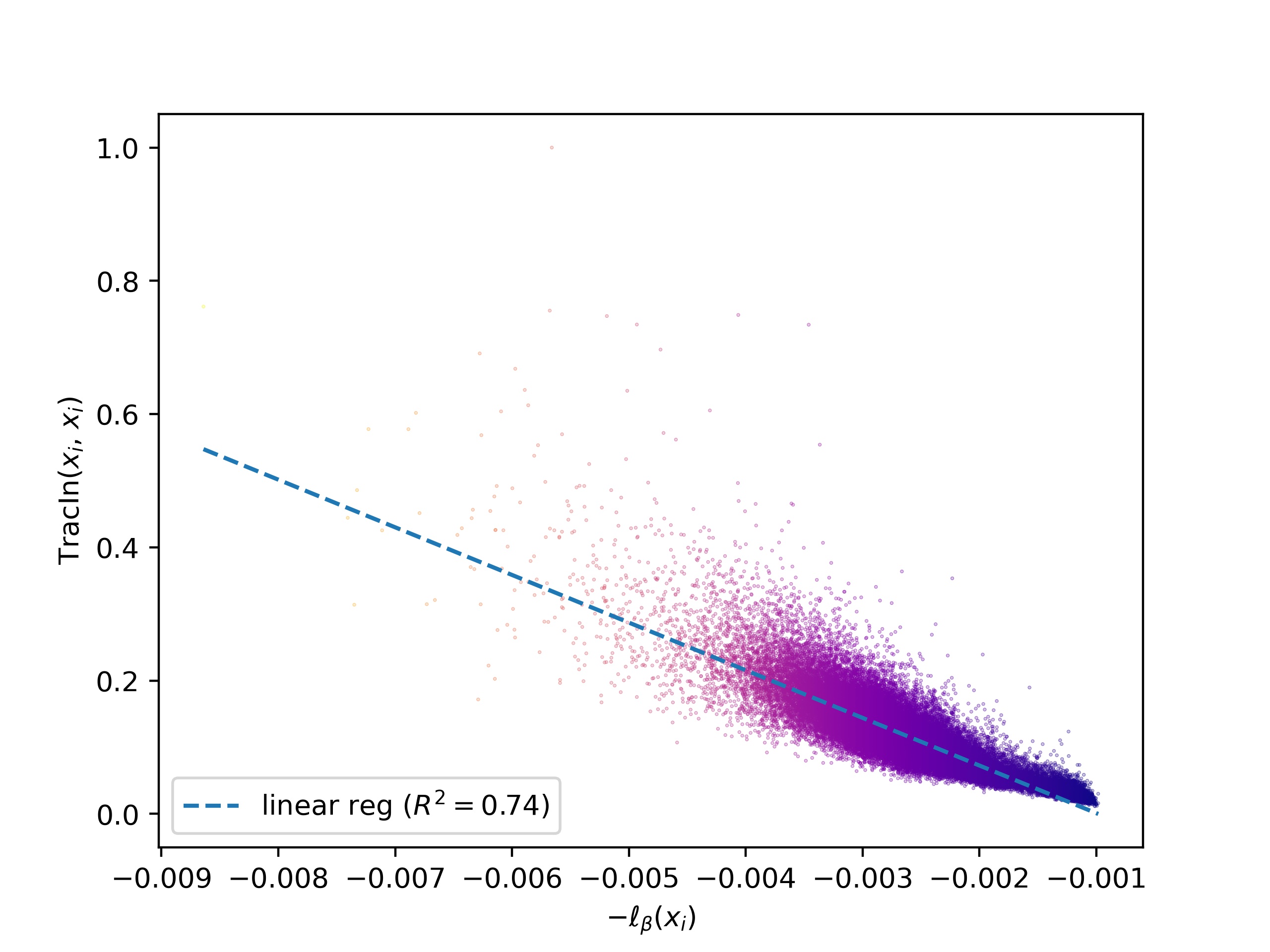}
    }
    \subfloat[][$\beta=2$, $\dlatent=128$]{ 
    	\includegraphics[trim=10 0 30 30, clip, width=0.32\textwidth]{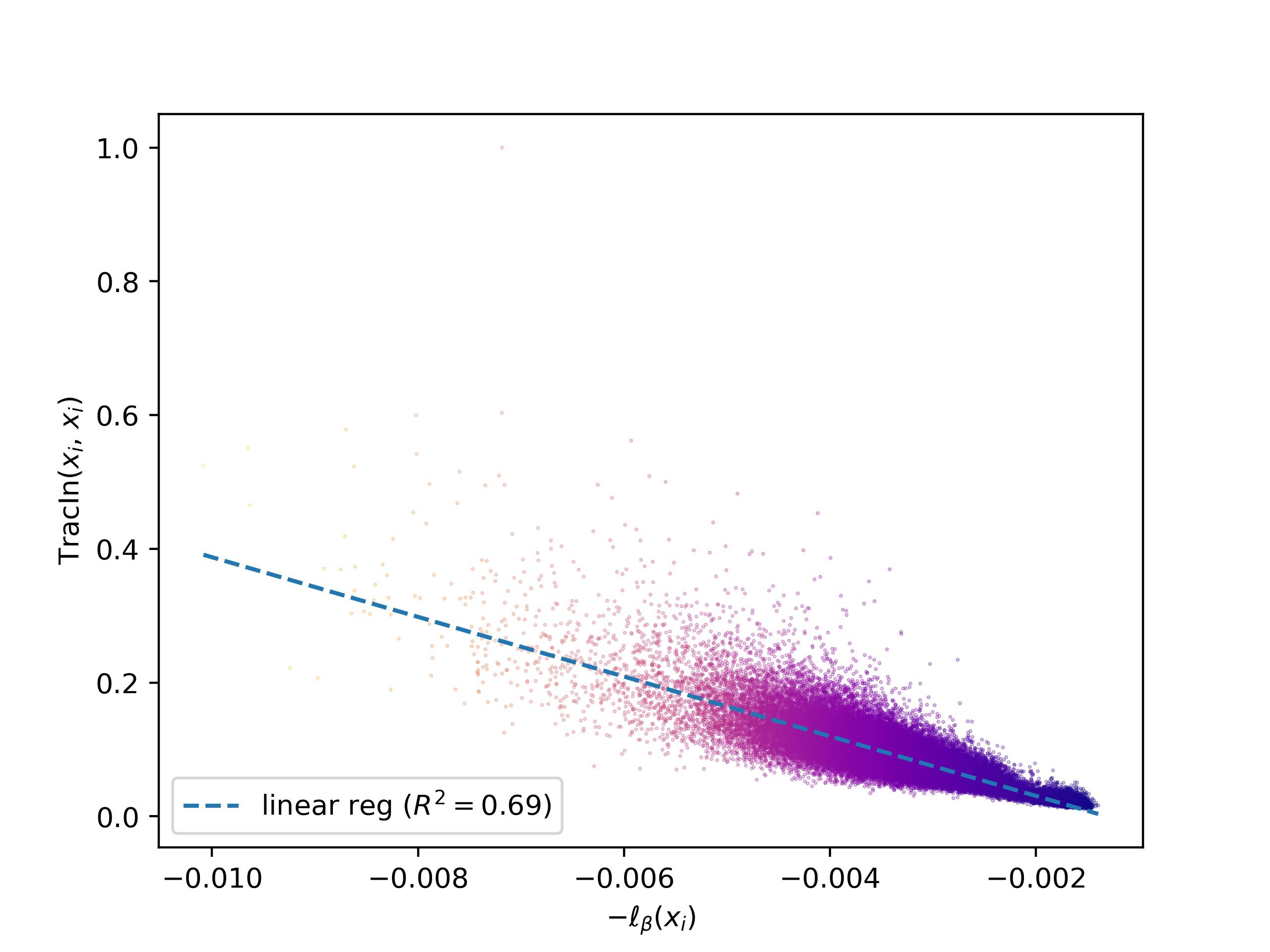}
    }
    \subfloat[][$\beta=4$, $\dlatent=128$]{ 
    	\includegraphics[trim=10 0 30 30, clip, width=0.32\textwidth]{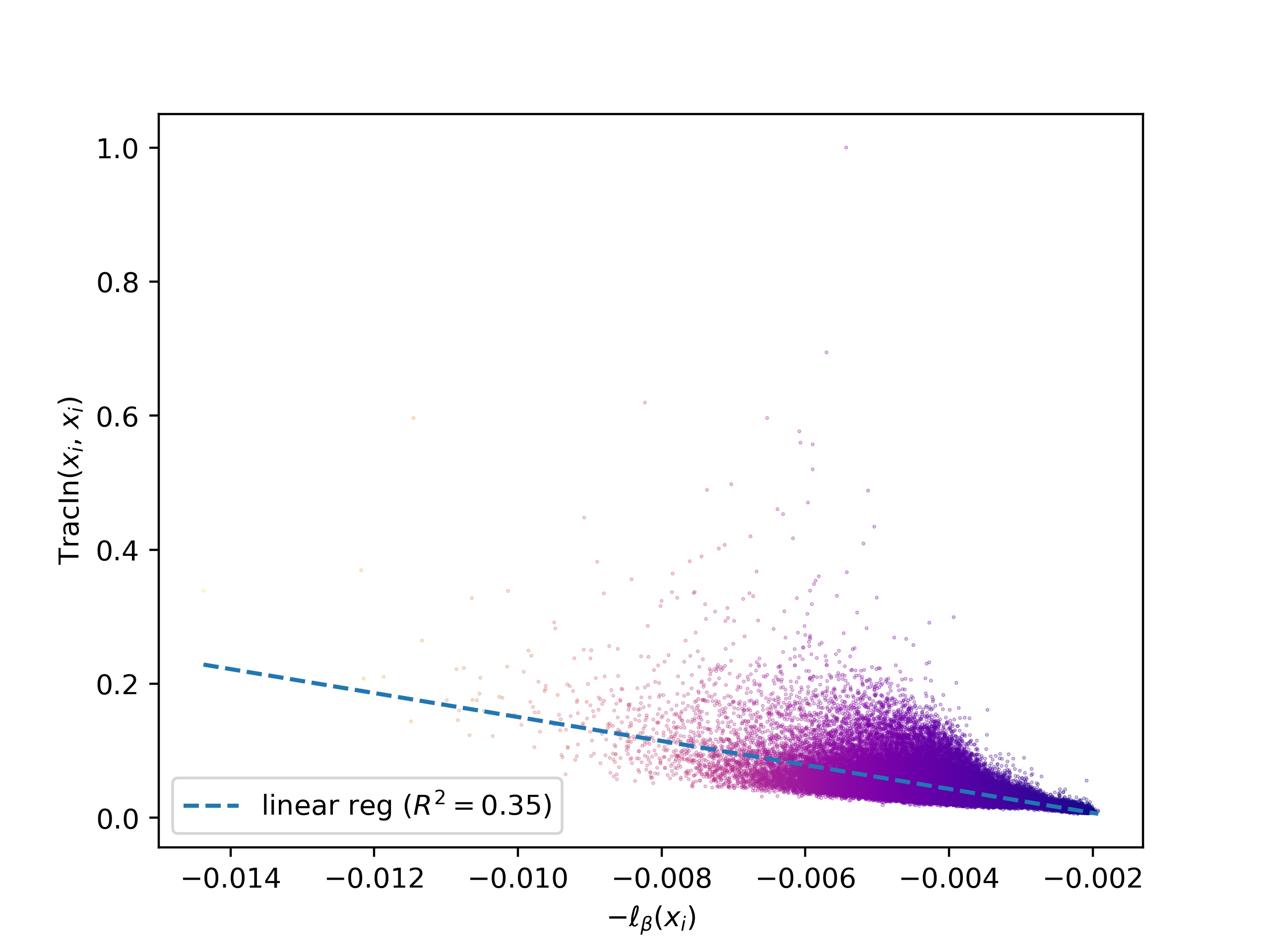}
    }\\ \vspace{-0.5em}
    \subfloat[][$\beta=1$, $\dlatent=128$]{ 
    	\includegraphics[trim=10 10 0 0, clip, width=0.32\textwidth]{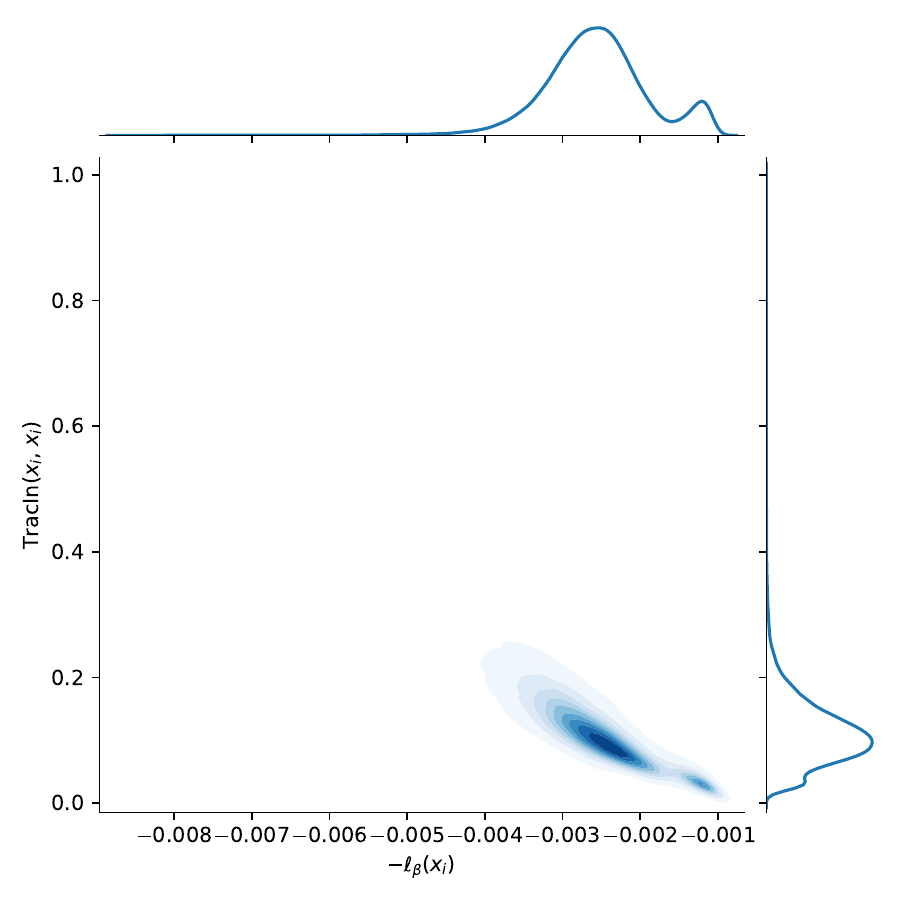} 
	}
     \subfloat[][$\beta=2$, $\dlatent=128$]{ 
    	\includegraphics[trim=10 10 0 0, clip, width=0.32\textwidth]{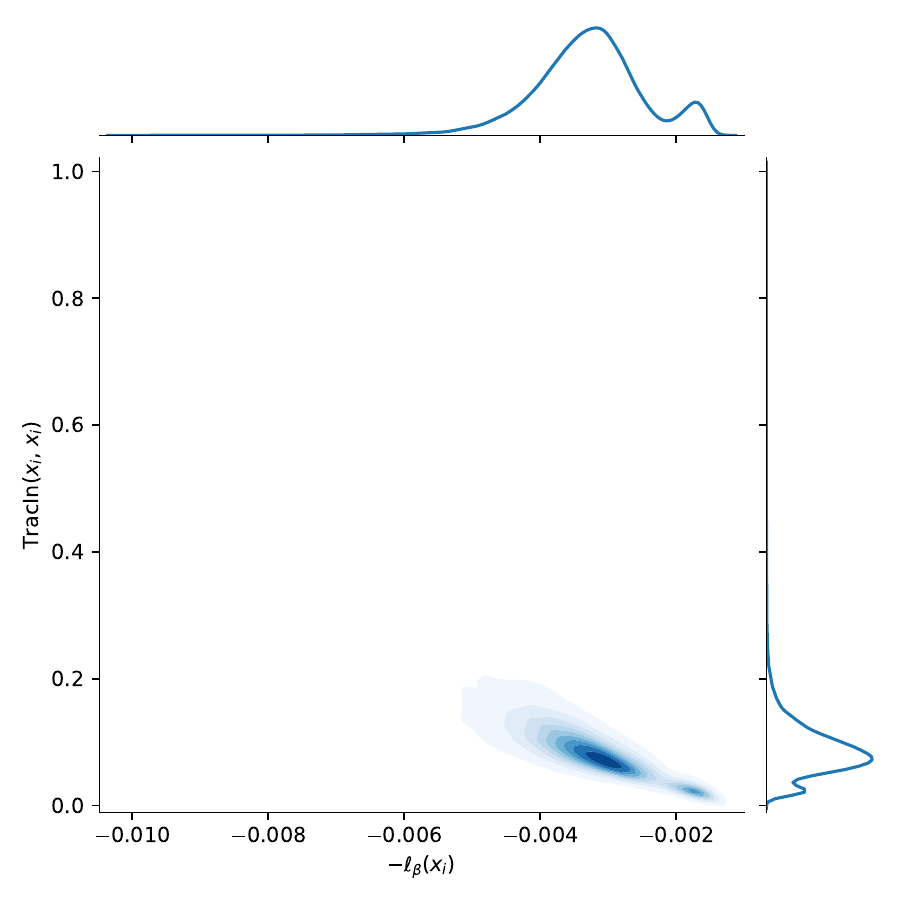} 
	}
     \subfloat[][$\beta=4$, $\dlatent=128$]{ 
    	\includegraphics[trim=10 10 0 0, clip, width=0.32\textwidth]{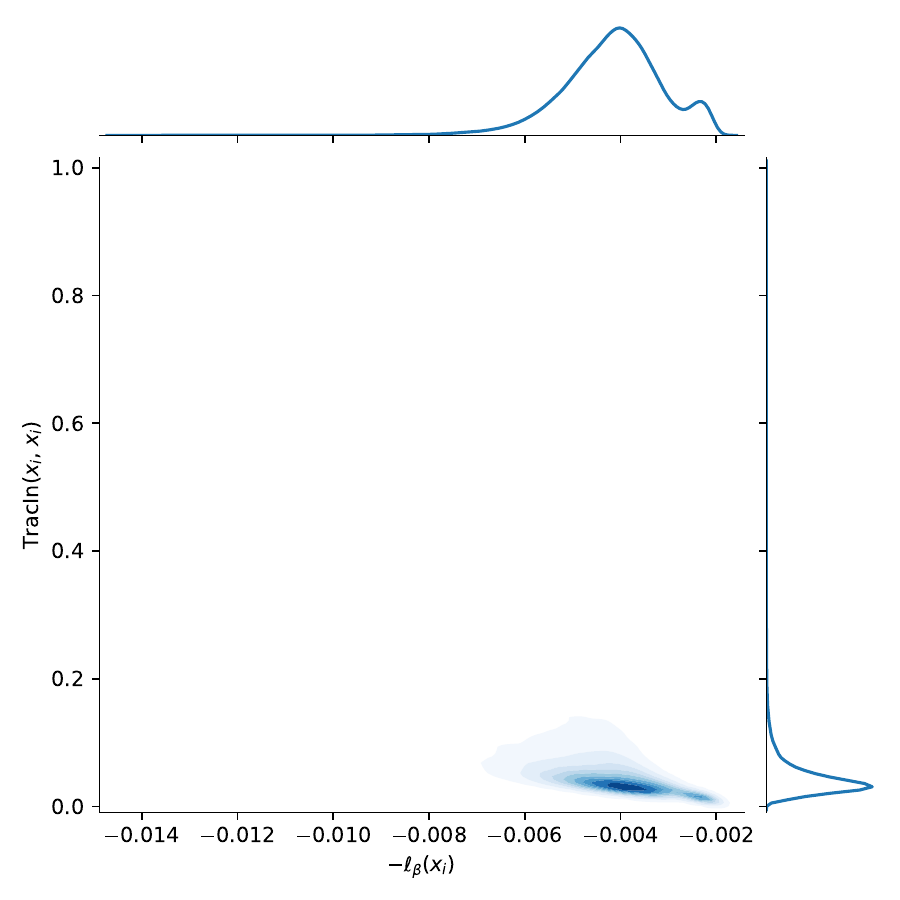}
	}\\
    \subfloat[][$\beta=1$, $\dlatent=16$]{ 
    	\includegraphics[trim=10 0 30 30, clip, width=0.32\textwidth]{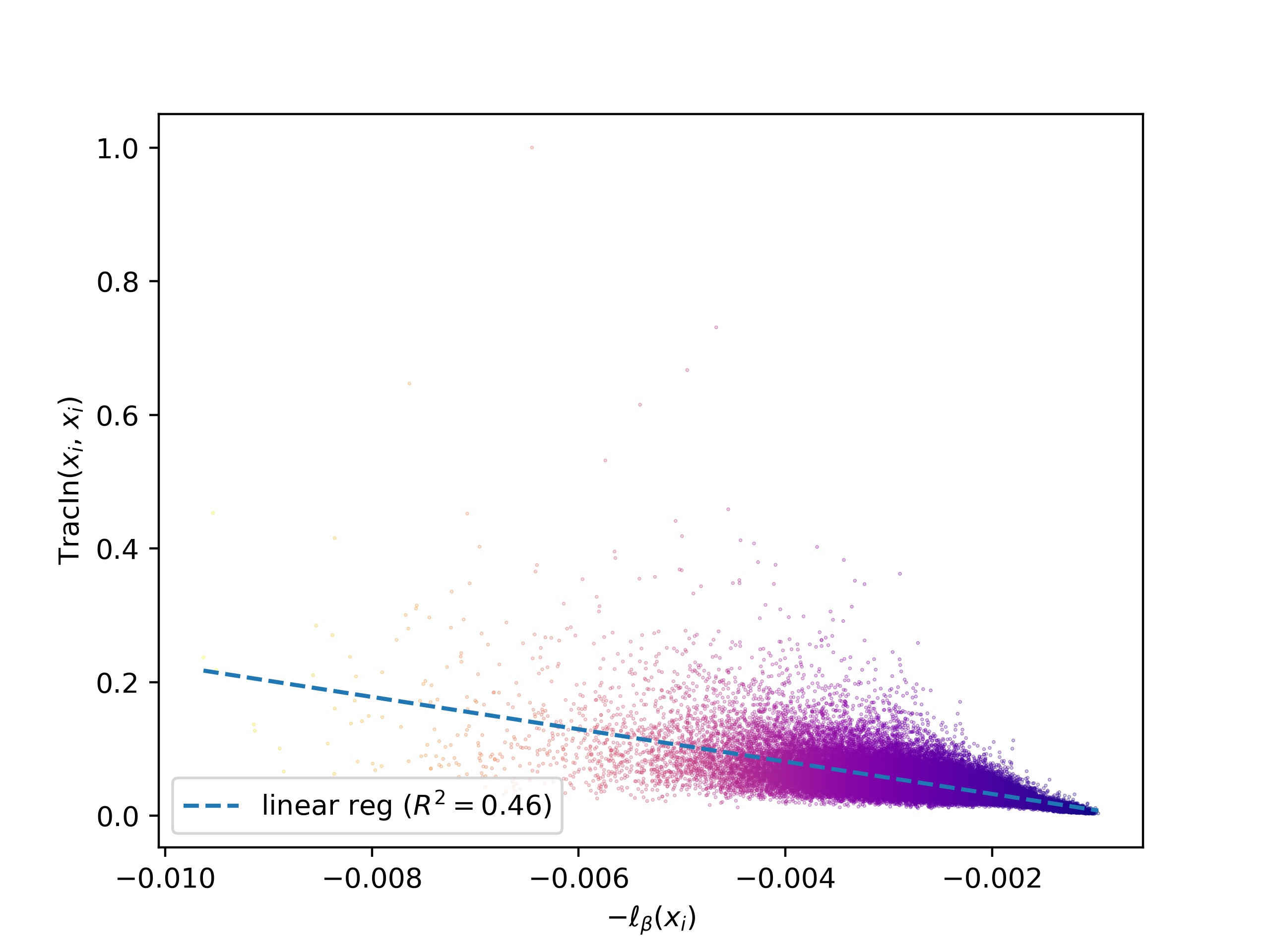}
    }
    \subfloat[][$\beta=1$, $\dlatent=64$]{ 
    	\includegraphics[trim=10 0 30 30, clip, width=0.32\textwidth]{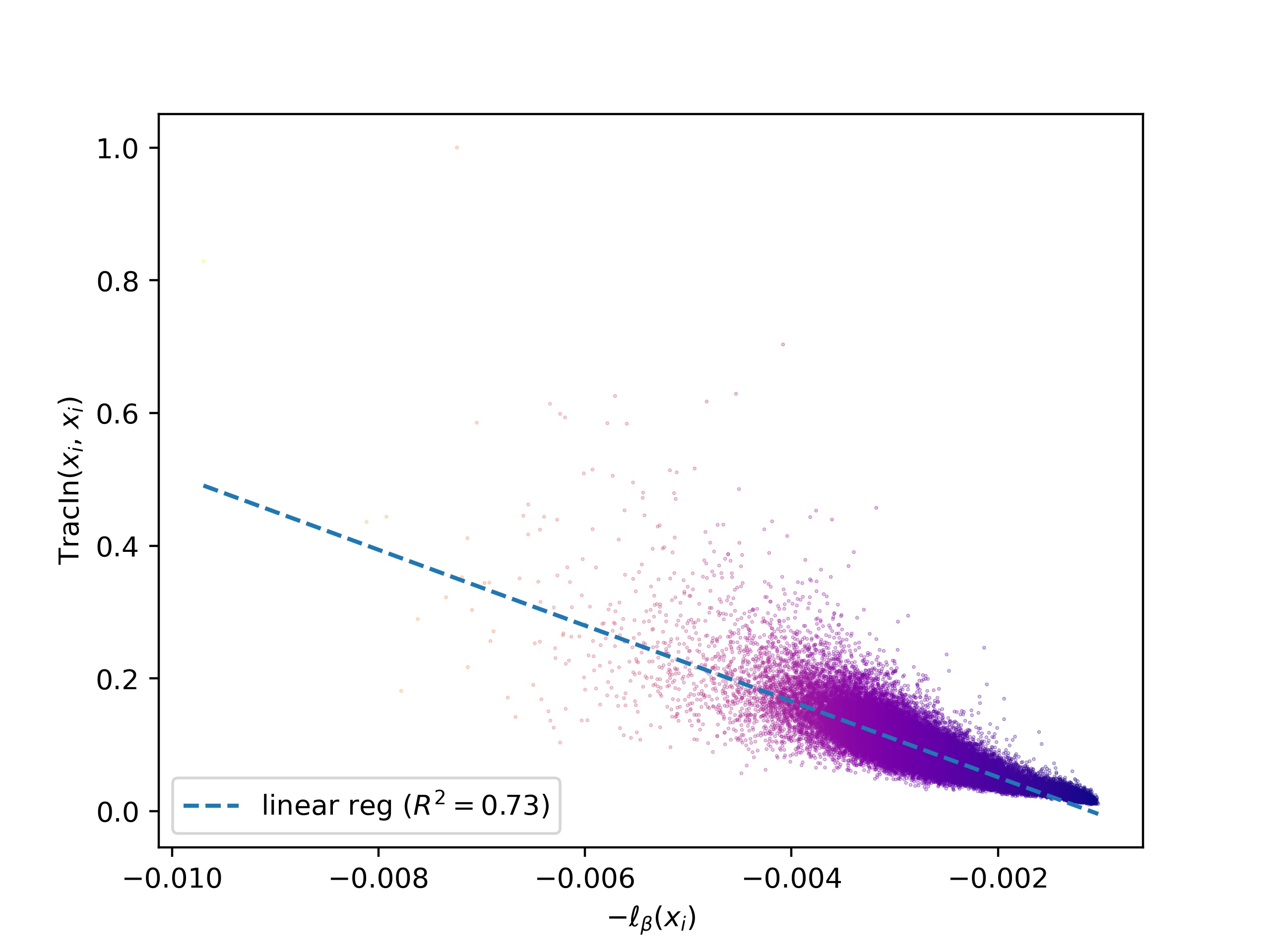}
    }
    \subfloat[][$\beta=1$, $\dlatent=128$]{ 
    	\includegraphics[trim=10 0 30 30, clip, width=0.32\textwidth]{selfinf_byELBO_mnist_emnist_outlier_beta1_z128.jpg}
    }\\ \vspace{-0.5em}
    \subfloat[][$\beta=1$, $\dlatent=16$]{ 
    	\includegraphics[trim=10 10 0 0, clip, width=0.32\textwidth]{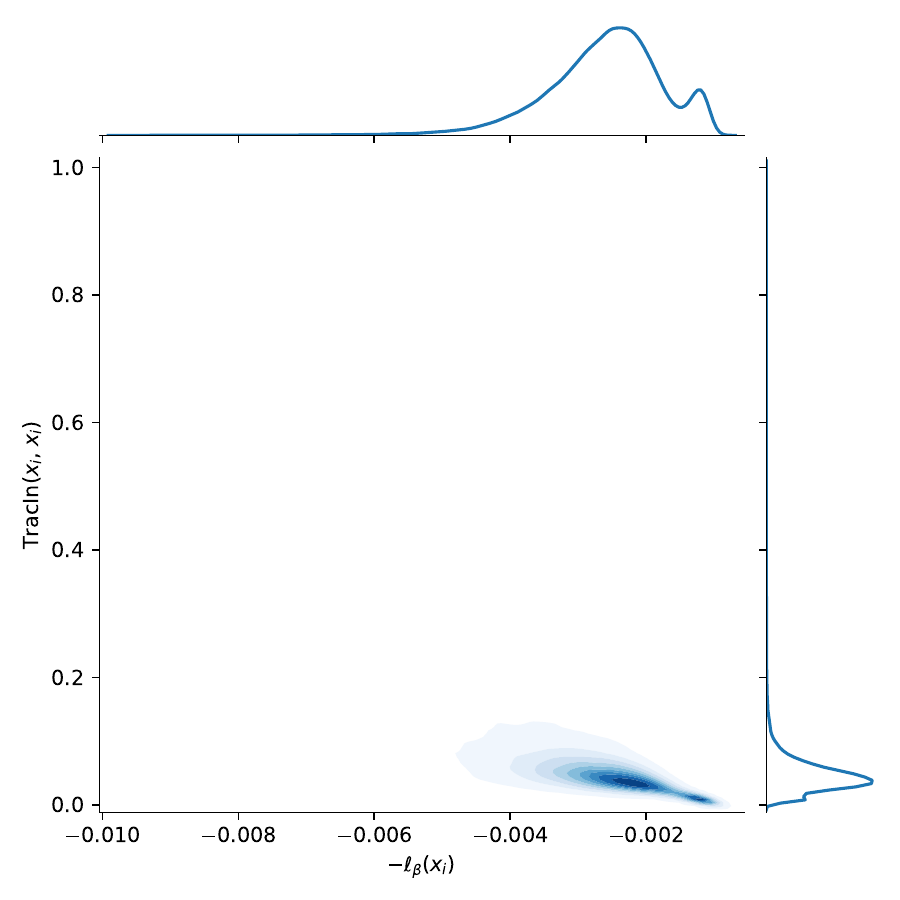} 
	}
     \subfloat[][$\beta=1$, $\dlatent=64$]{ 
    	\includegraphics[trim=10 10 0 0, clip, width=0.32\textwidth]{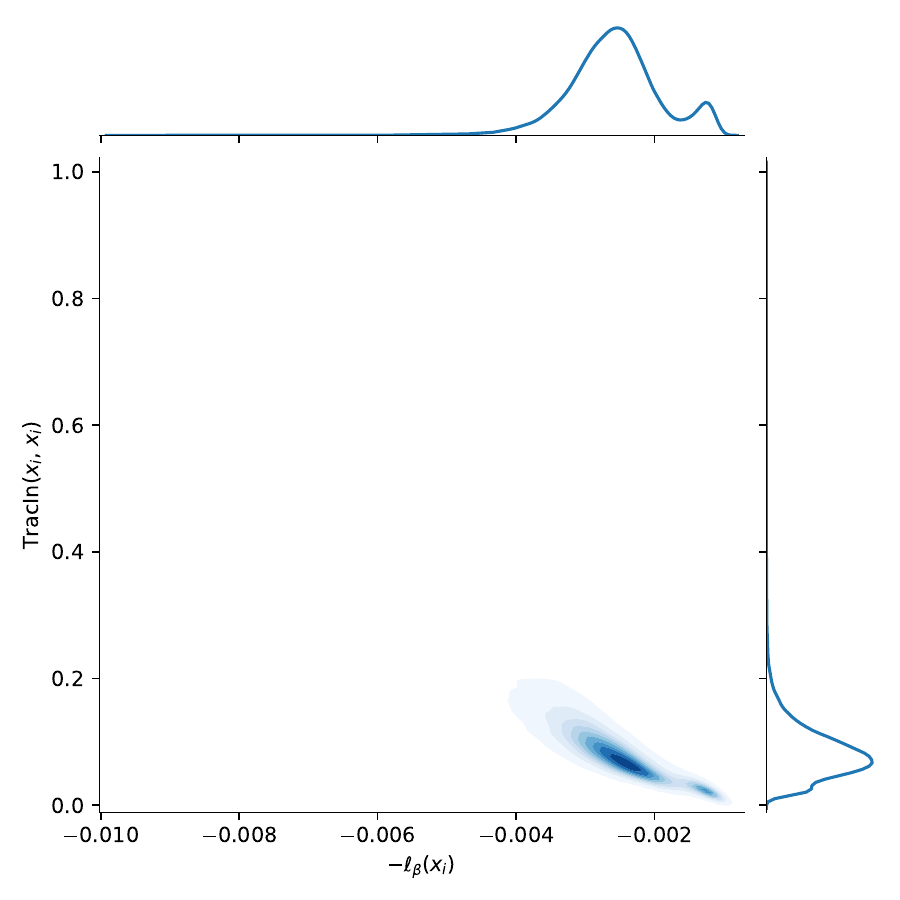} 
	}
     \subfloat[][$\beta=1$, $\dlatent=128$]{ 
    	\includegraphics[trim=10 10 0 0, clip, width=0.32\textwidth]{selfinf_kde_byELBO_mnist_emnist_outlier_beta1_z128.pdf}
	}
    \caption{Scatter and density plots of self influences versus negative losses of all training samples in MNIST. The linear regressors show that high self influence samples have large losses.}
    \label{fig: mnist self inf vs loss appendix}
\end{figure}

In MNIST experiments, we compare self influences and losses across different hyperparameters. The scatter and density plots are shown in Figure \ref{fig: mnist self inf vs loss appendix}. We fit linear regression models to these points and report $R^2$ scores. In all settings high self influence samples have large losses. We find $R^2$ is larger under high latent dimensions or smaller $\beta$.

\newpage
\subsection{Self Influences (CIFAR)}\label{appendix: self-inf cifar}

\begin{figure}[!h]
    \centering
    \subfloat[][$\dlatent=64$]{ 
    	\includegraphics[trim=10 0 30 30, clip, width=0.26\textwidth]{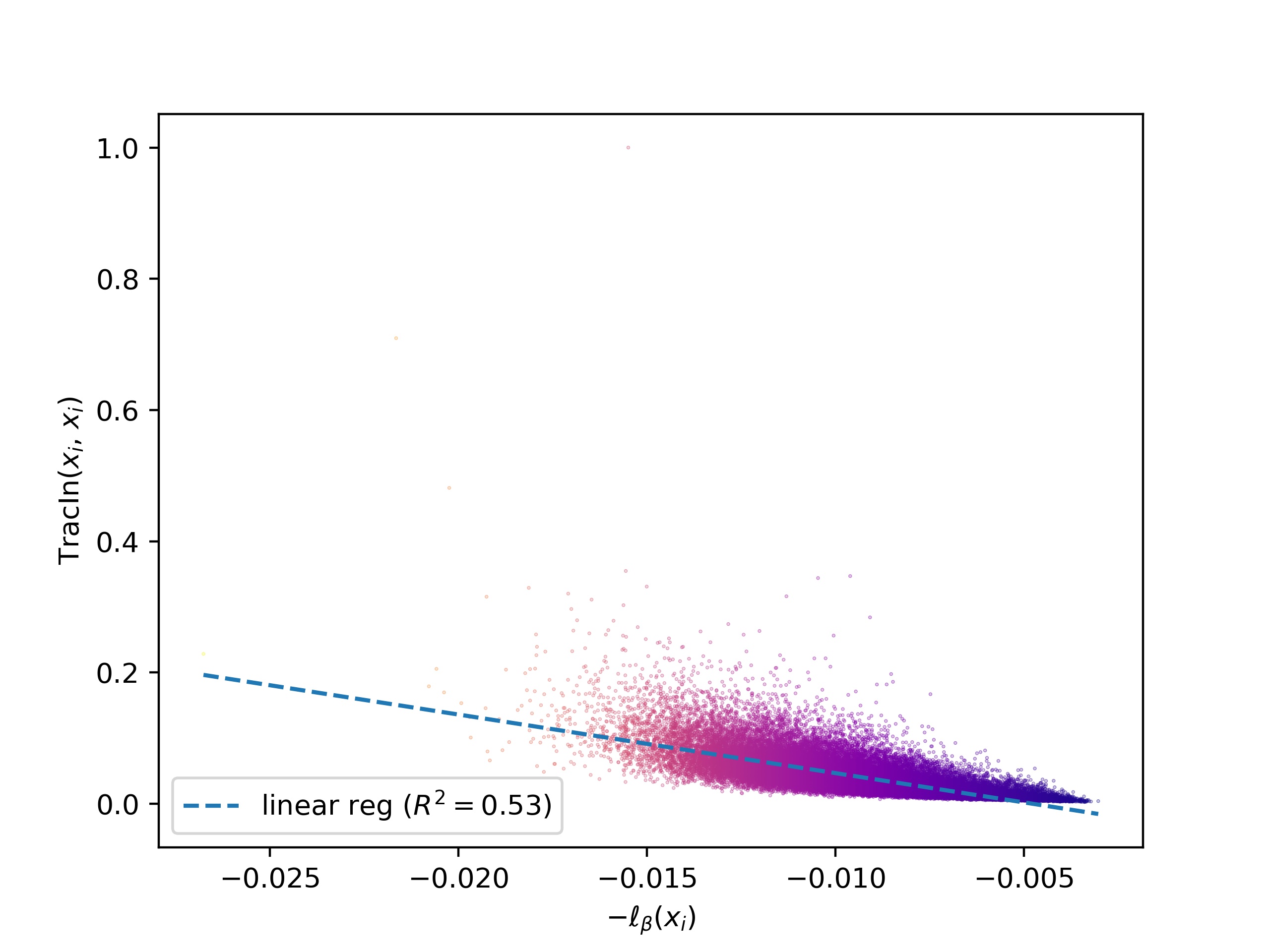}
    }
    \subfloat[][$\dlatent=128$]{ 
    	\includegraphics[trim=10 0 30 30, clip, width=0.26\textwidth]{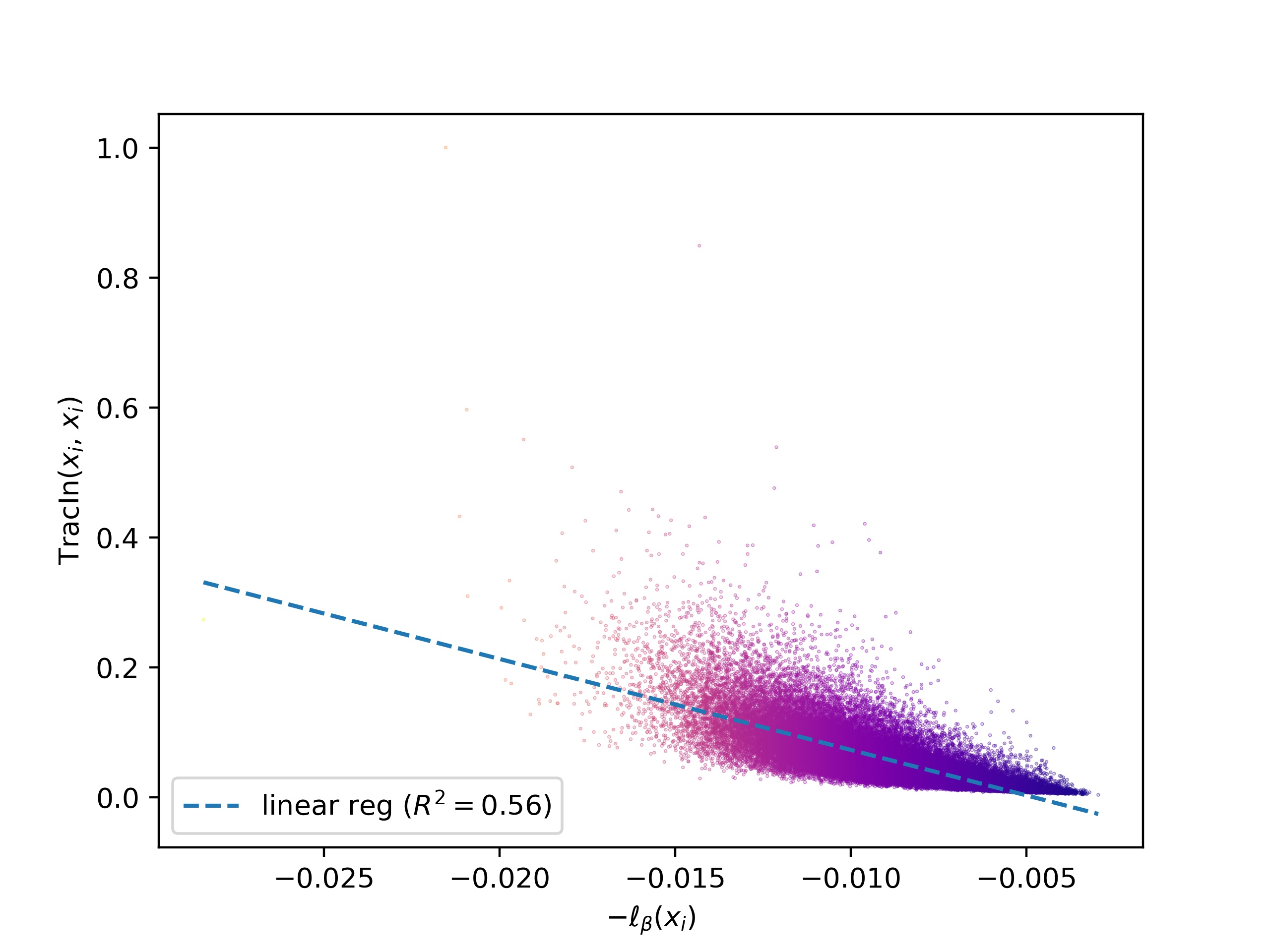}
    }
    \subfloat[][$\dlatent=64$]{ 
    	\includegraphics[trim=10 0 0 0, clip, width=0.22\textwidth]{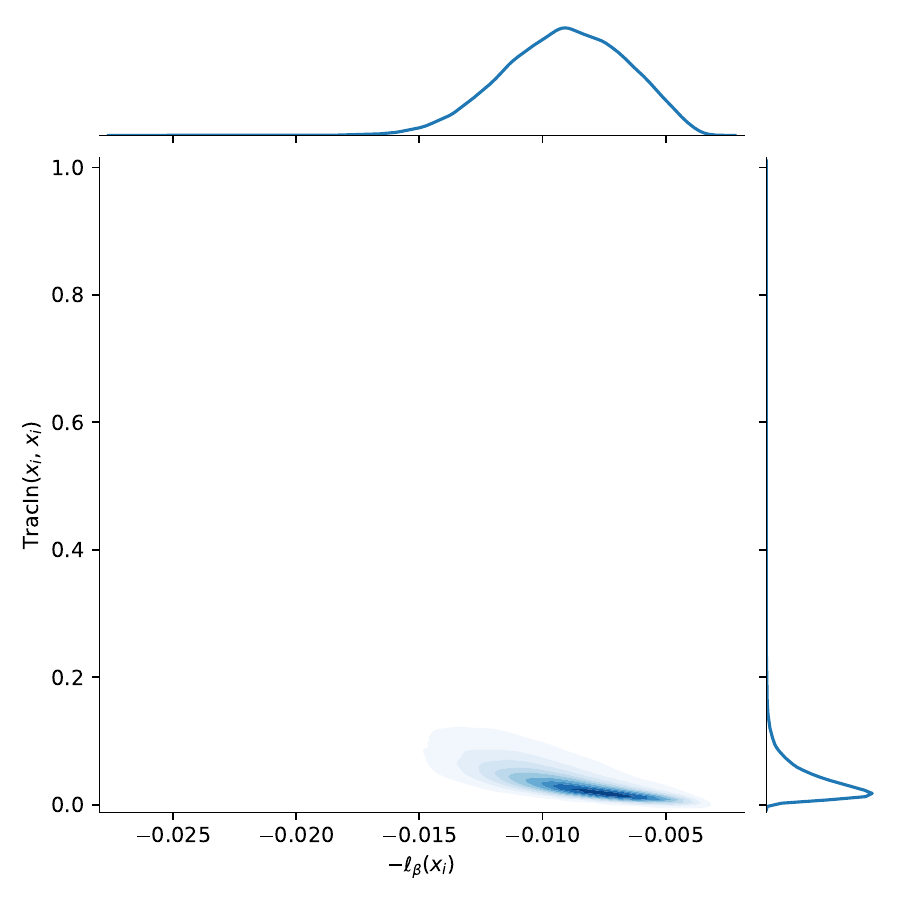} 
	}
     \subfloat[][$\dlatent=128$]{ 
    	\includegraphics[trim=10 0 0 0, clip, width=0.22\textwidth]{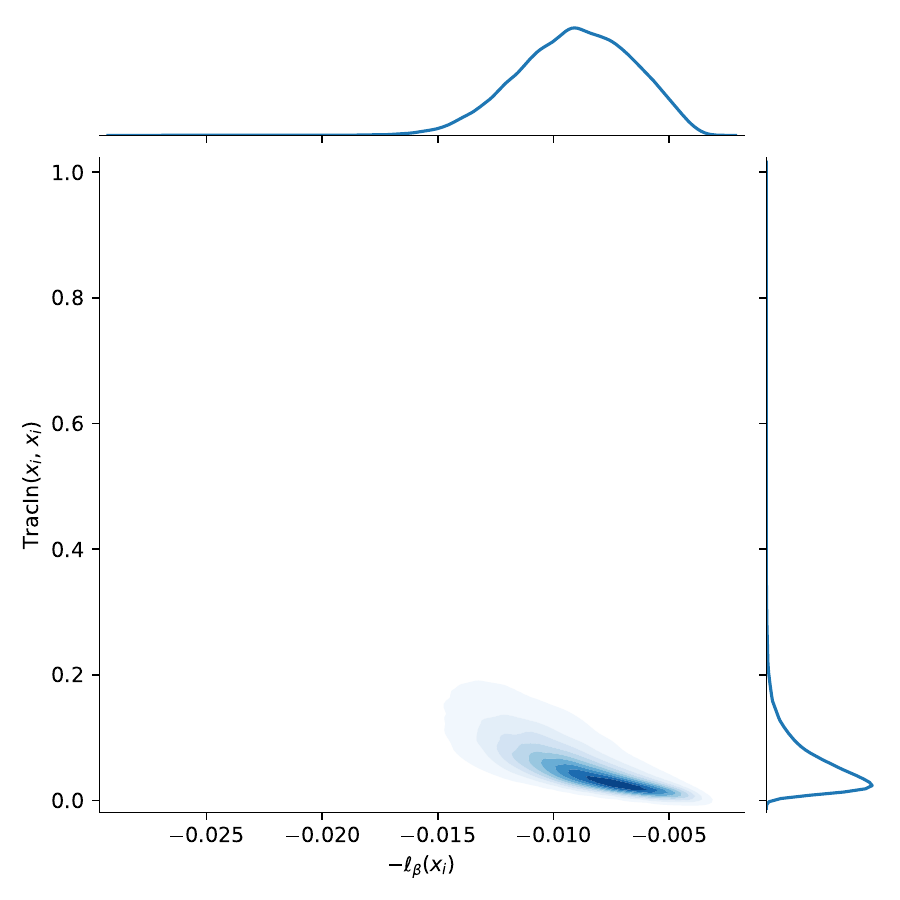} 
	}
    \caption{Scatter and density plots of self influences versus negative losses of all training samples in CIFAR. The linear regressors show that high self influence samples have large losses.}
    \label{fig: cifar self inf vs loss dlatent appendix}
\end{figure}

\begin{figure}[!h]
    \centering
    \subfloat[][CIFAR$_0$]{ 
    	\includegraphics[trim=10 0 30 30, clip, width=0.19\textwidth]{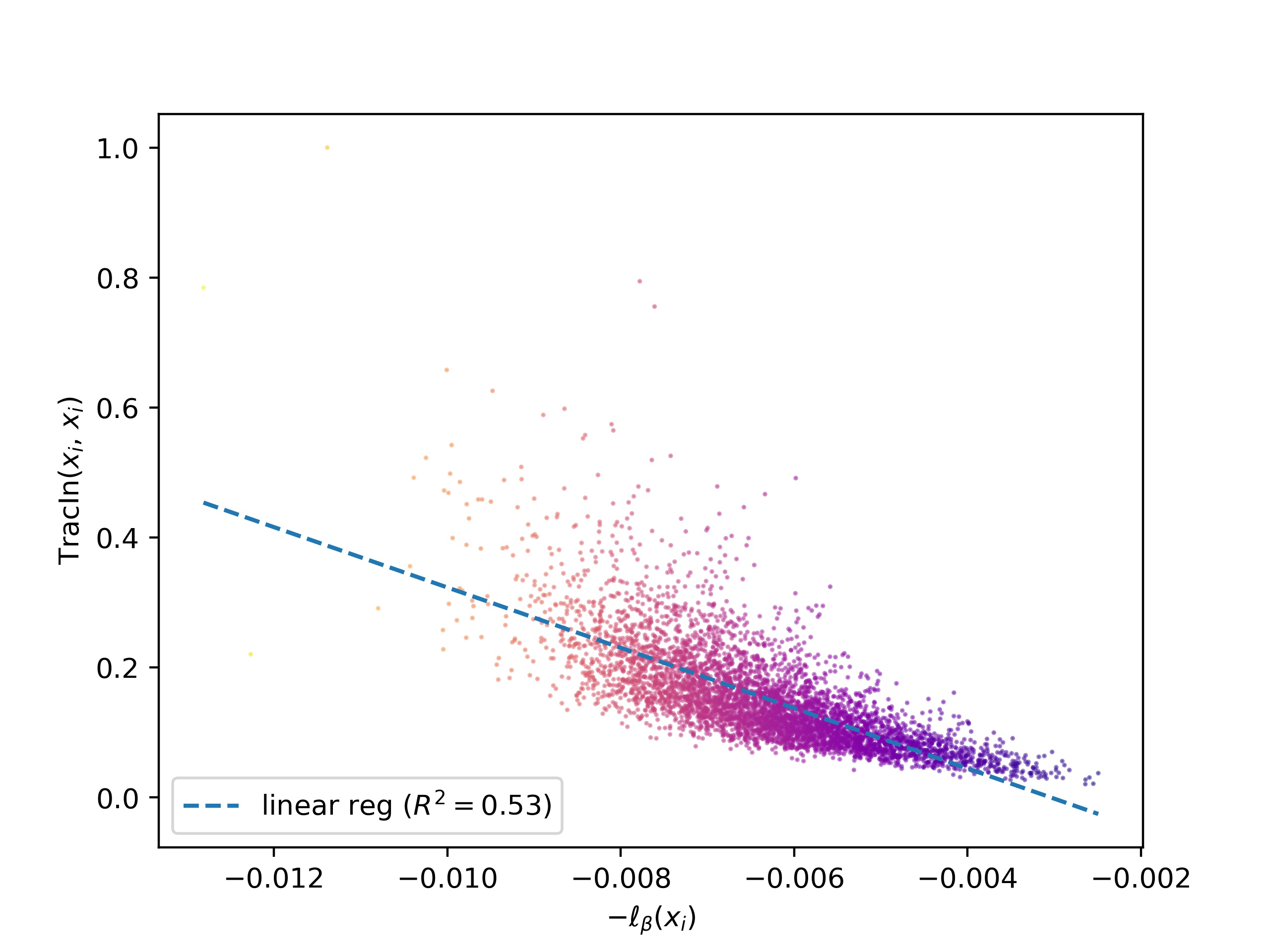}
    }
    \subfloat[][CIFAR$_1$]{ 
    	\includegraphics[trim=10 0 30 30, clip, width=0.19\textwidth]{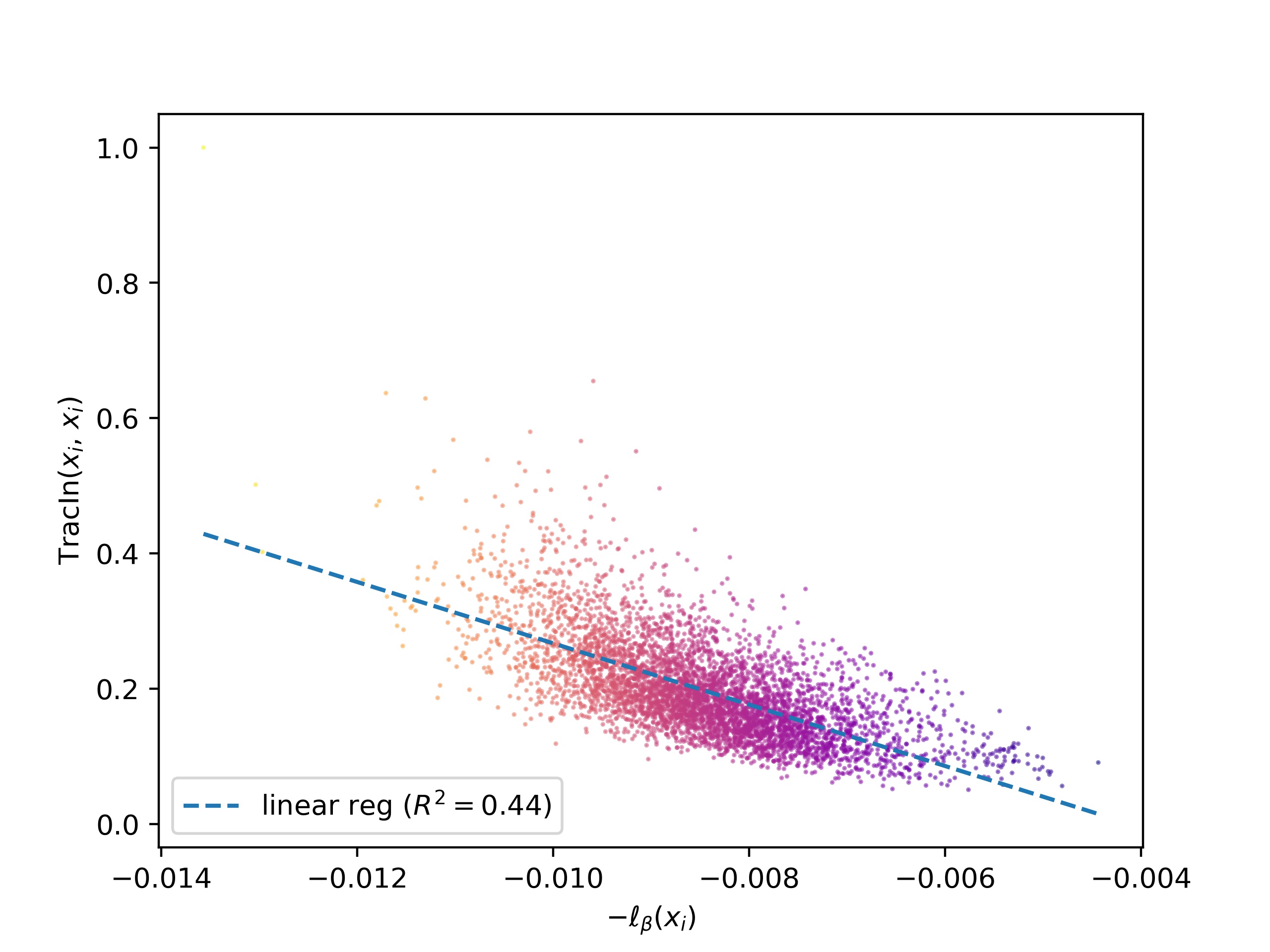}
    }
    \subfloat[][CIFAR$_2$]{ 
    	\includegraphics[trim=10 0 30 30, clip, width=0.19\textwidth]{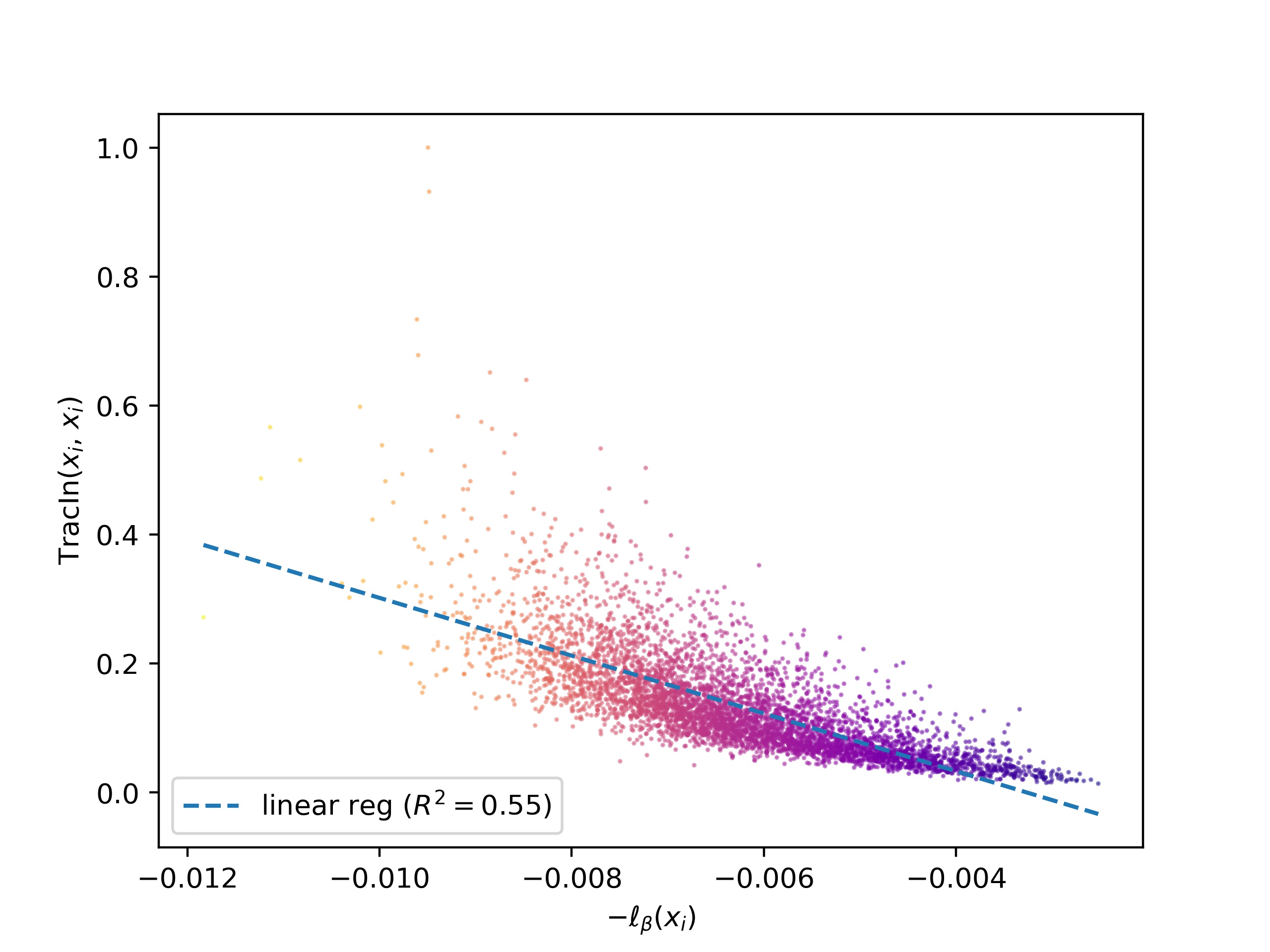}
    }
    \subfloat[][CIFAR$_3$]{ 
    	\includegraphics[trim=10 0 30 30, clip, width=0.19\textwidth]{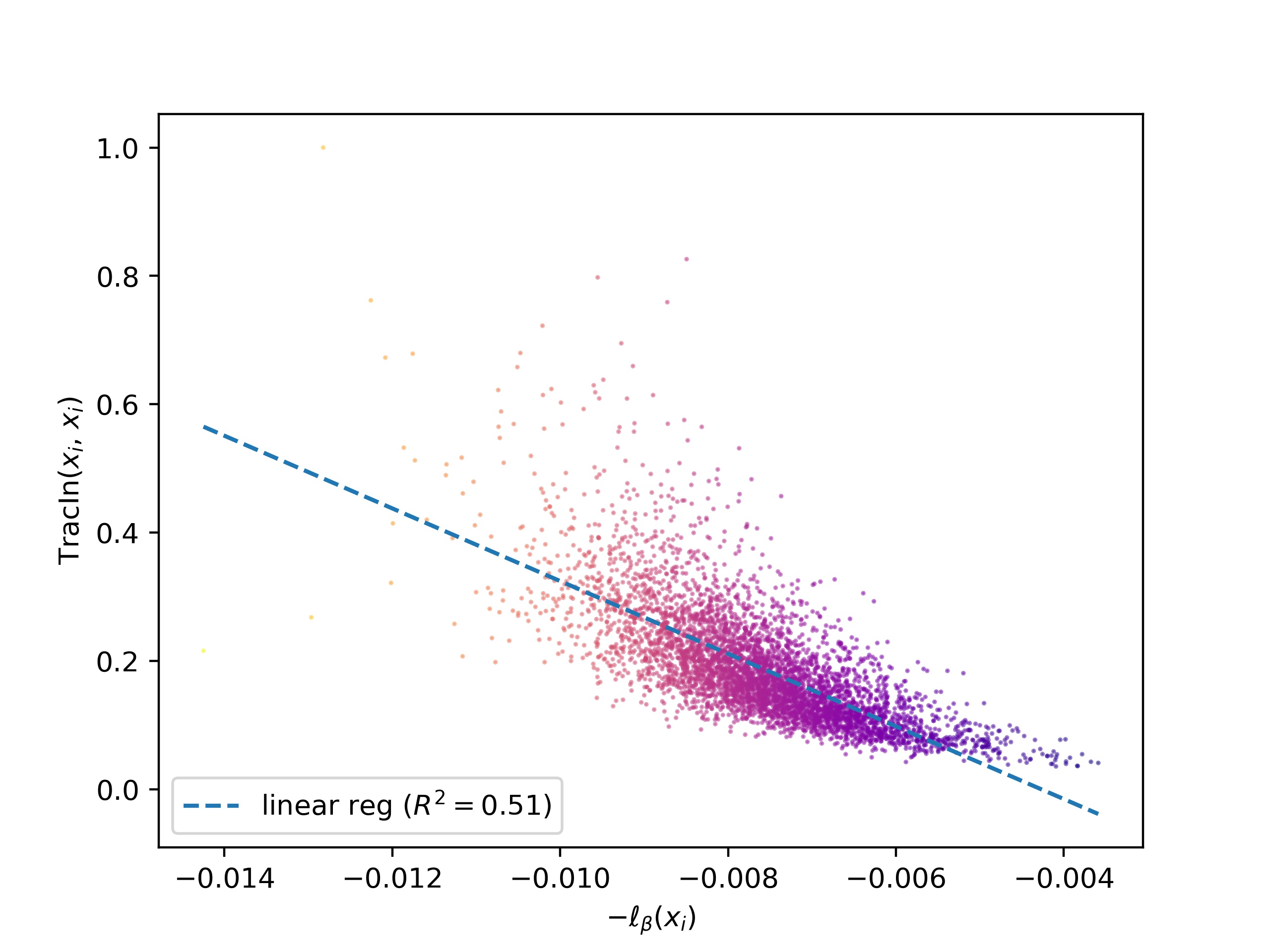}
    }
    \subfloat[][CIFAR$_4$]{ 
    	\includegraphics[trim=10 0 30 30, clip, width=0.19\textwidth]{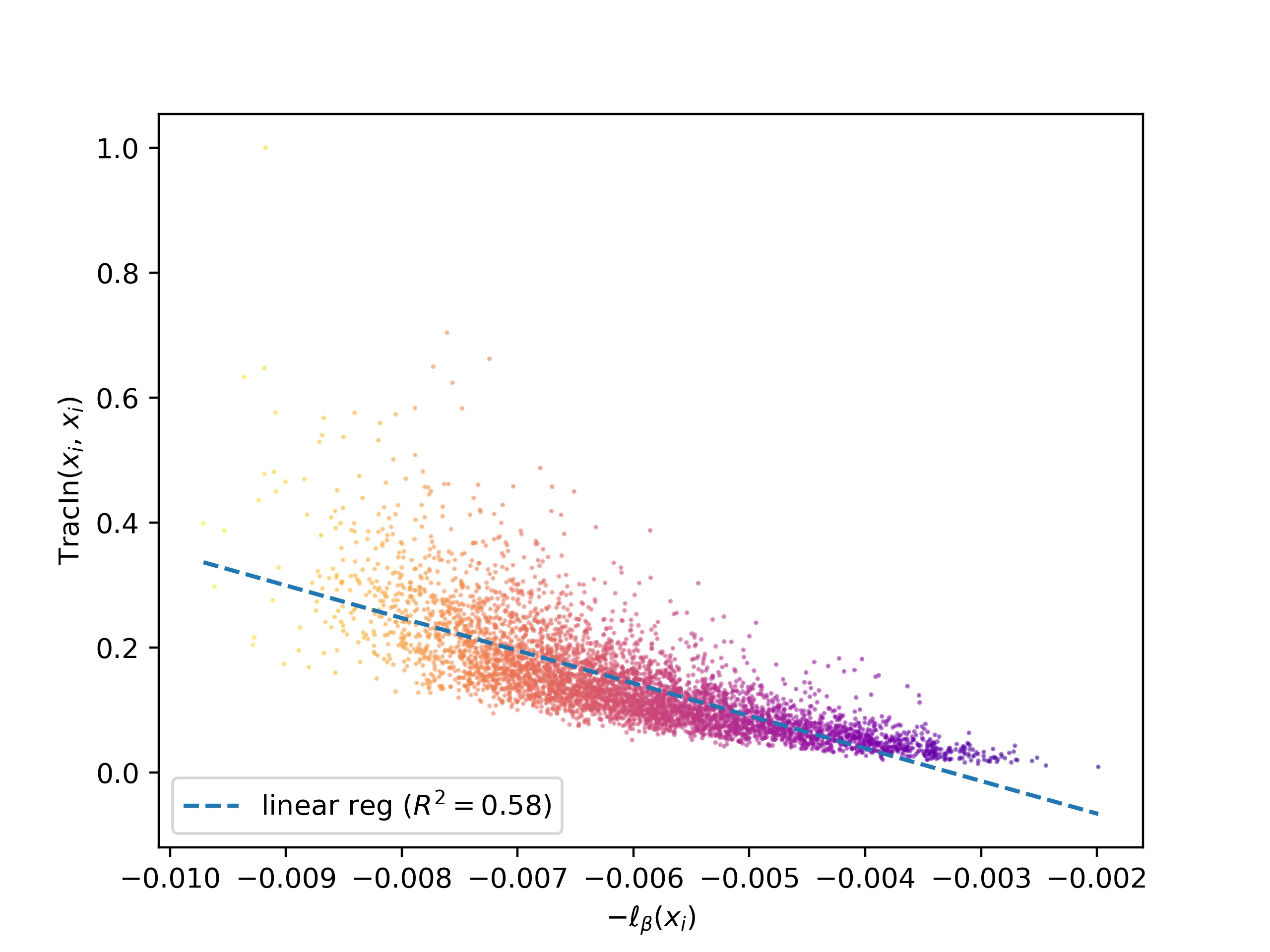}
    }\\ \vspace{-0.5em}
    \subfloat[][CIFAR$_5$]{ 
    	\includegraphics[trim=10 0 30 30, clip, width=0.19\textwidth]{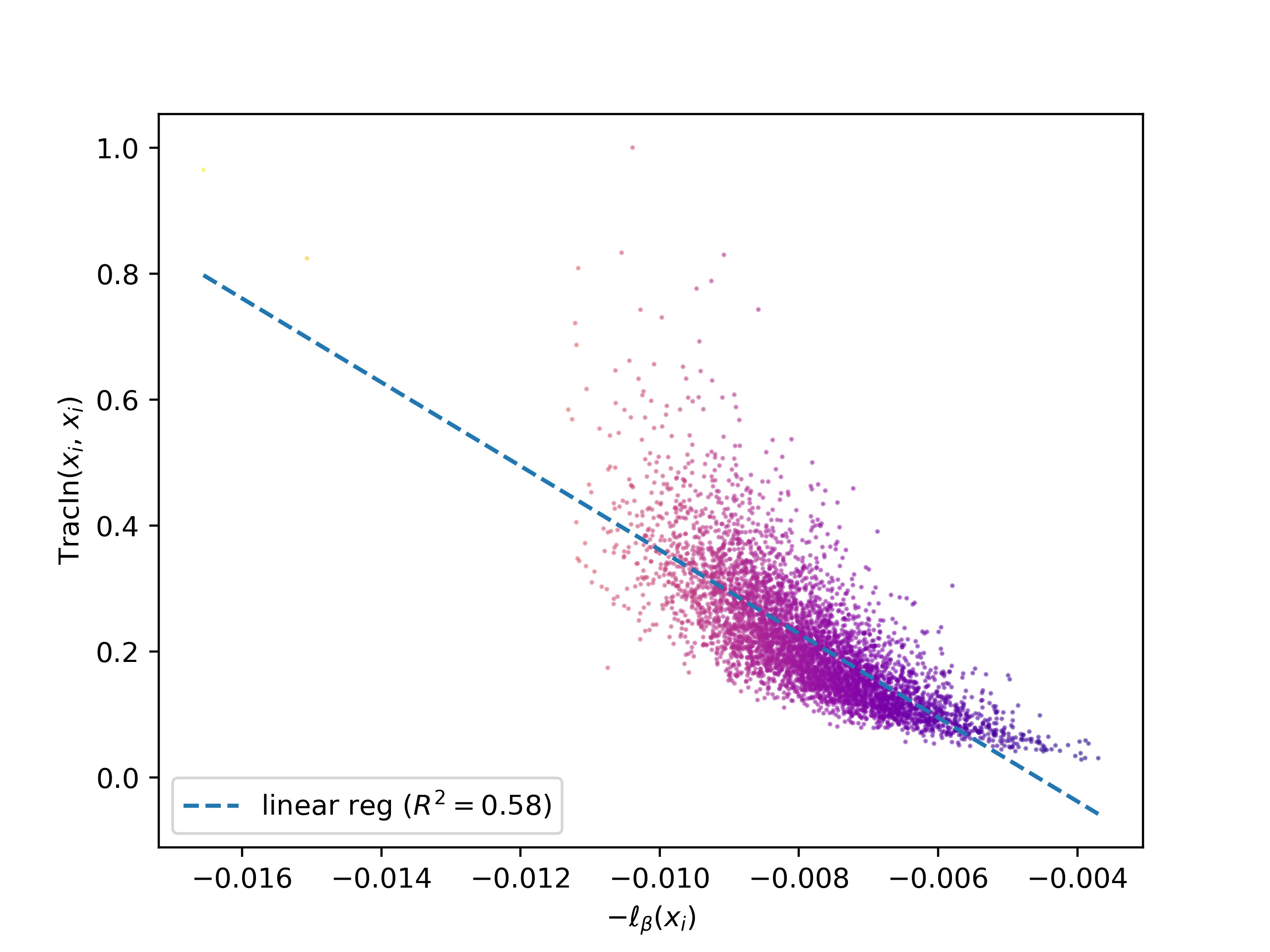}
    }
    \subfloat[][CIFAR$_6$]{ 
    	\includegraphics[trim=10 0 30 30, clip, width=0.19\textwidth]{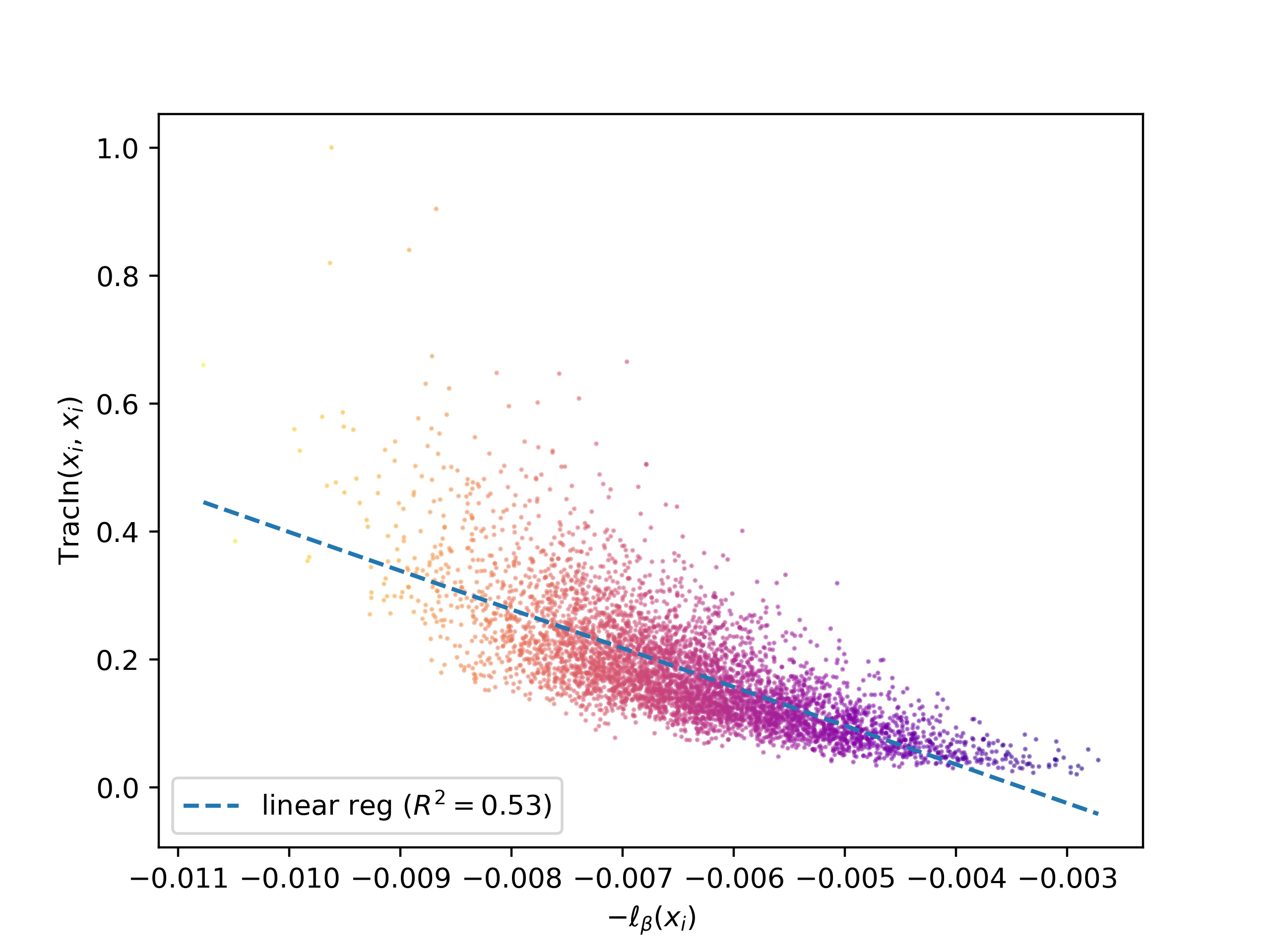}
    }
    \subfloat[][CIFAR$_7$]{ 
    	\includegraphics[trim=10 0 30 30, clip, width=0.19\textwidth]{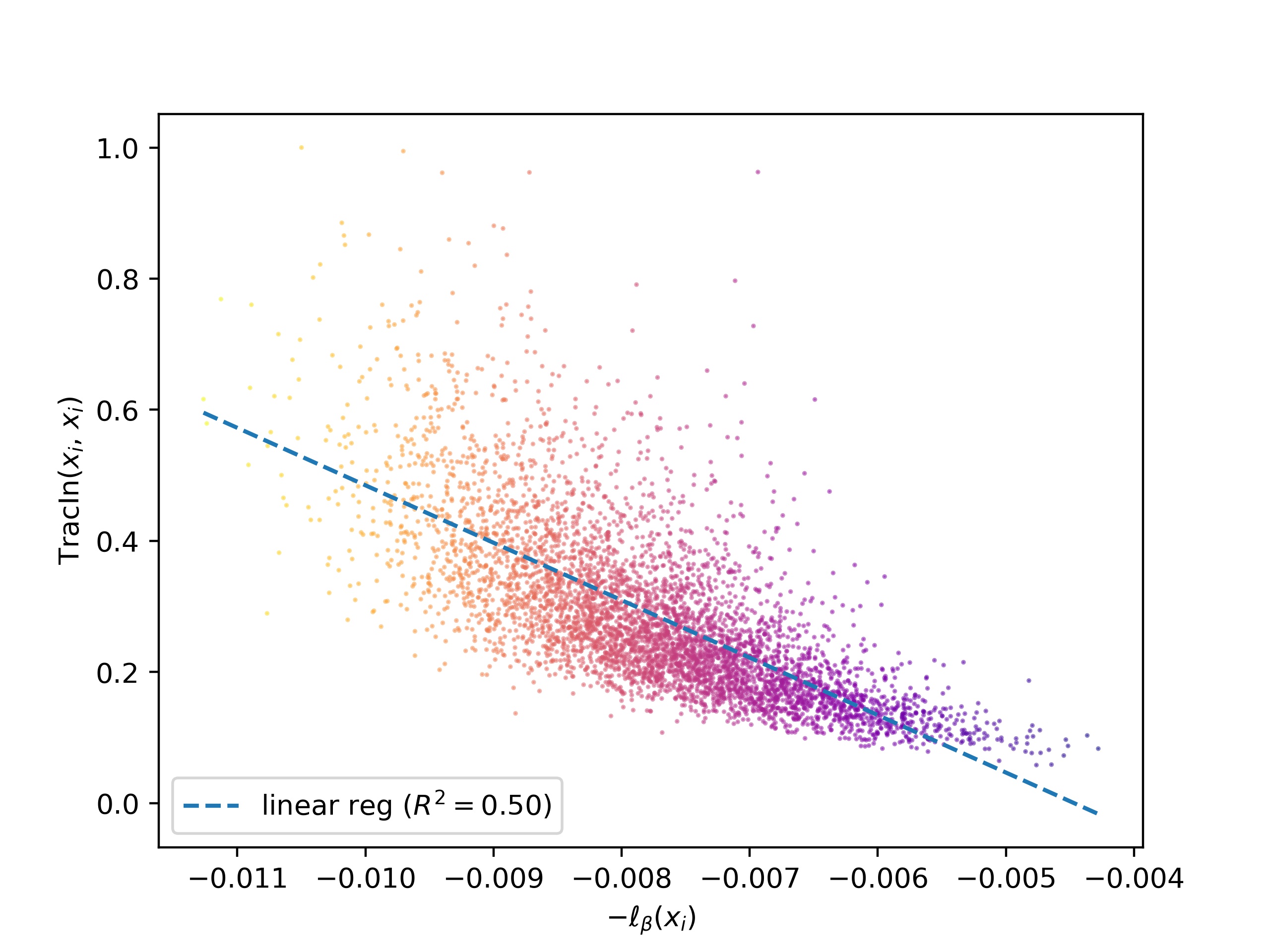}
    }
    \subfloat[][CIFAR$_8$]{ 
    	\includegraphics[trim=10 0 30 30, clip, width=0.19\textwidth]{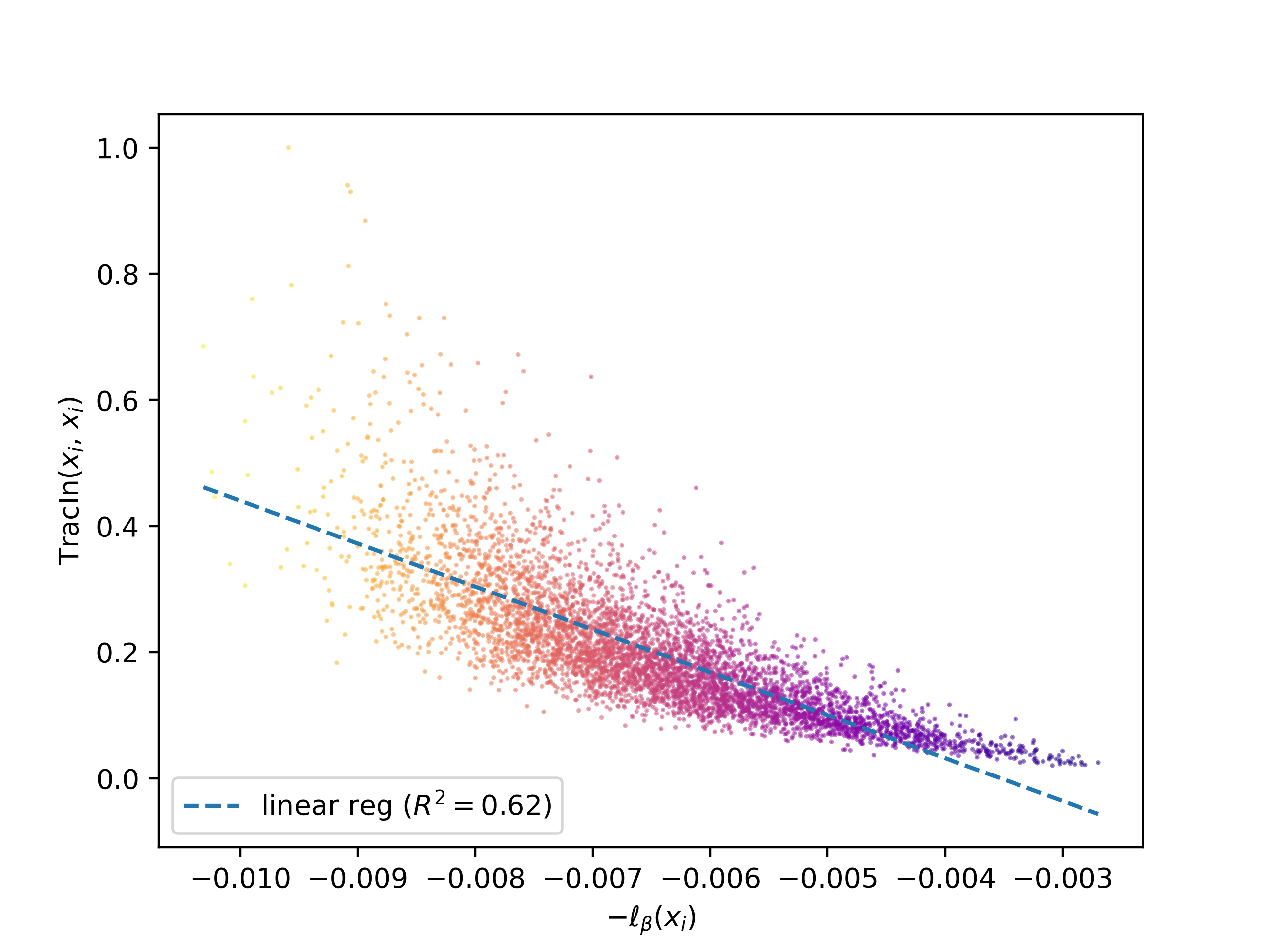}
    }
    \subfloat[][CIFAR$_9$]{ 
    	\includegraphics[trim=10 0 30 30, clip, width=0.19\textwidth]{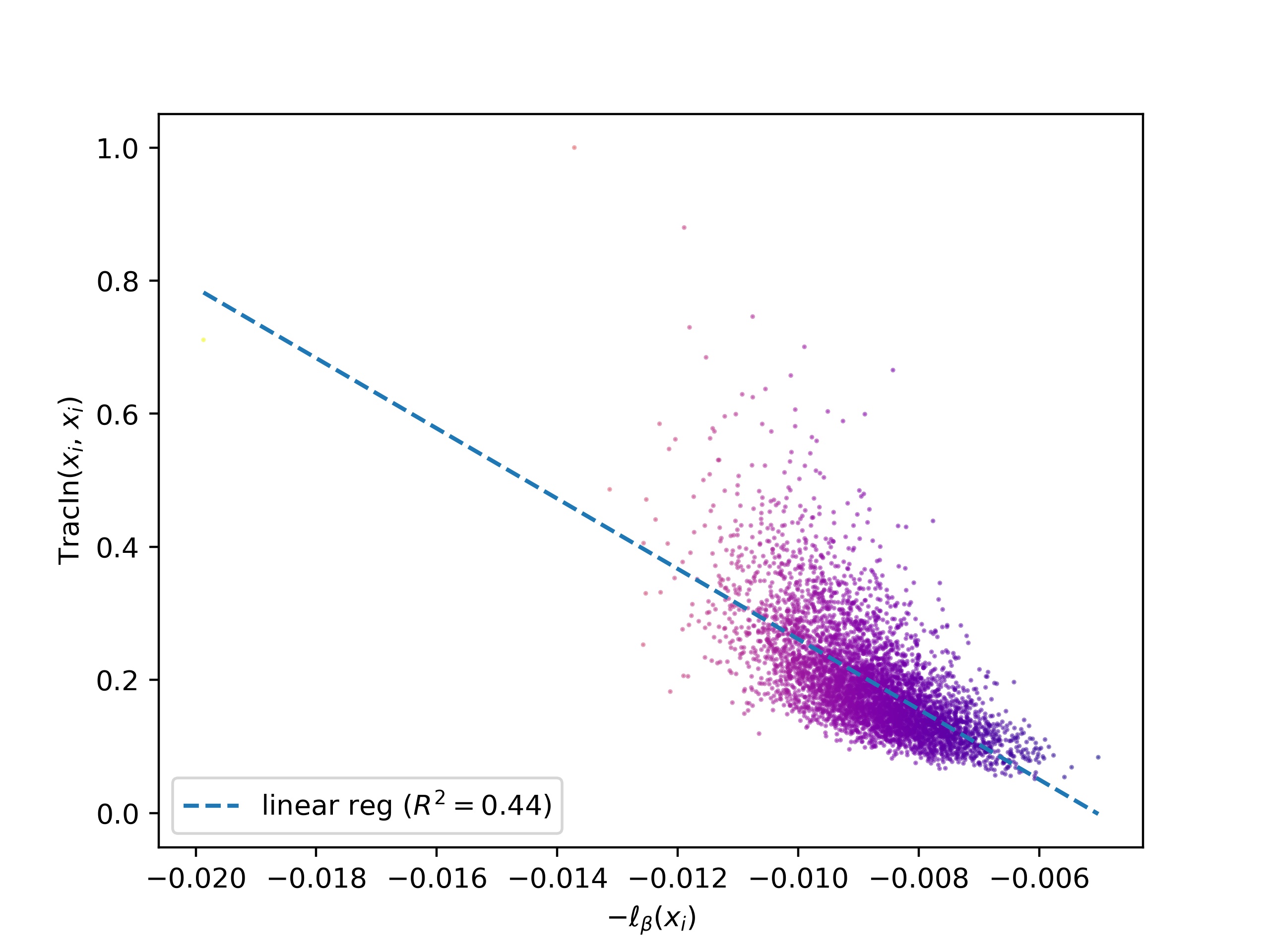}
    }\\
    \subfloat[][CIFAR$_0$]{ 
    	\includegraphics[trim=10 10 0 0, clip, width=0.19\textwidth]{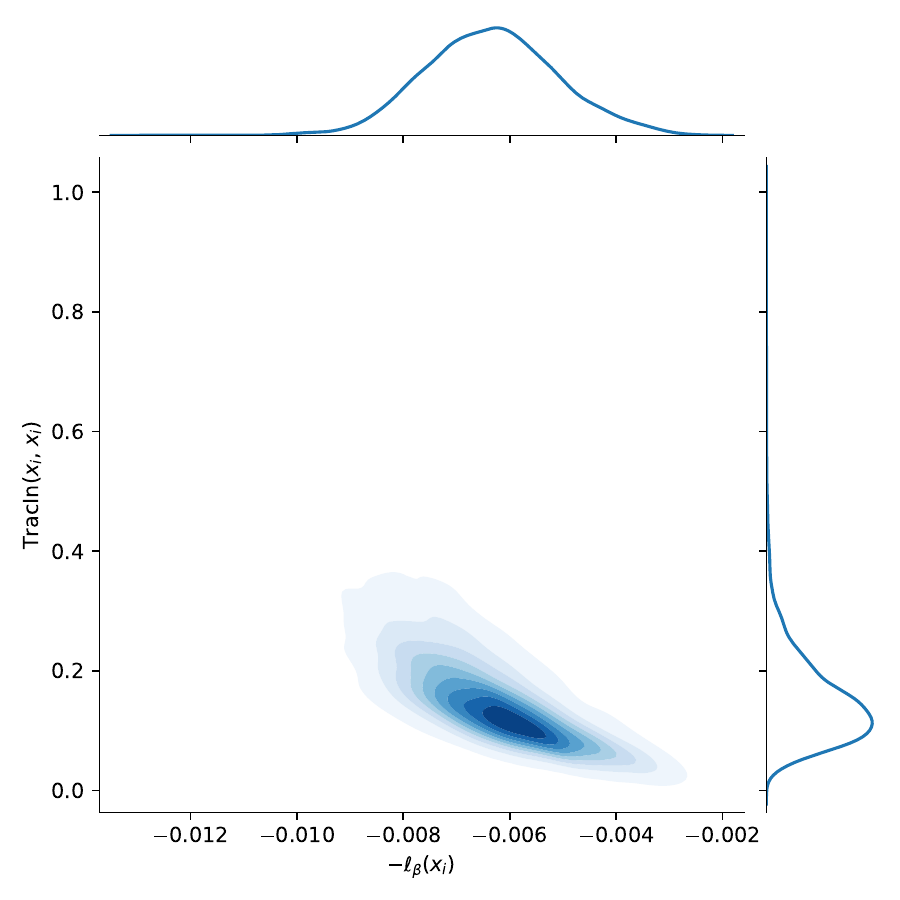}
    }
    \subfloat[][CIFAR$_1$]{ 
    	\includegraphics[trim=10 10 0 0, clip, width=0.19\textwidth]{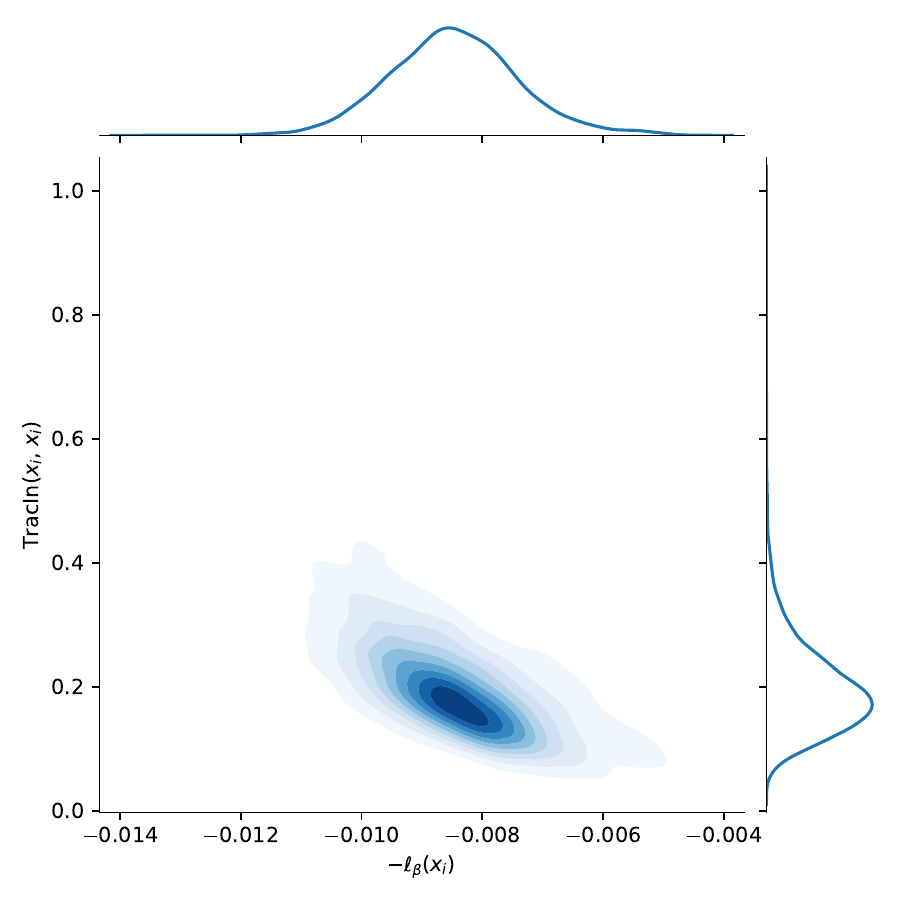}
    }
    \subfloat[][CIFAR$_2$]{ 
    	\includegraphics[trim=10 10 0 0, clip, width=0.19\textwidth]{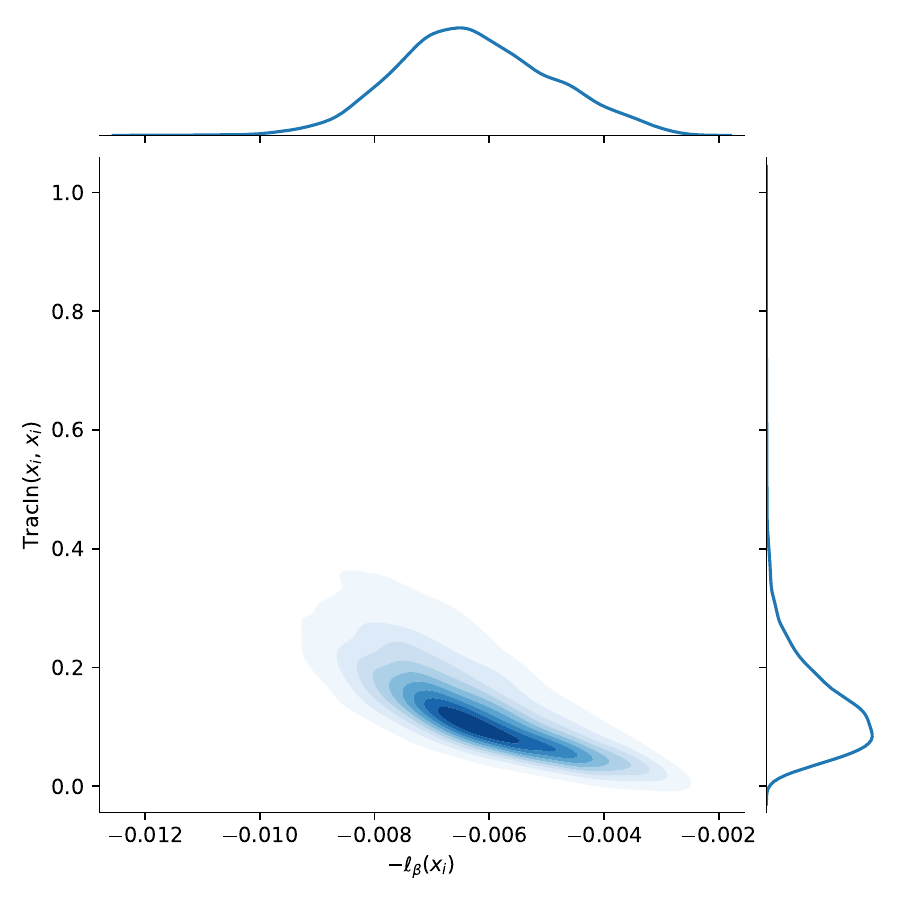}
    }
    \subfloat[][CIFAR$_3$]{ 
    	\includegraphics[trim=10 10 0 0, clip, width=0.19\textwidth]{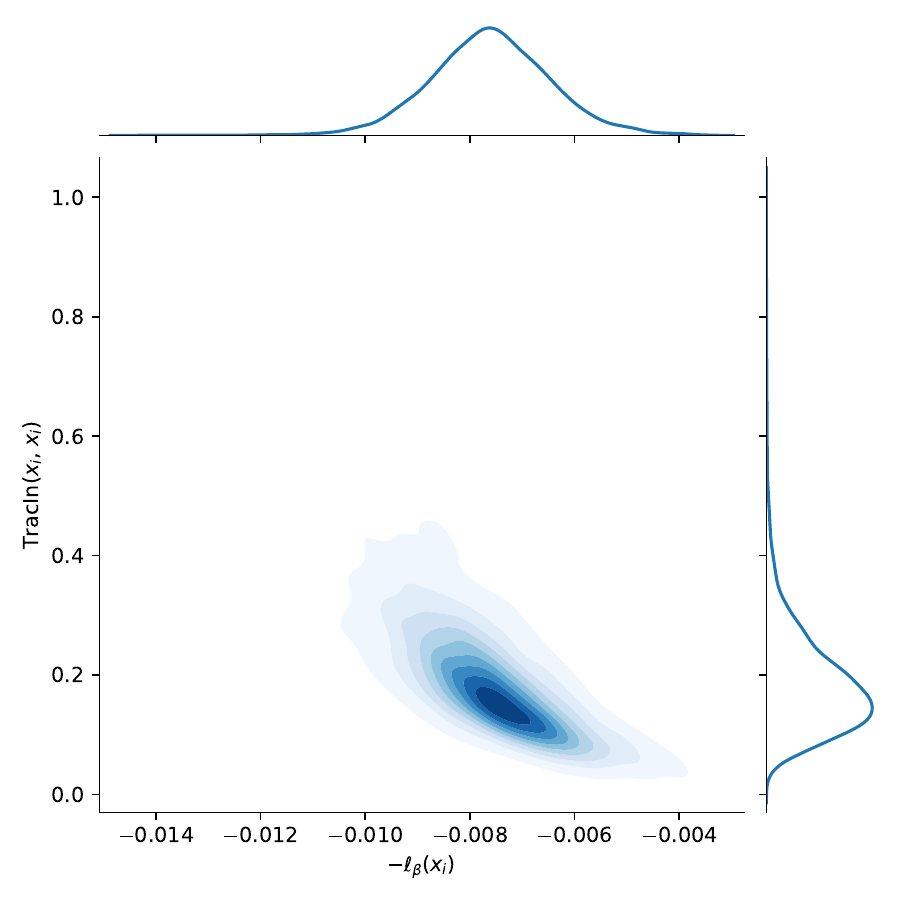}
    }
    \subfloat[][CIFAR$_4$]{ 
    	\includegraphics[trim=10 10 0 0, clip, width=0.19\textwidth]{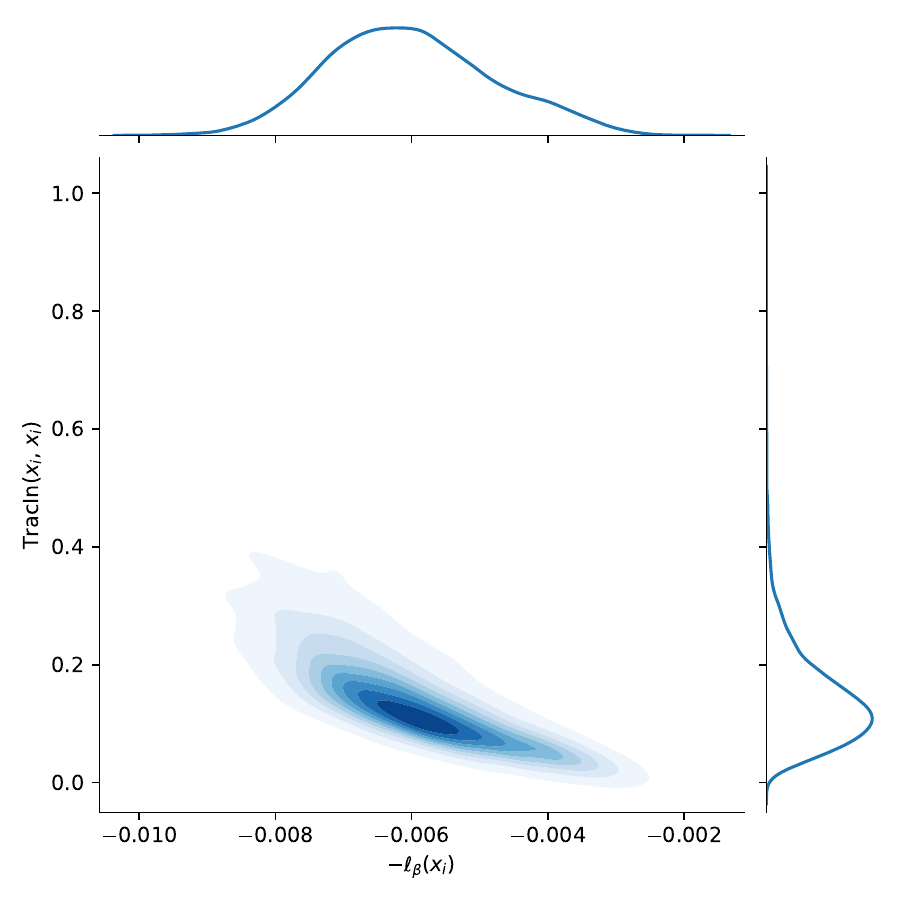}
    }\\ \vspace{-0.5em}
    \subfloat[][CIFAR$_5$]{ 
    	\includegraphics[trim=10 10 0 0, clip, width=0.19\textwidth]{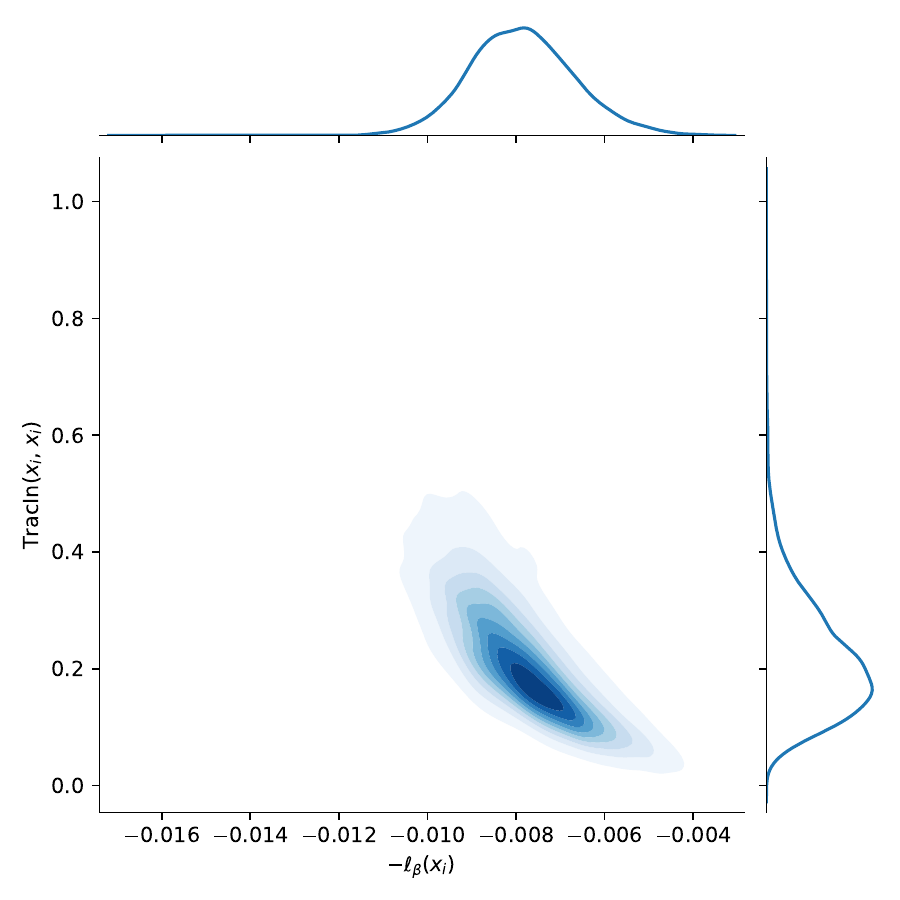}
    }
    \subfloat[][CIFAR$_6$]{ 
    	\includegraphics[trim=10 10 0 0, clip, width=0.19\textwidth]{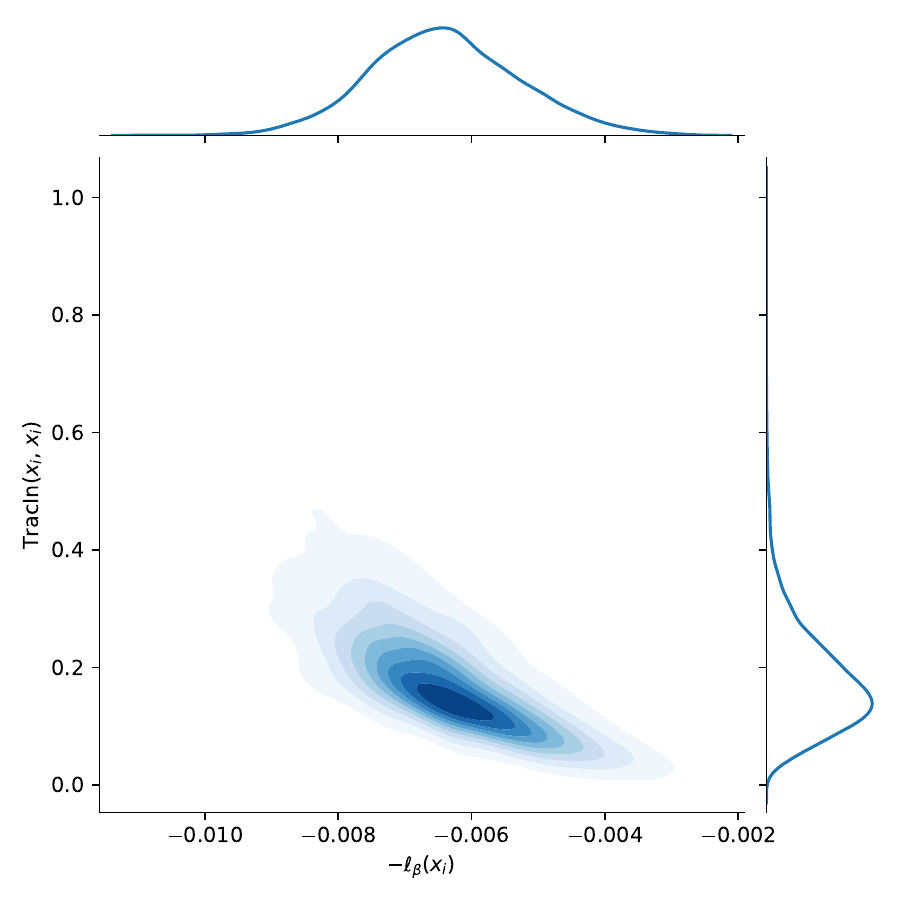}
    }
    \subfloat[][CIFAR$_7$]{ 
    	\includegraphics[trim=10 10 0 0, clip, width=0.19\textwidth]{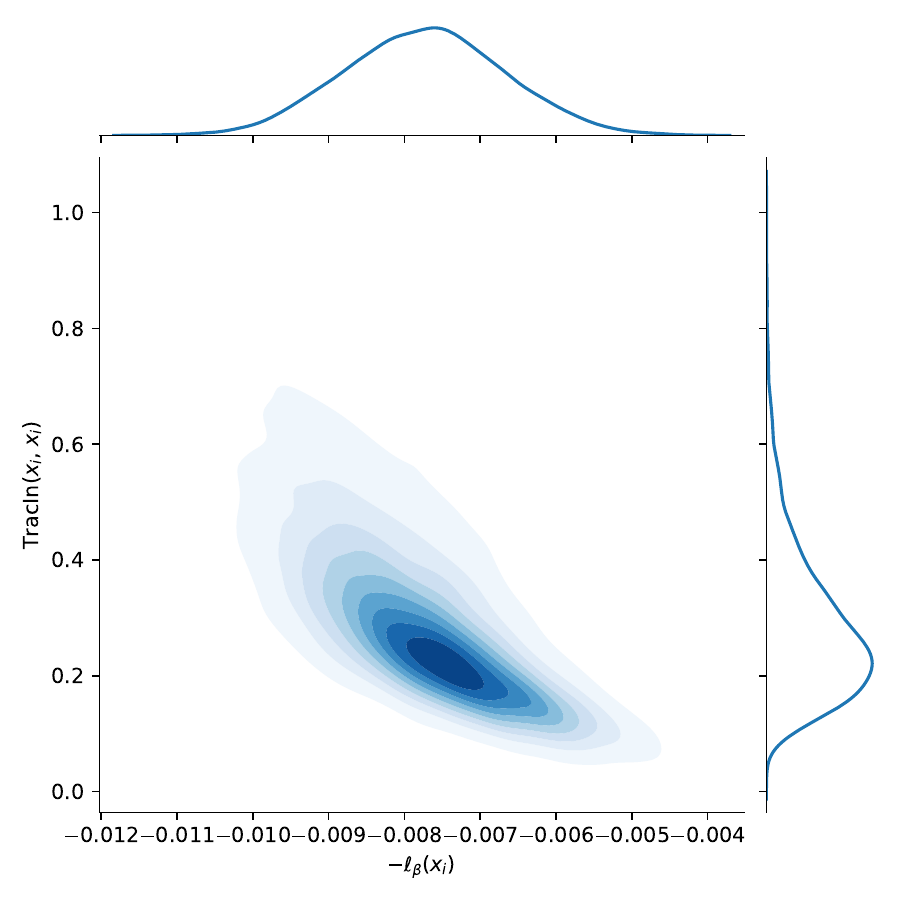}
    }
    \subfloat[][CIFAR$_8$]{ 
    	\includegraphics[trim=10 10 0 0, clip, width=0.19\textwidth]{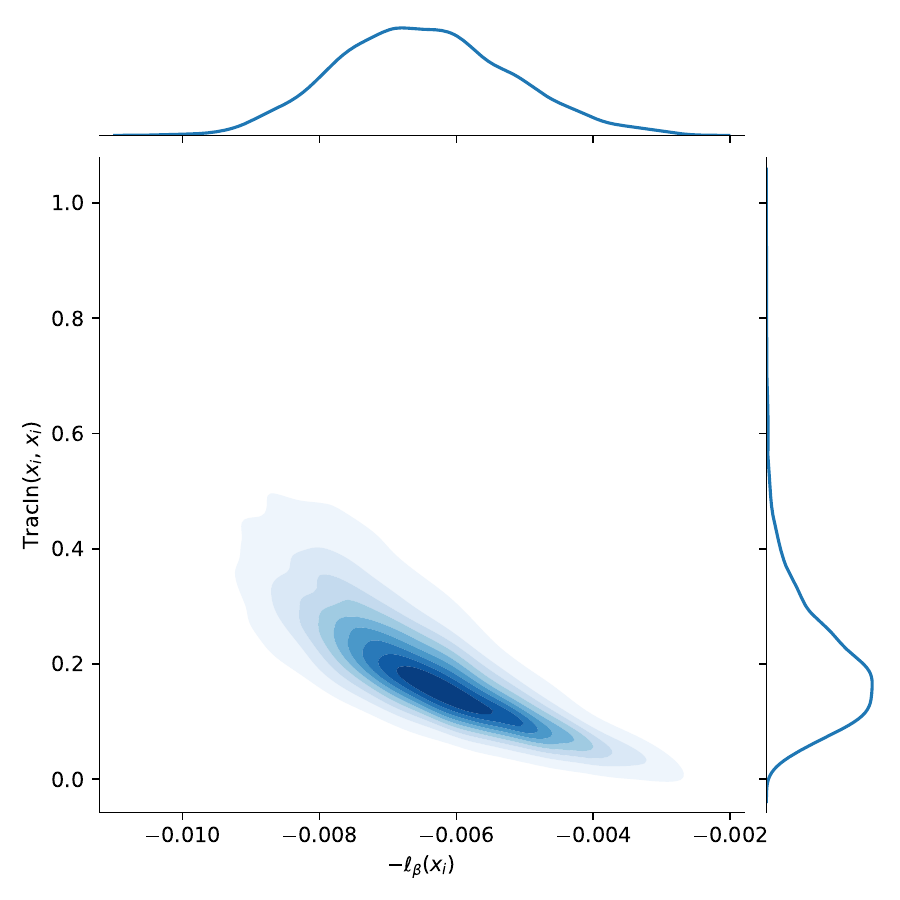}
    }
    \subfloat[][CIFAR$_9$]{ 
    	\includegraphics[trim=10 10 0 0, clip, width=0.19\textwidth]{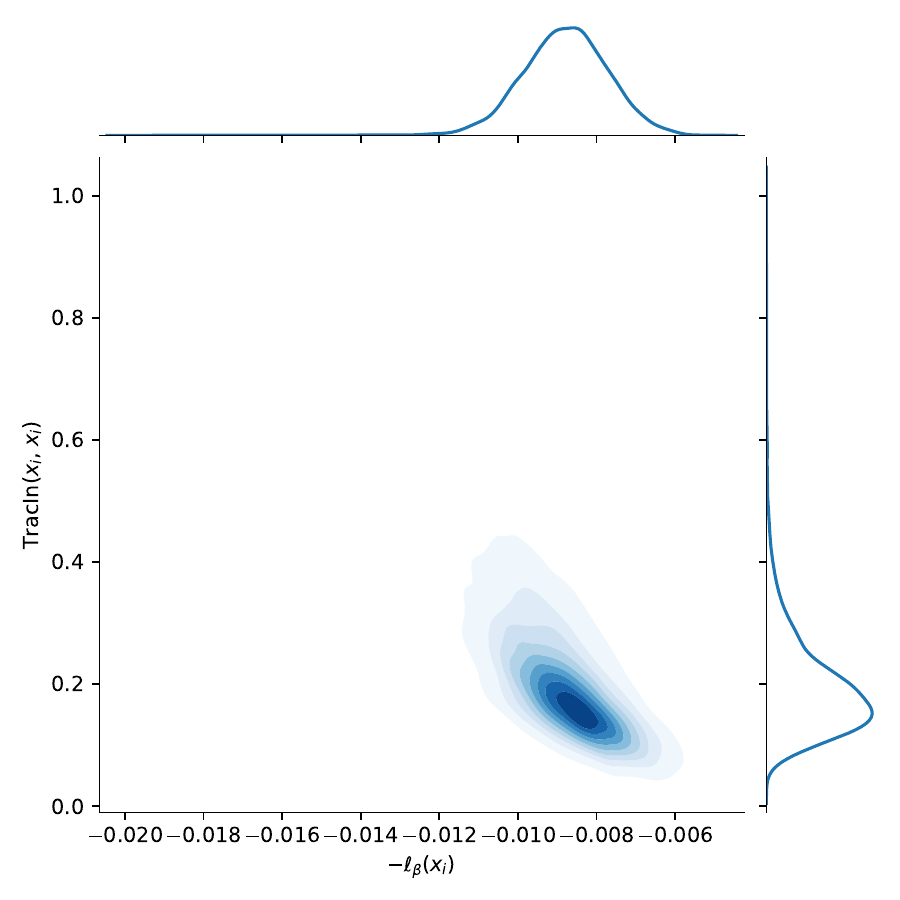}
    }
    \caption{Scatter and density plots of self influences versus negative losses of all training samples in each CIFAR subclass. The linear regressors show that high self influence samples have large losses.}
    \label{fig: cifar_k self inf vs loss appendix}
\end{figure}

\begin{figure}[!h]
    \centering
    \subfloat[][CIFAR$_1$]{ 
    	\includegraphics[width=0.32\textwidth]{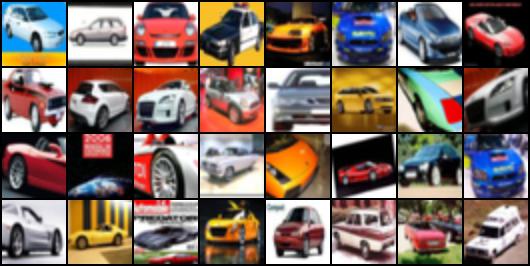}
    }
    \subfloat[][CIFAR$_2$]{ 
    	\includegraphics[width=0.32\textwidth]{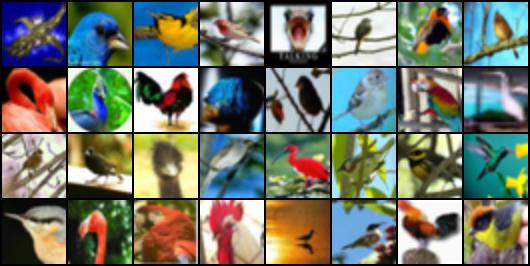}
    }
    \subfloat[][CIFAR$_3$]{ 
    	\includegraphics[width=0.32\textwidth]{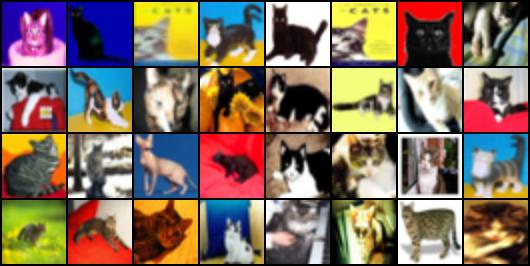}
    }\\ \vspace{-0.5em}
    \subfloat[][CIFAR$_4$]{ 
    	\includegraphics[width=0.32\textwidth]{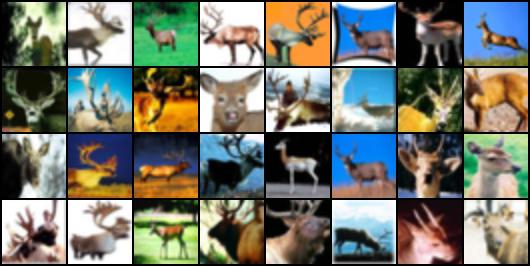}
    }
    \subfloat[][CIFAR$_5$]{ 
    	\includegraphics[width=0.32\textwidth]{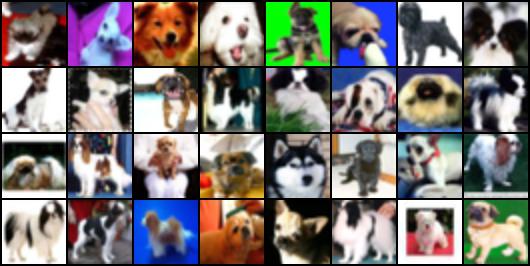}
    }
    \subfloat[][CIFAR$_6$]{ 
    	\includegraphics[width=0.32\textwidth]{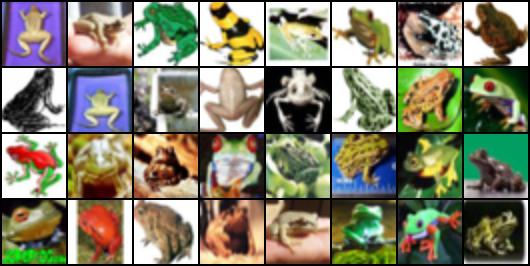}
    }\\ \vspace{-0.5em}
    \subfloat[][CIFAR$_7$]{ 
    	\includegraphics[width=0.32\textwidth]{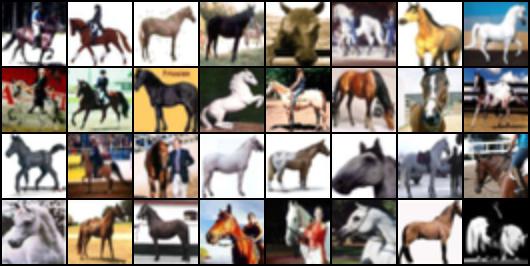}
    }
    \subfloat[][CIFAR$_8$]{ 
    	\includegraphics[width=0.32\textwidth]{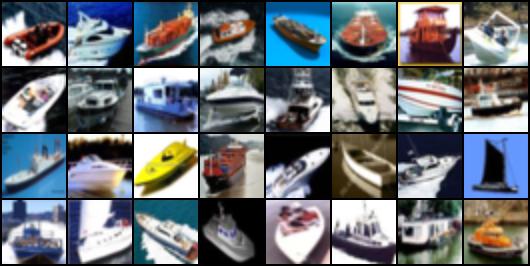}
    }
    \subfloat[][CIFAR$_9$]{ 
    	\includegraphics[width=0.32\textwidth]{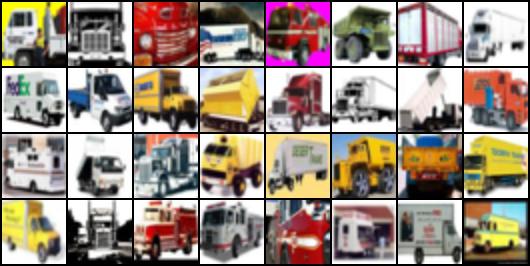}
    }
    \caption{High self influence samples in each CIFAR subclass. These samples are visually high-contrast and bright.}
    \label{fig: cifar high self inf visualization appendix}
\end{figure}

\begin{figure}[!h]
    \centering
    \subfloat[][CIFAR$_1$]{ 
    	\includegraphics[width=0.32\textwidth]{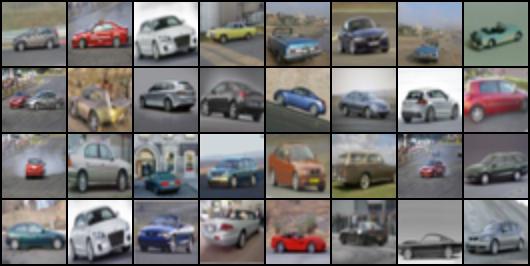}
    }
    \subfloat[][CIFAR$_2$]{ 
    	\includegraphics[width=0.32\textwidth]{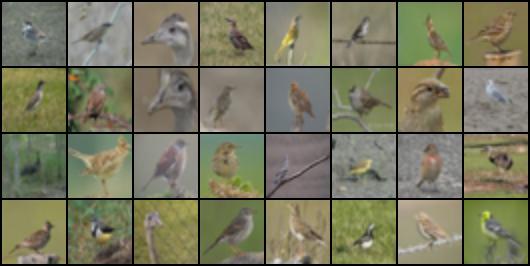}
    }
    \subfloat[][CIFAR$_3$]{ 
    	\includegraphics[width=0.32\textwidth]{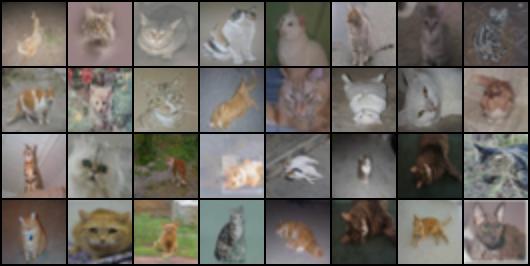}
    }\\ \vspace{-0.5em}
    \subfloat[][CIFAR$_4$]{ 
    	\includegraphics[width=0.32\textwidth]{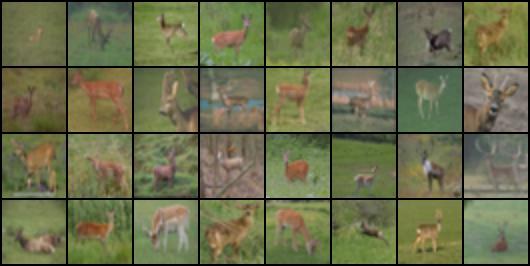}
    }
    \subfloat[][CIFAR$_5$]{ 
    	\includegraphics[width=0.32\textwidth]{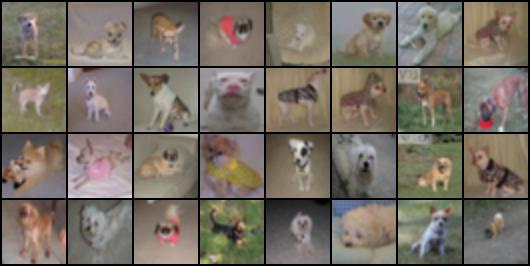}
    }
    \subfloat[][CIFAR$_6$]{ 
    	\includegraphics[width=0.32\textwidth]{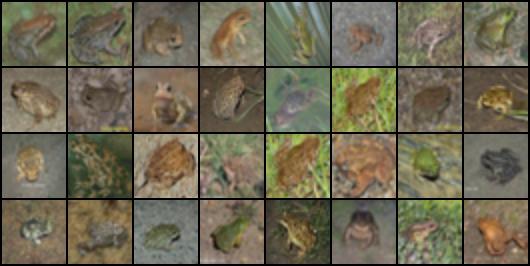}
    }\\ \vspace{-0.5em}
    \subfloat[][CIFAR$_7$]{ 
    	\includegraphics[width=0.32\textwidth]{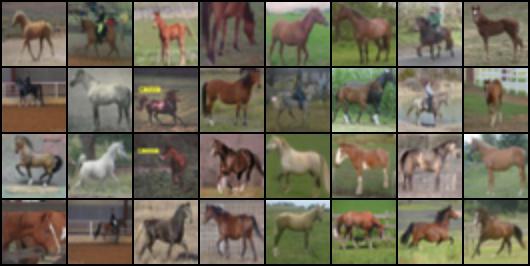}
    }
    \subfloat[][CIFAR$_8$]{ 
    	\includegraphics[width=0.32\textwidth]{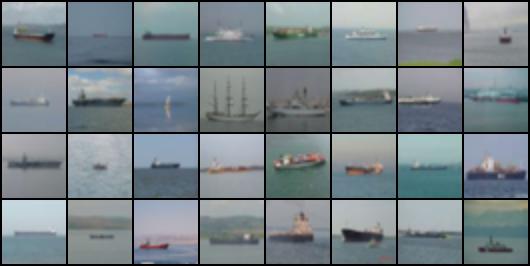}
    }
    \subfloat[][CIFAR$_9$]{ 
    	\includegraphics[width=0.32\textwidth]{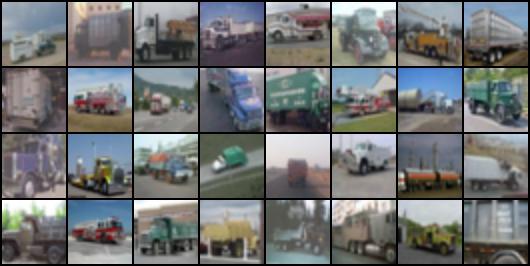}
    }\\
    \caption{Low self influence samples in each CIFAR subclass. These samples are visually similar in shape or background.}
    \label{fig: cifar low self inf visualization appendix}
\end{figure}

In CIFAR and CIFAR subclass experiments, we compare self influences and losses across different hyperparameters. Similar to Appendix \ref{appendix: self-inf mnist}, we demonstrate scatter and density plots, and report $R^2$ scores of linear regression models fit to these data. Comparisons on CIFAR are shown in Figure \ref{fig: cifar self inf vs loss dlatent appendix}, and CIFAR subclasses in Figure \ref{fig: cifar_k self inf vs loss appendix}. In all settings high self influence samples have large losses. We then visualize high and low self influence samples from each CIFAR subclass in Figure \ref{fig: cifar high self inf visualization appendix} and Figure \ref{fig: cifar low self inf visualization appendix}, respectively. 

\newpage

\newpage
\subsection{Application on Unsupervised Data Cleaning}\label{appendix: data cleaning}

\begin{figure}[!h]
    \centering
    \subfloat[][EMNIST versus MNIST]{ 
    	\includegraphics[trim=30 5 30 190, clip, width=0.48\textwidth]{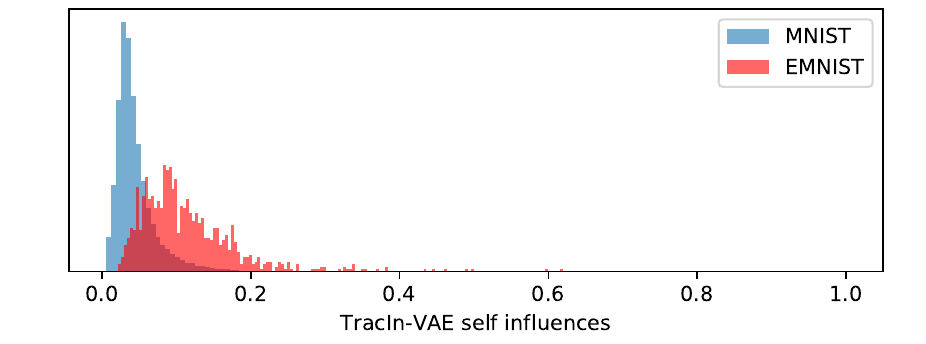}
    	\label{fig: rare distr mnist}
    }
    \subfloat[][CelebA versus CIFAR]{ 
    	\includegraphics[trim=30 5 30 190, clip, width=0.48\textwidth]{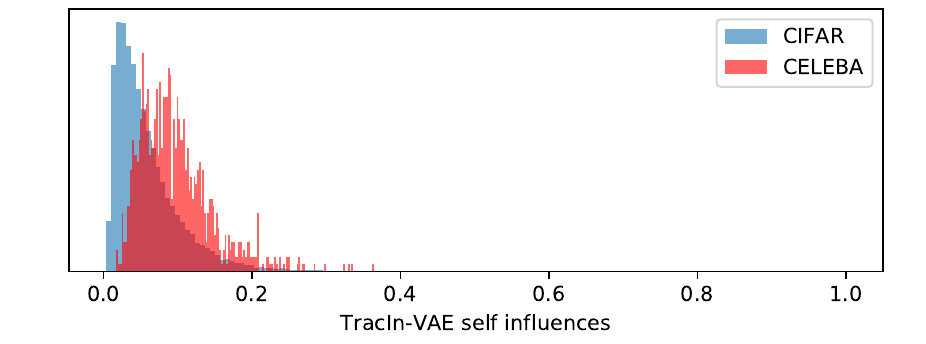}
	\label{fig: rare distr cifar}
    }
    \caption{Distributions of self influences of 1k extra samples versus samples from the original dataset. Distributions are normalized as densities for better visualization. It is shown that extra samples have higher self influences.}
    \label{fig: rare distr}
\end{figure}

\begin{figure}[!h]
    \centering
        \centering
        \subfloat[][EMNIST versus MNIST]{ 
            \includegraphics[trim=20 5 30 30, clip, width=0.4\textwidth]{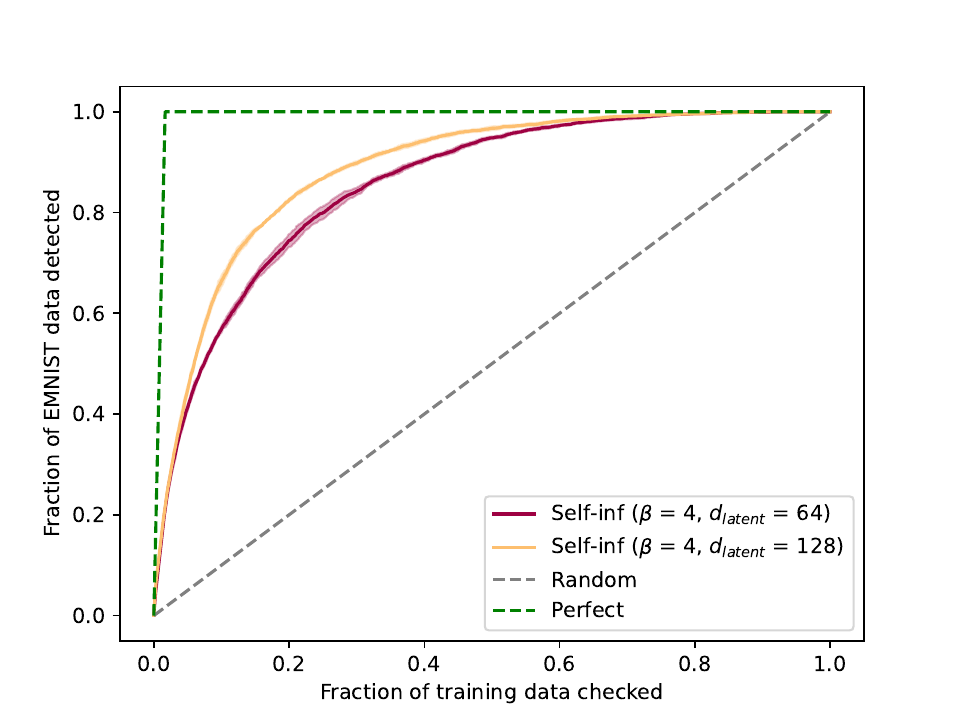}
            \label{fig: rare auc mnist}
        }
        \subfloat[][CelebA versus CIFAR]{ 
            \includegraphics[trim=20 5 30 30, clip, width=0.4\textwidth]{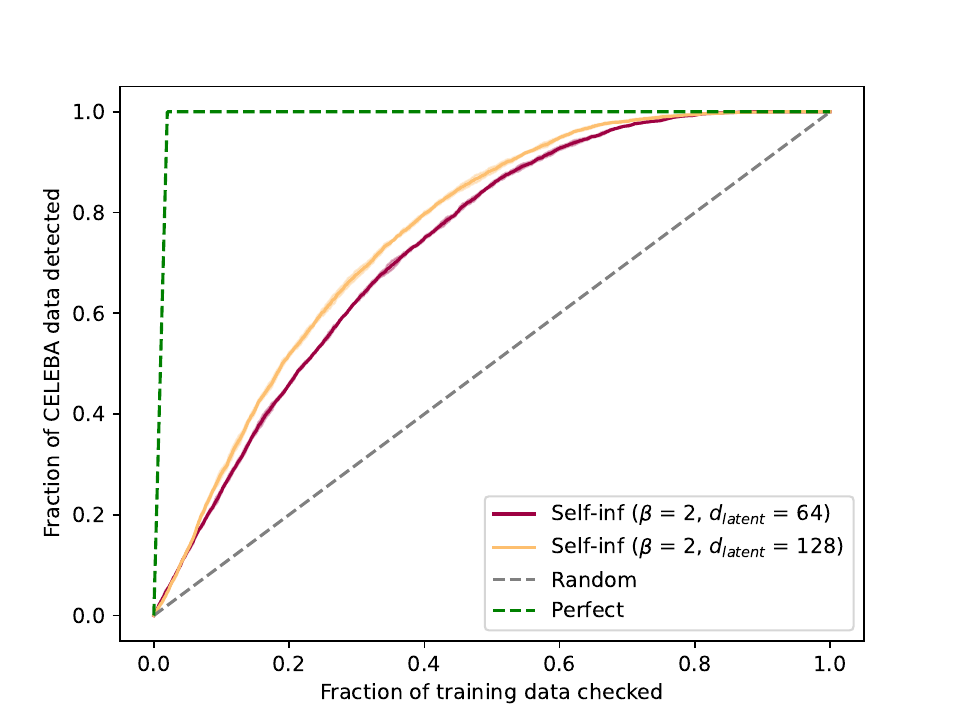} 
       	    \label{fig: rare auc cifar}
        }
        \caption{Detection curves (fraction of extra samples detected versus fraction of training data checked in the self influence order) with standard errors. It is shown that extra samples can be detected by sorting self influences.}
        \label{fig: rare auc}
\end{figure}

We plot the distribution of self influences of extra samples (EMNIST or CelebA) and original samples (MNIST or CIFAR) in Figure \ref{fig: rare distr}. We plot the detection curves in Figure \ref{fig: rare auc}, where the horizontal axis is the fraction of all samples checked when they are sorted in the self influence order, and the vertical axis is the fraction of extra samples found. The area under these detection curves (AUC) are reported in Table \ref{tab: rare auc}. These experiments are repeated five times to reduce randomness. The results indicate that extra samples have higher self influences than original samples. This justifies the potential to apply VAE-TracIn to unsupervised data cleaning.

\begin{table}[!h]
        \caption{Mean area-under-curve (AUC) $\pm$ standard errors of detection curves in Figure \ref{fig: rare auc}. A higher AUC indicates extra samples can be better detected by sorting self influences, where AUC$\approx1$ implies perfect detection and AUC $\approx0.5$ implies random selection. The results indicate detection on the simple MNIST + EMNIST datasets is better than the more complicated CIFAR + CelebA datasets. In addition, a higher $\dlatent$ leads to slightly better AUC.}
        \vspace{0.5em}
        \label{tab: rare auc}
        \centering
        \begin{tabular}{ccc|c}
           \hline
           Original dataset & Extra samples & $\dlatent$ & AUC \\ \hline
           MNIST & EMNIST & 64 & 0.858$\pm$0.003 \\
           MNIST & EMNIST & 128 & 0.887$\pm$0.002 \\
           CIFAR & CelebA & 64 & 0.735$\pm$0.002 \\
           CIFAR & CelebA & 128 & 0.760$\pm$0.001 \\ \hline
        \end{tabular}
\end{table}

\begin{figure}[!h]
    \centering
    \subfloat[][$\beta=1$, $\dlatent=128$ ]{ 
    	\includegraphics[trim=30 5 30 30, clip, width=0.32\textwidth]{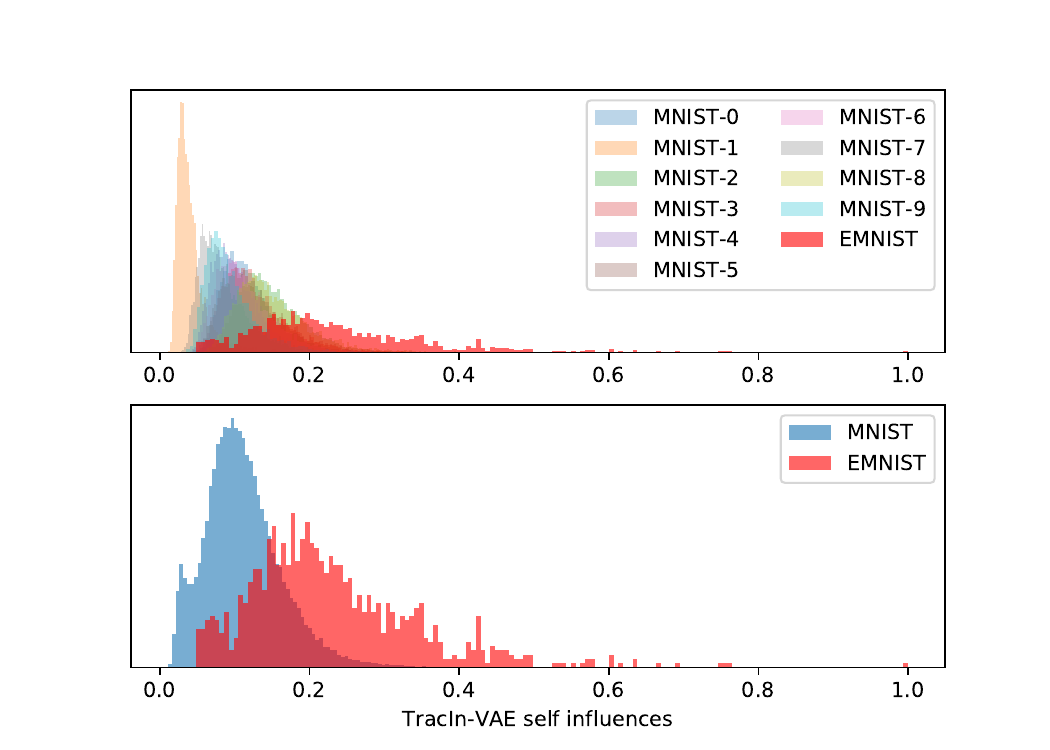}
    }
    \subfloat[][$\beta=2$, $\dlatent=128$ ]{ 
    	\includegraphics[trim=30 5 30 30, clip, width=0.32\textwidth]{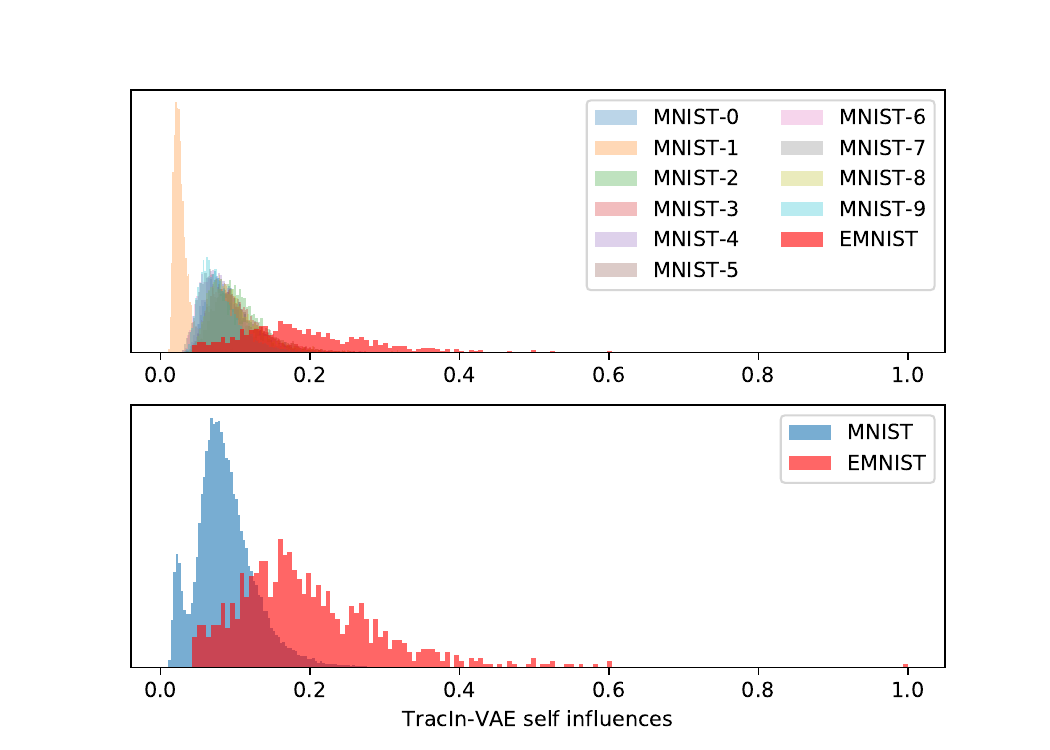}
    }
    \subfloat[][$\beta=4$, $\dlatent=128$ ]{ 
    	\includegraphics[trim=30 5 30 30, clip, width=0.32\textwidth]{outlier_mnist_emnist_outlier_beta4_z128.pdf}
    }\\
    \vspace{1em}
    
    \subfloat[][$\beta=1$ , $\dlatent=16$]{ 
    	\includegraphics[trim=30 5 30 30, clip, width=0.32\textwidth]{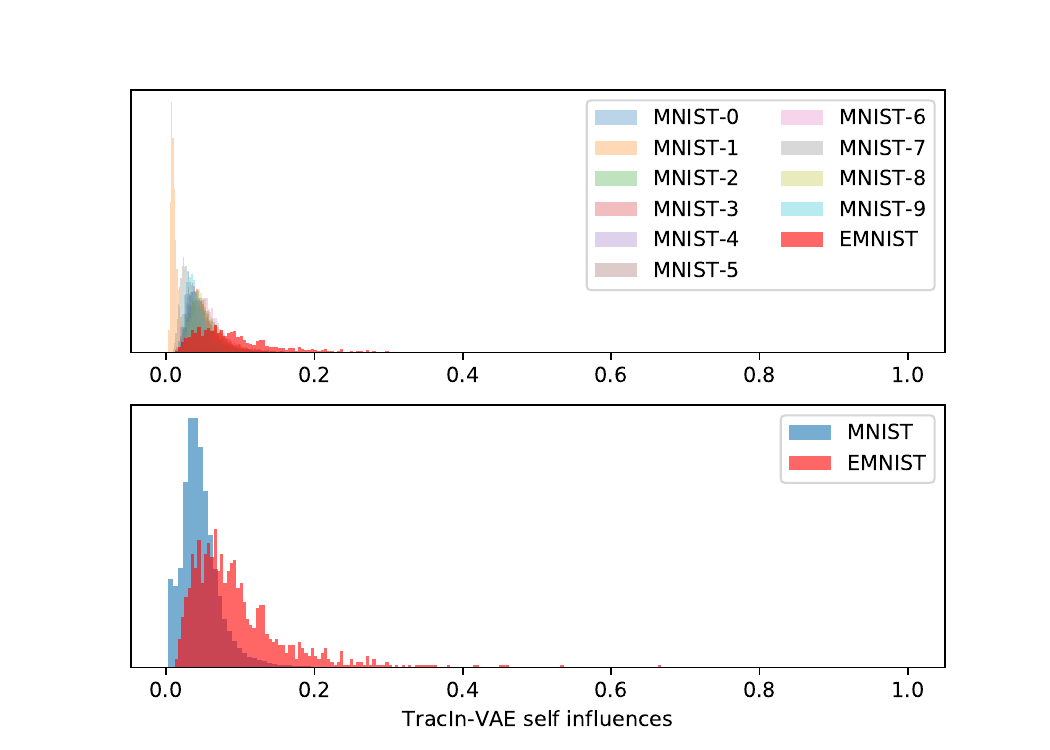}
    }
    \subfloat[][$\beta=1$ , $\dlatent=64$]{ 
    	\includegraphics[trim=30 5 30 30, clip, width=0.32\textwidth]{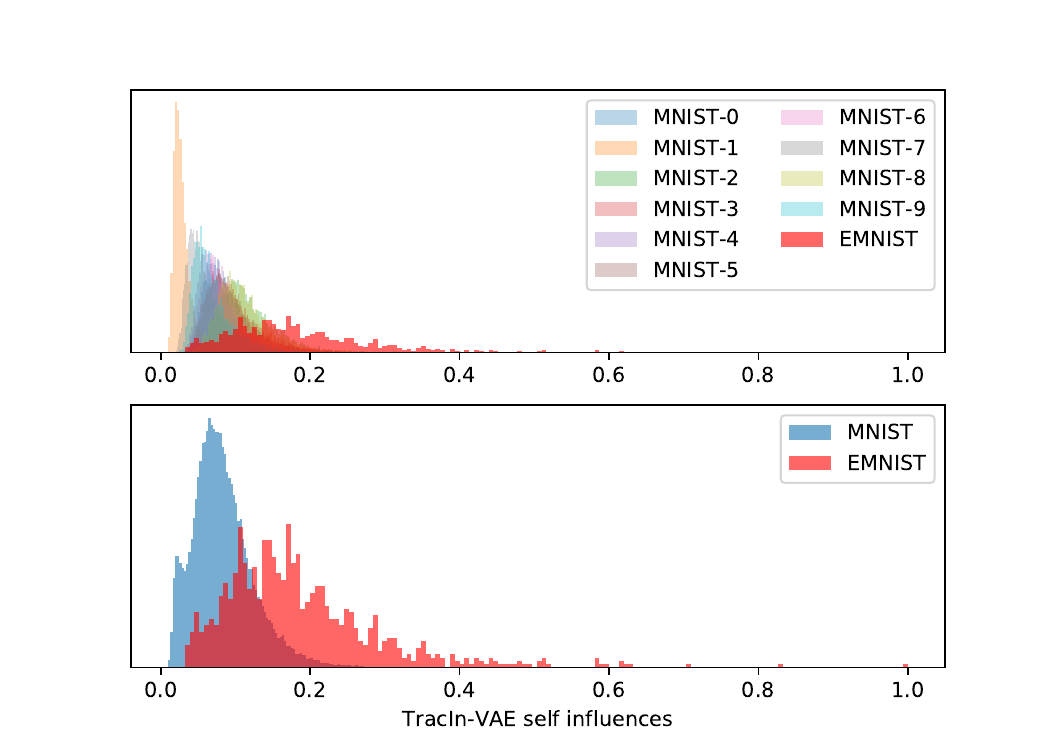}
    }
    \subfloat[][$\beta=1$ , $\dlatent=128$]{ 
    	\includegraphics[trim=30 5 30 30, clip, width=0.32\textwidth]{outlier_mnist_emnist_outlier_beta1_z128.pdf}
    }
    \caption{Distribution of self influences of extra sample and samples from the original MNIST dataset. In all settings, extra samples have higher self influences than original samples.}
    \label{fig: rare distr mnist appendix}
\end{figure}

\begin{figure}[!h]
    \centering
    \includegraphics[trim=20 5 30 30, clip, width=0.32\textwidth]{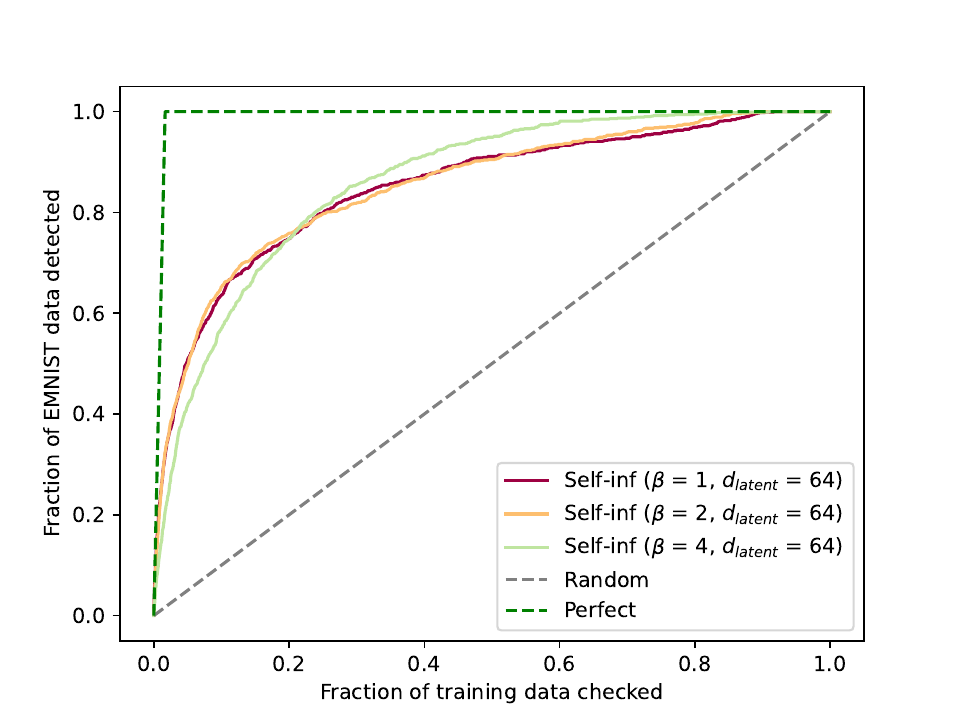} 
    \includegraphics[trim=20 5 30 30, clip, width=0.32\textwidth]{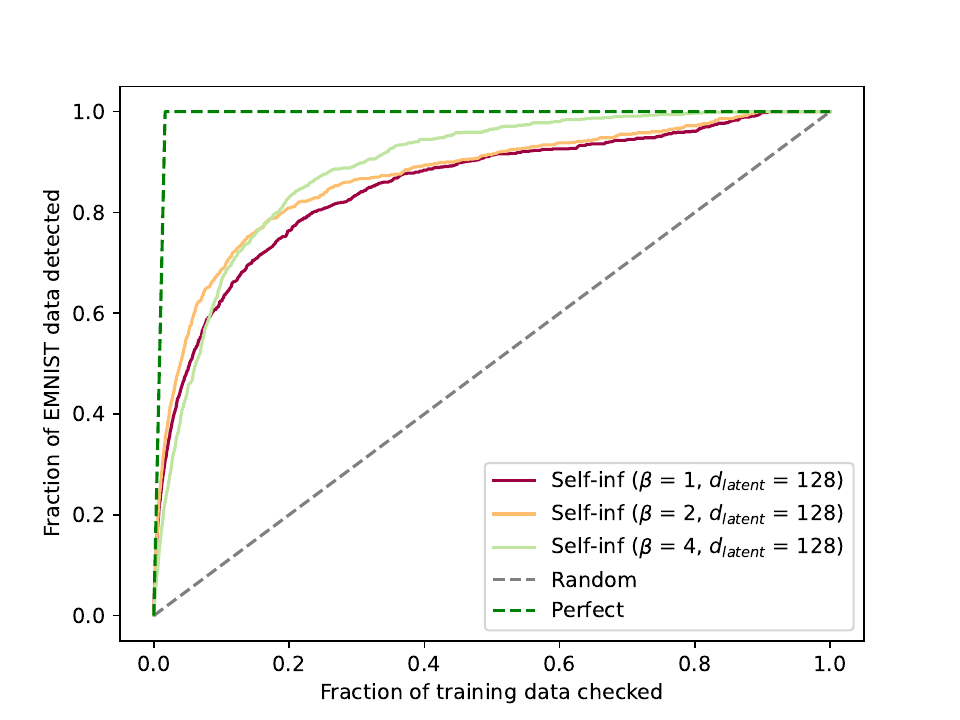} 
    \includegraphics[trim=20 5 30 30, clip, width=0.32\textwidth]{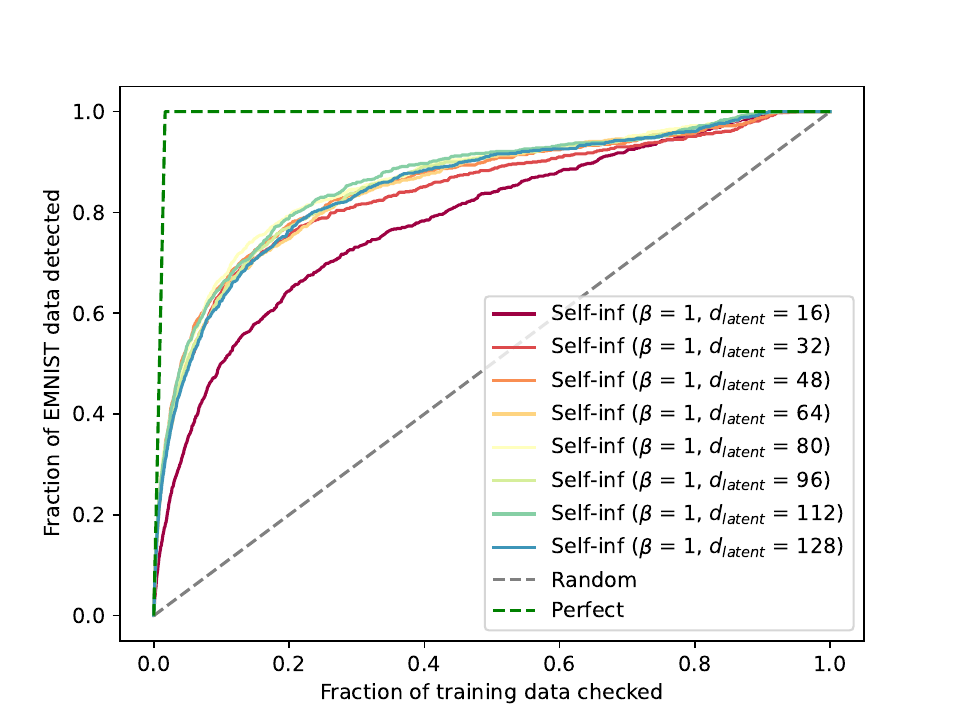} 
    \caption{Detection curves (EMNIST versus MNIST). It is shown that extra samples can be detected by sorting self influences. Increasing $\beta$ or $\dlatent$ can slightly improve detection.}
    \label{fig: rare auc appendix}
\end{figure}

\newpage
We next compare different hyperparameters under the MNIST + EMNIST setting. Distributions of self influences are shown in Figure \ref{fig: rare distr mnist appendix} and detection curves are shown in Figure \ref{fig: rare auc appendix}. In all settings, extra samples have higher self influences than original samples. Increasing $\beta$ or $\dlatent$ can slightly improve detection. 

In order to understand whether high self influence samples indeed harm the model, we remove a certain number of high self influence samples and retrain the $\beta$-VAE model under the MNIST + EMNIST setting. We report the results in Table \ref{tab: self inf retraining appendix}. We can observe consistent loss drop after deletion of high self influence samples.

\begin{table}[!h]
    \centering
    \caption{Average loss of the MNIST test set when performing retraining after removing a certain number of high self influence samples under the MNIST + EMNIST setting.}
    \vspace{0.5em}
    \begin{tabular}{r|cccccccccc}
    \hline
        \# removed & 0 & 4 & 8 & 16 & 32 & 64 & 128 & 256 & 512 & 1024 \\ \hline
        Loss ($\times10^{-3}$) & 4.24 & 4.21 & 4.18 & 4.19 & 4.18 & 4.21 & 4.17 & 4.17 & 4.18 & 4.18 \\
    \hline
    \end{tabular}
    \label{tab: self inf retraining appendix}
\end{table}

~~\newpage
\subsection{Influences over Test Data (MNIST)}\label{appendix: test inf mnist}

\begin{figure}[!h]
    \centering
    \includegraphics[width=0.9\textwidth]{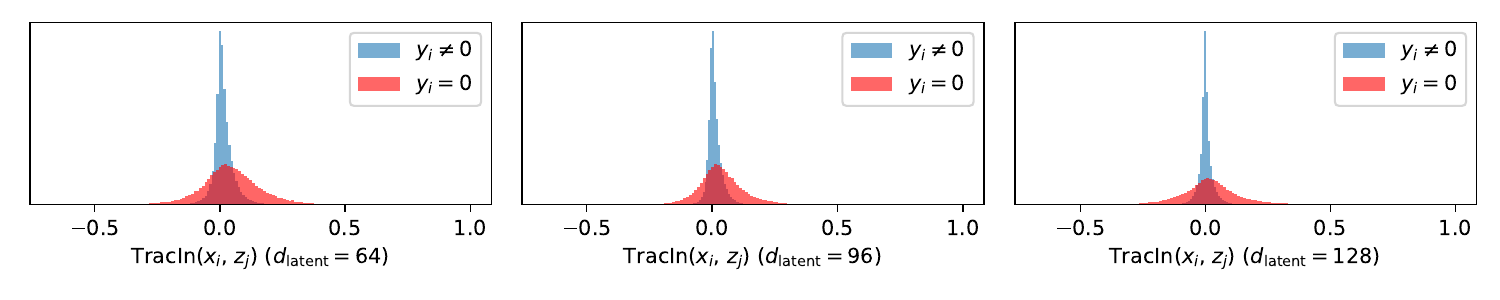}
    \caption{Distributions of influences of training samples ($x_i$) over test zeroes ($z_j$). The red distributions are training zeroes and the blue distributions are training non-zeroes. It is shown that a large proportion of strong proponents and opponents of test zeroes are training zeroes. }
    \label{fig: mnist test inf by class}
\end{figure}

\begin{figure}[!h]
    \centering
    \subfloat[][$\beta=4$, $\dlatent=64$]{ 
        \includegraphics[width=0.9\textwidth]{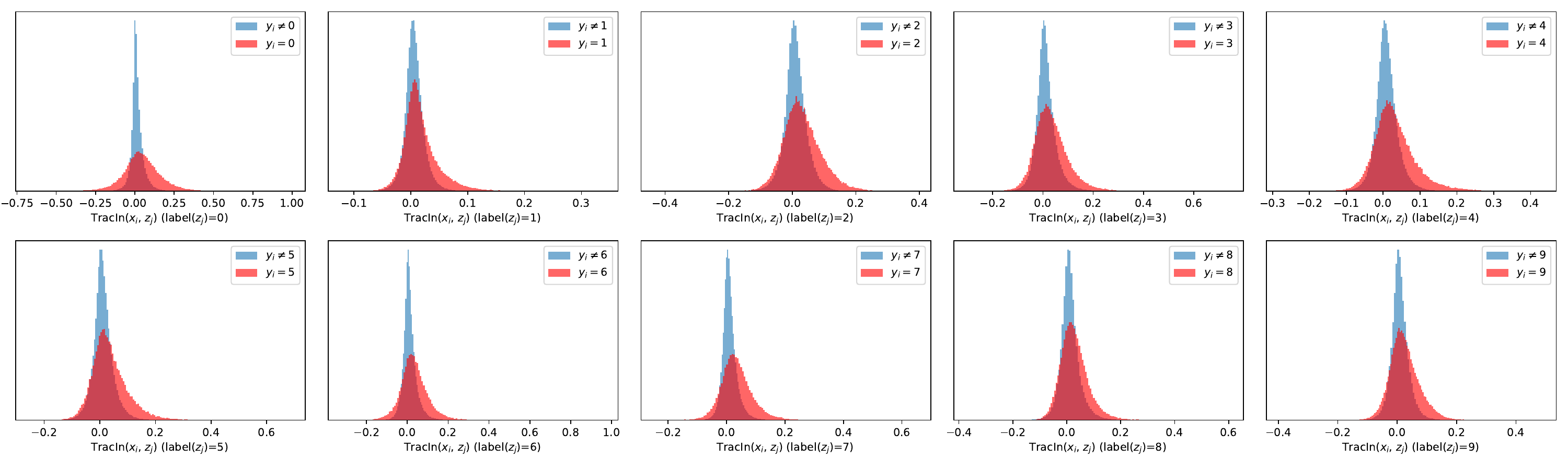}
    }\\ \vspace{-0.5em}
    \subfloat[][$\beta=4$, $\dlatent=96$]{ 
        \includegraphics[width=0.9\textwidth]{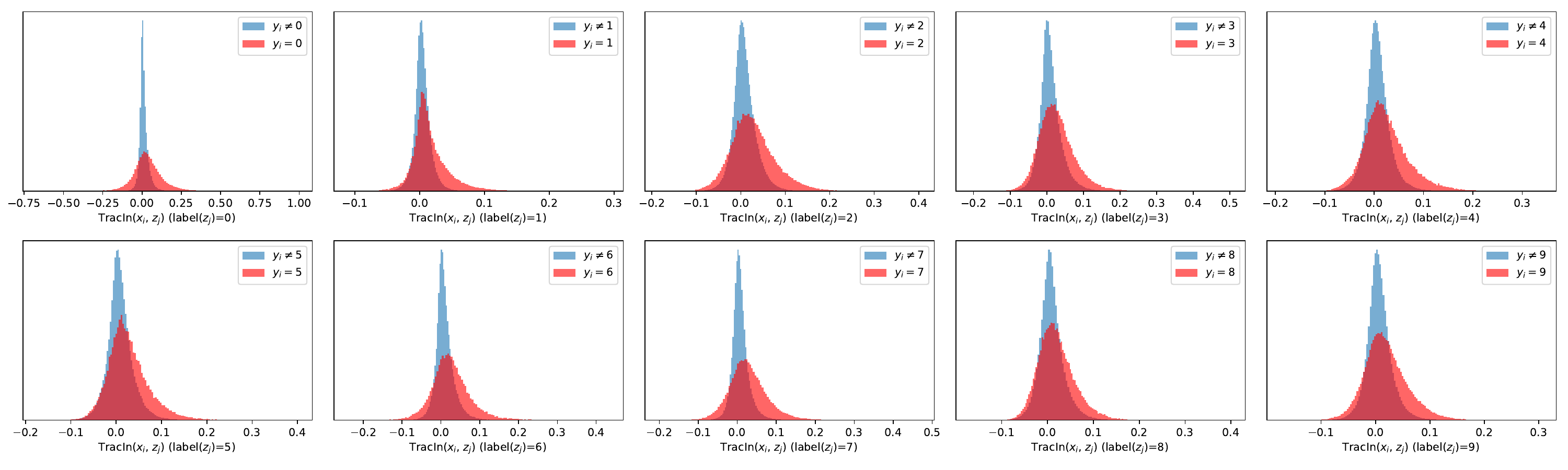}
    }\\ \vspace{-0.5em}
    \subfloat[][$\beta=4$, $\dlatent=128$]{ 
        \includegraphics[width=0.9\textwidth]{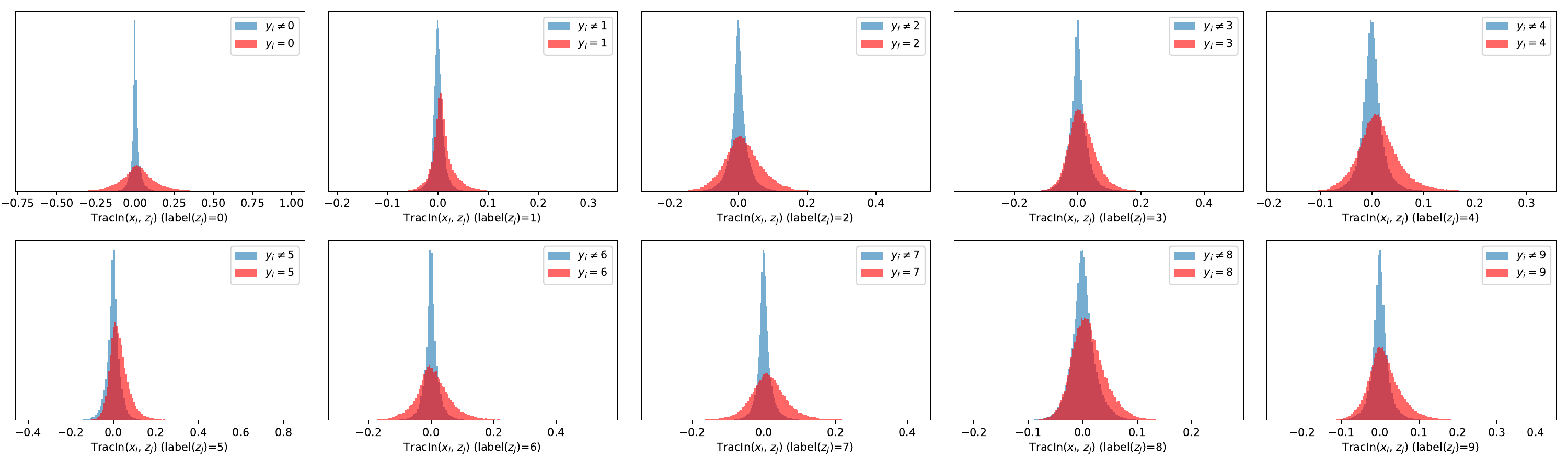}
    }
    \caption{Distributions of influences of training samples ($x_i$) over test samples ($z_j$). The red distributions are $x_i$ in the same class as $z_j$, and the blue distributions are $x_i$ in a different class as $z_j$. For most labels the strongest proponents and opponents are very likely from the same class. For the rest the strongest opponents seem equally likely from the same or a different class.}
    \label{fig: mnist test inf by class appendix}
\end{figure}

In Figure \ref{fig: mnist test inf by class}, we plot the distributions of influences of training zeroes (red distributions) and non-zeroes (blue distributions) over test zeroes. Full results on all classes are shown in Figure \ref{fig: mnist test inf by class appendix}. For most labels including 0, 2, 4, 6, 7, and 9, the strongest proponents and opponents are very likely from the same class. For the rest of the labels including 1, 3, 5, and 8, the strongest opponents seem equally likely from the same or a different class. 

\begin{figure}[!h]
    \centering
    \subfloat[][$\beta=1$, $\dlatent=128$]{ 
    	\includegraphics[trim=0 0 30 30, clip, width=0.25\textwidth]{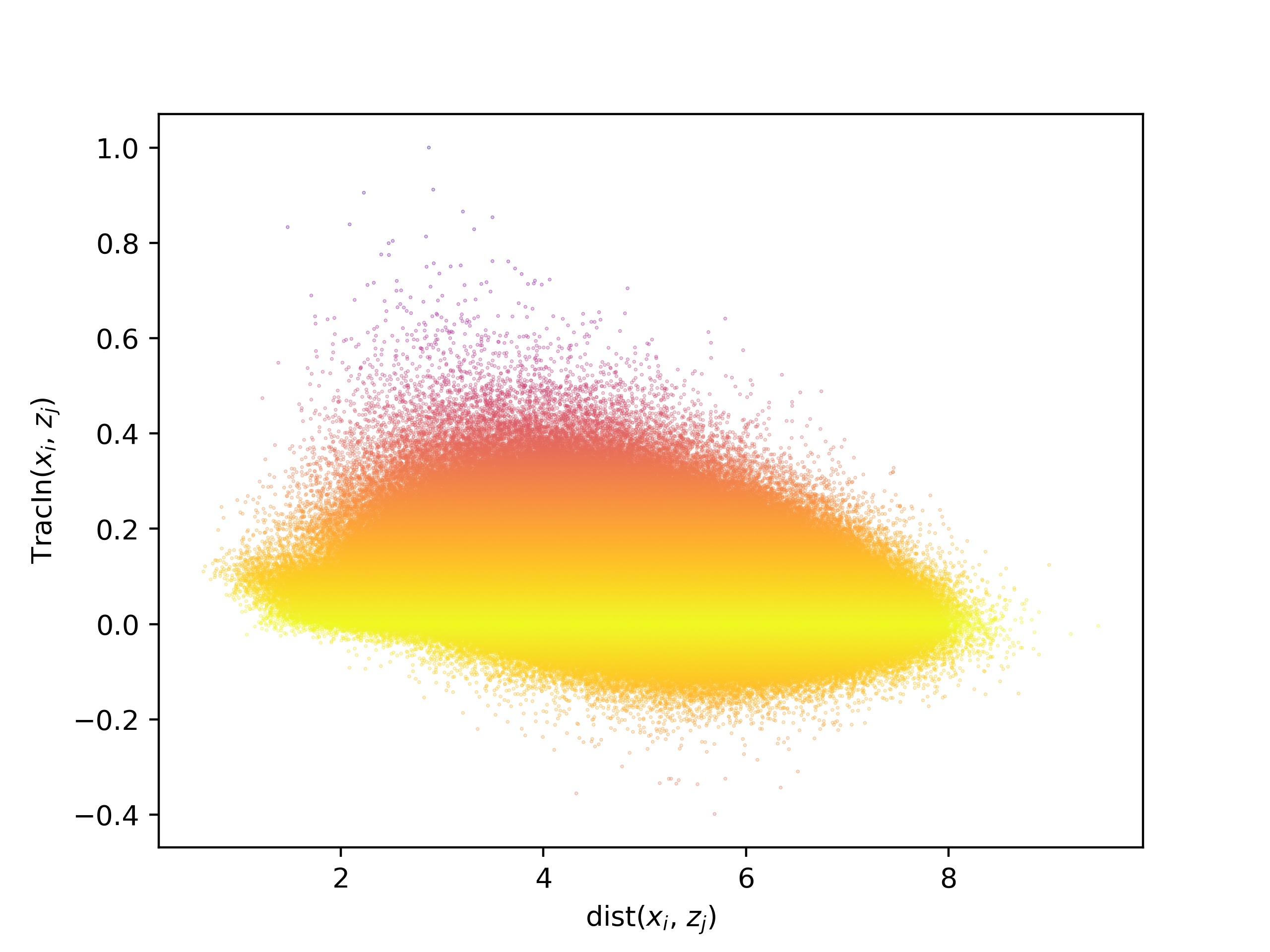}
    }
    \subfloat[][$\beta=2$, $\dlatent=128$]{ 
    	\includegraphics[trim=0 0 30 30, clip, width=0.25\textwidth]{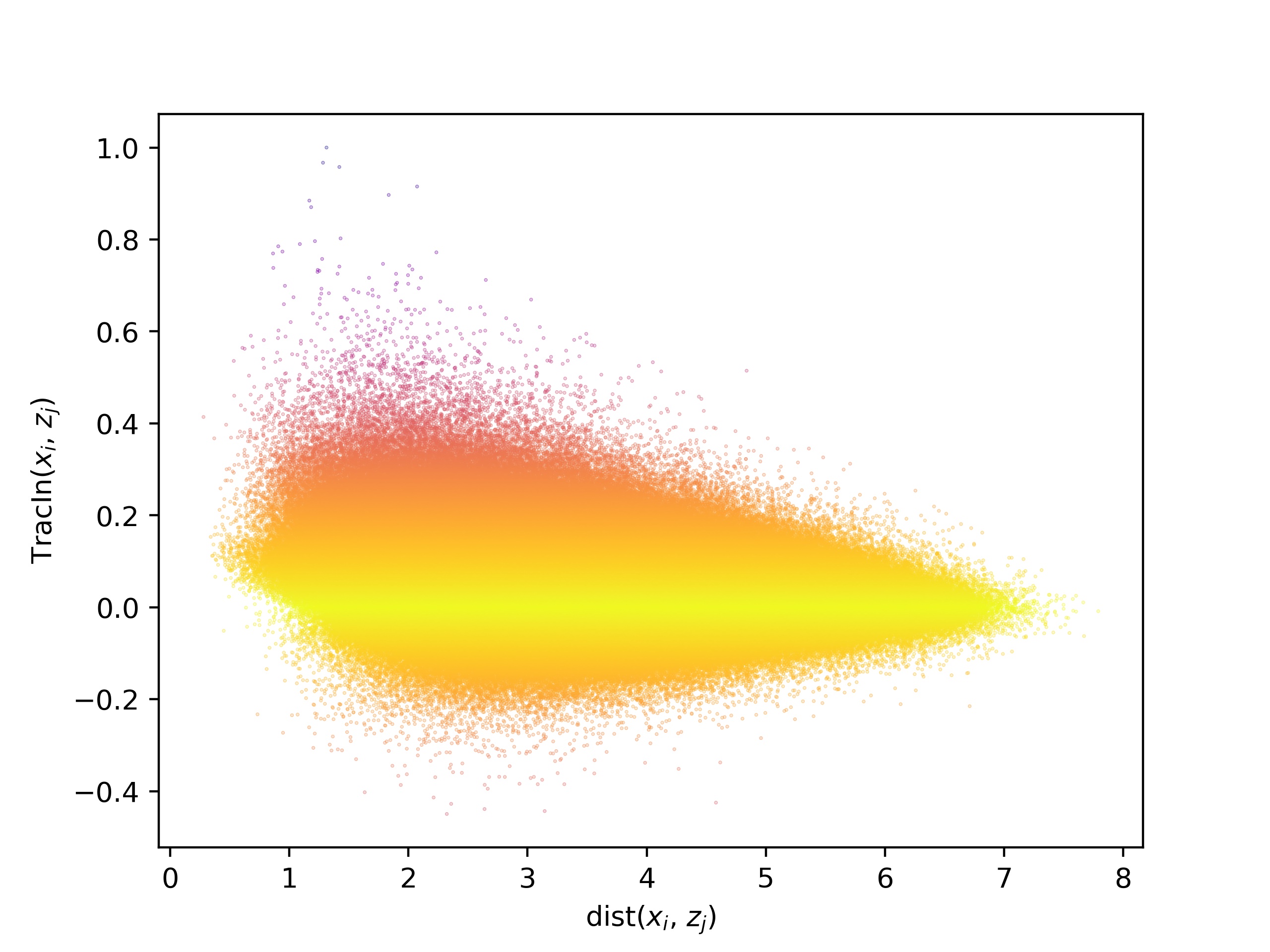}
    }
    \subfloat[][$\beta=4$, $\dlatent=128$]{ 
    	\includegraphics[trim=0 0 30 30, clip, width=0.25\textwidth]{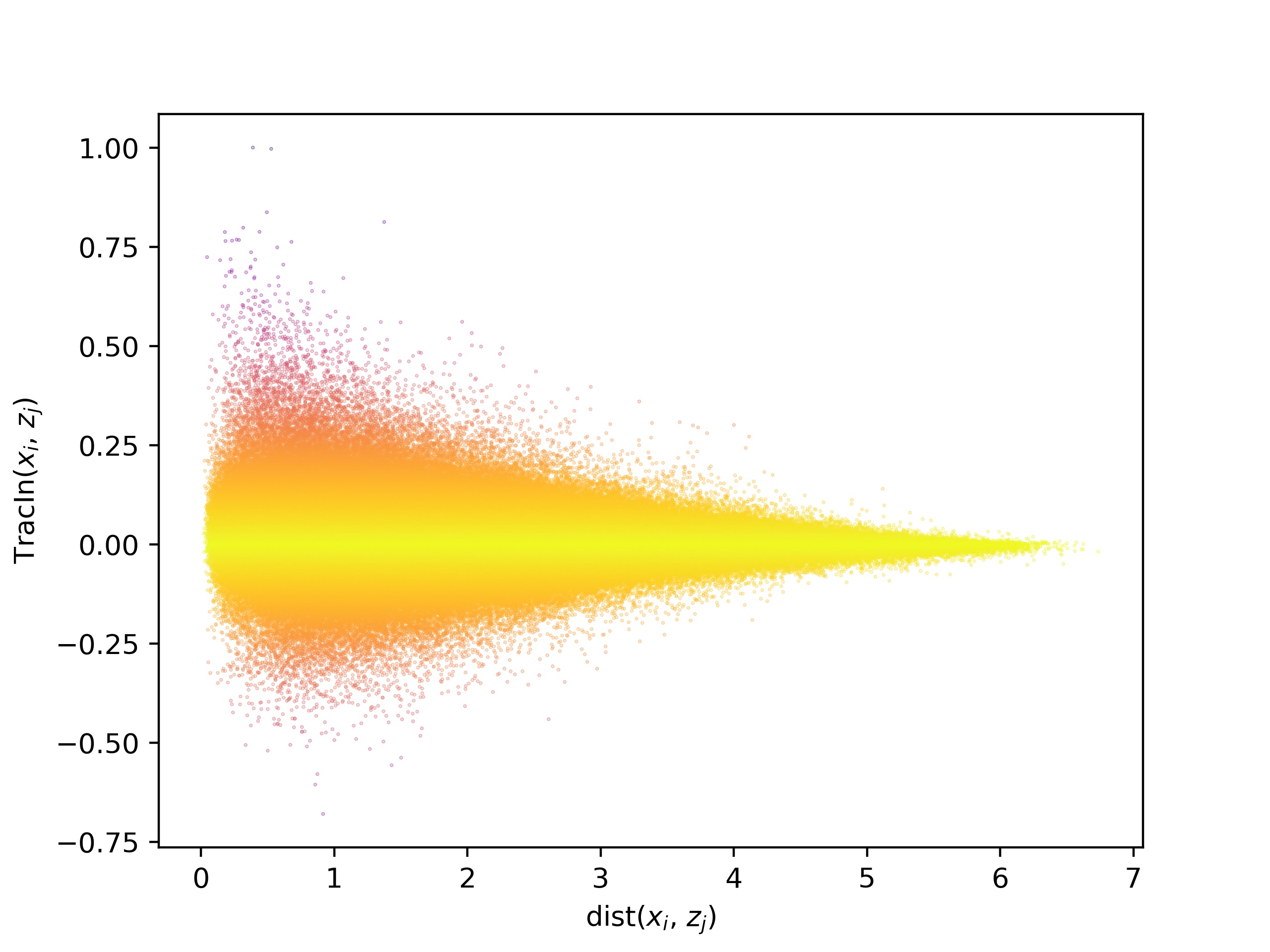}
    }\\
    \subfloat[][$\beta=4$, $\dlatent=64$]{ 
    	\includegraphics[trim=0 0 30 30, clip, width=0.25\textwidth]{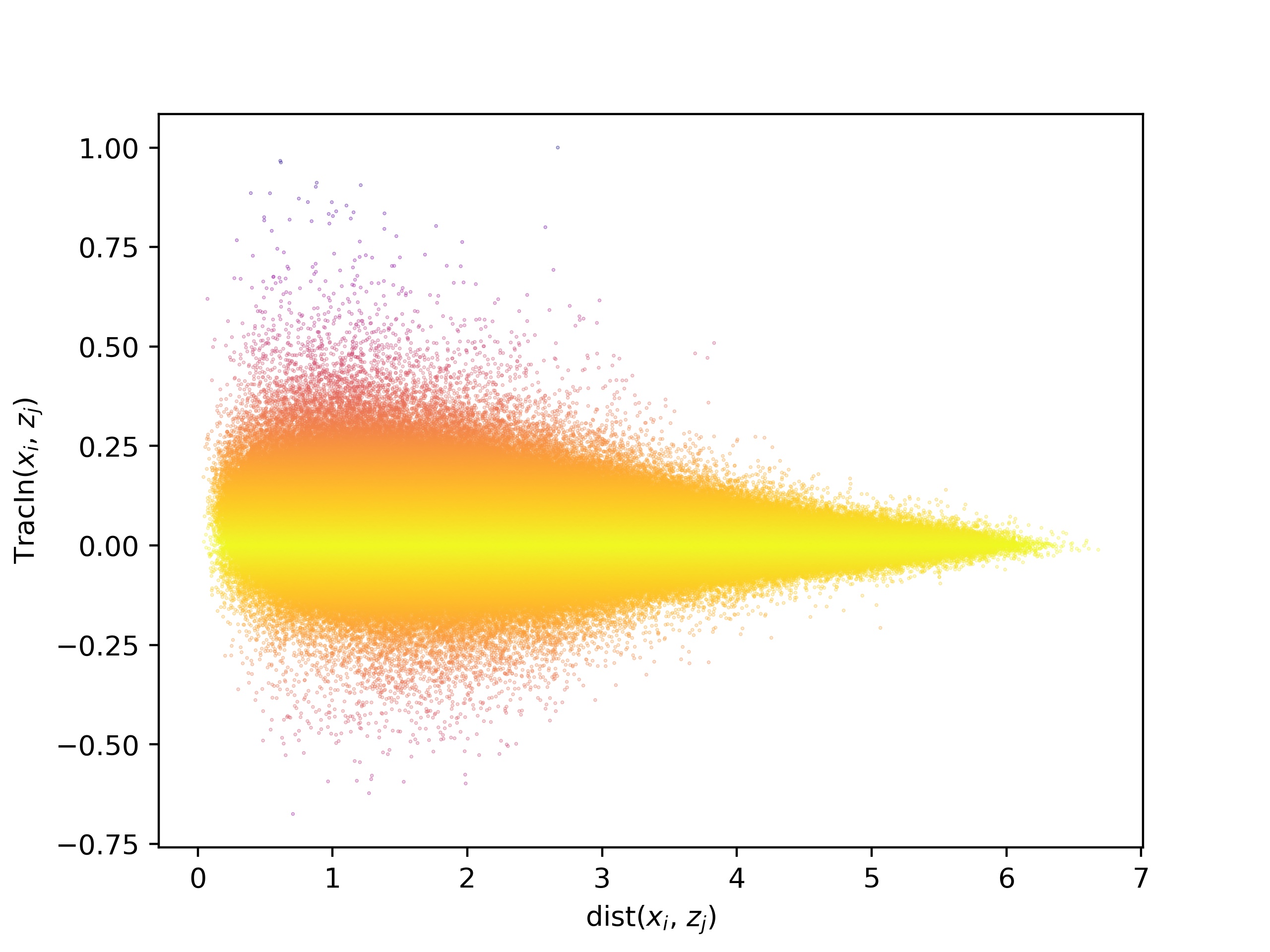}
    }
    \subfloat[][$\beta=4$, $\dlatent=96$]{ 
    	\includegraphics[trim=0 0 30 30, clip, width=0.25\textwidth]{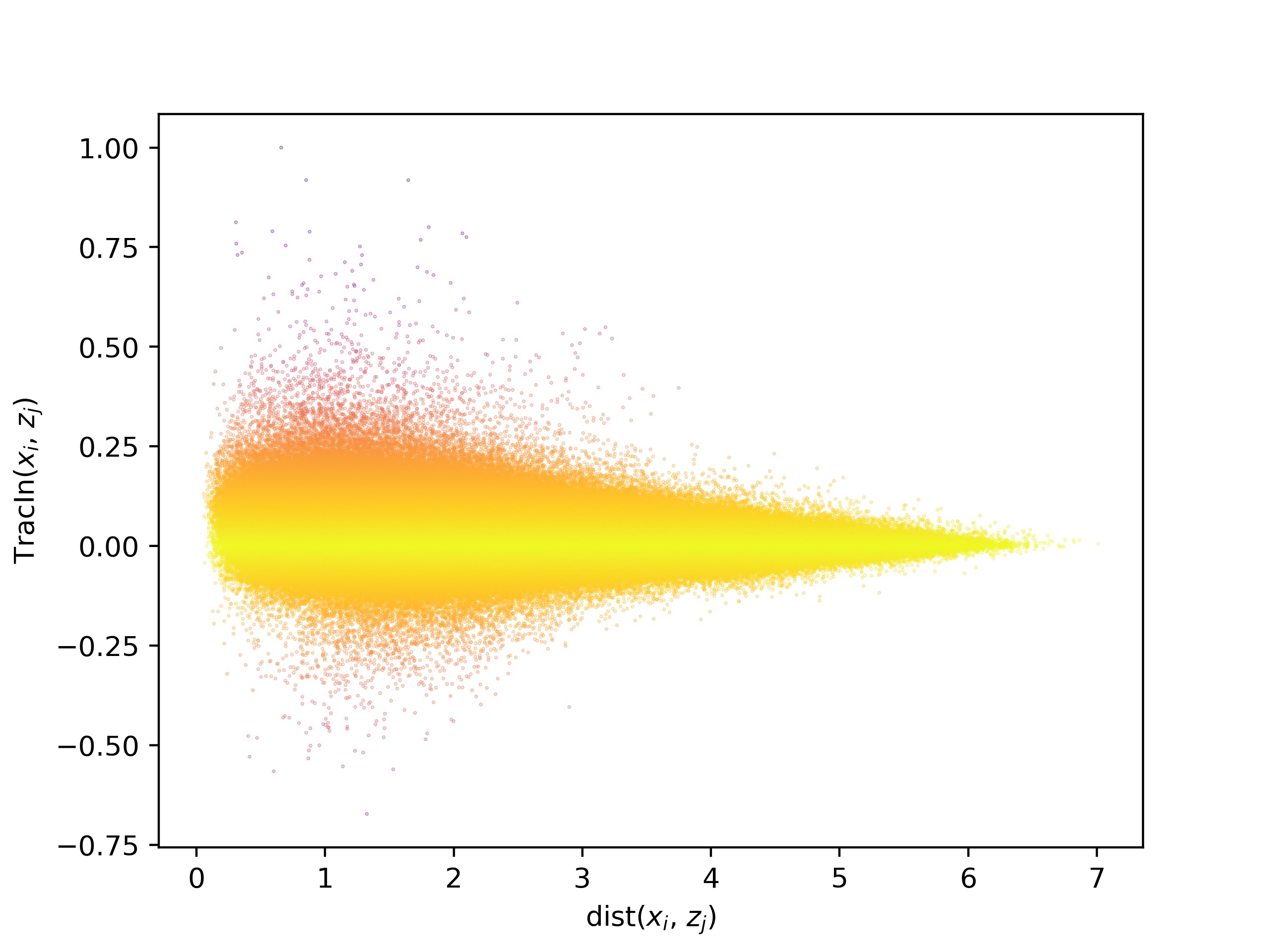}
    }
    \subfloat[][$\beta=4$, $\dlatent=128$]{ 
    	\includegraphics[trim=0 0 30 30, clip, width=0.25\textwidth]{traininf_bydist_mnist_beta4_z128.jpg}
	\label{fig: mnist test inf by latent dist}
    }
    \caption{Influences of training over test samples versus pairwise distances between them in the latent space. It is shown that both proponents and opponents are very close to test samples in the latent space especially when $\beta=4$.}
    \label{fig: mnist test-inf by dist appendix}
\end{figure}

We then compare the influences of training over test samples to the distances between them in the latent space in Figure \ref{fig: mnist test-inf by dist appendix}. We observe that both proponents and opponents are very close to test samples in the latent space, which indicates strong similarity between them. This phenomenon is more obvious when $\beta=4$.

\newpage

\begin{figure}[!h]
    \centering
    \subfloat[][$z$]{ 
    	\includegraphics[height=16cm]{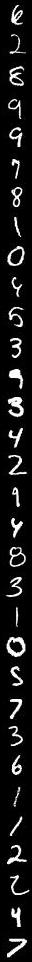}
	} \hspace{2em}
    \subfloat[][Strongest proponents]{ 
    	\includegraphics[height=16cm]{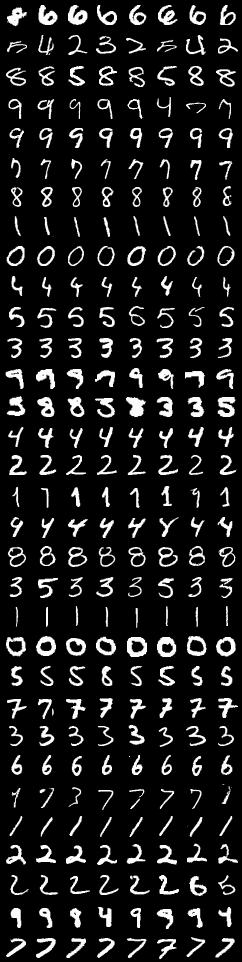}
	}
    \subfloat[][Strongest opponents]{ 
    	\includegraphics[height=16cm]{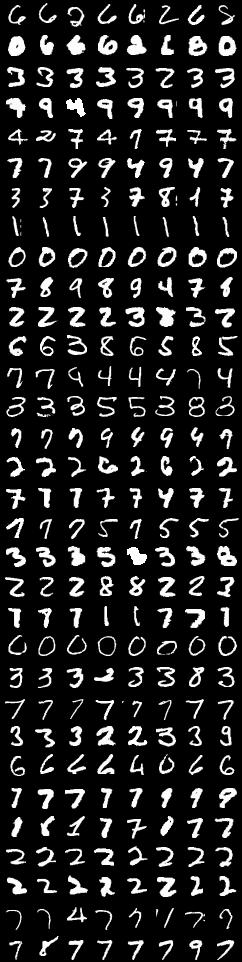}
	}
    \caption{Test samples $z$ from MNIST, their strongest proponents, and their strongest opponents. The strongest proponents look very similar to test samples, while the strongest opponents are often the same digit but are visually very different.}
    \label{fig: mnist test inf visualization appendix}
\end{figure}

We display the first 32 test samples in MNIST, their strongest proponents, and their strongest opponents in Figure \ref{fig: mnist test inf visualization appendix}. The strongest proponents look very similar to test samples. The strongest opponents are often the same digit but are visually very different. For instance, many strong opponents have very different thickness, shapes, or styles.

~~\newpage
\subsection{Influences over Test Data (CIFAR)}\label{appendix: test inf cifar}

\begin{figure}[!h]
    \centering
    \subfloat[][$\dlatent=64$]{ 
    	\includegraphics[trim=10 0 30 30, clip, width=0.3\textwidth]{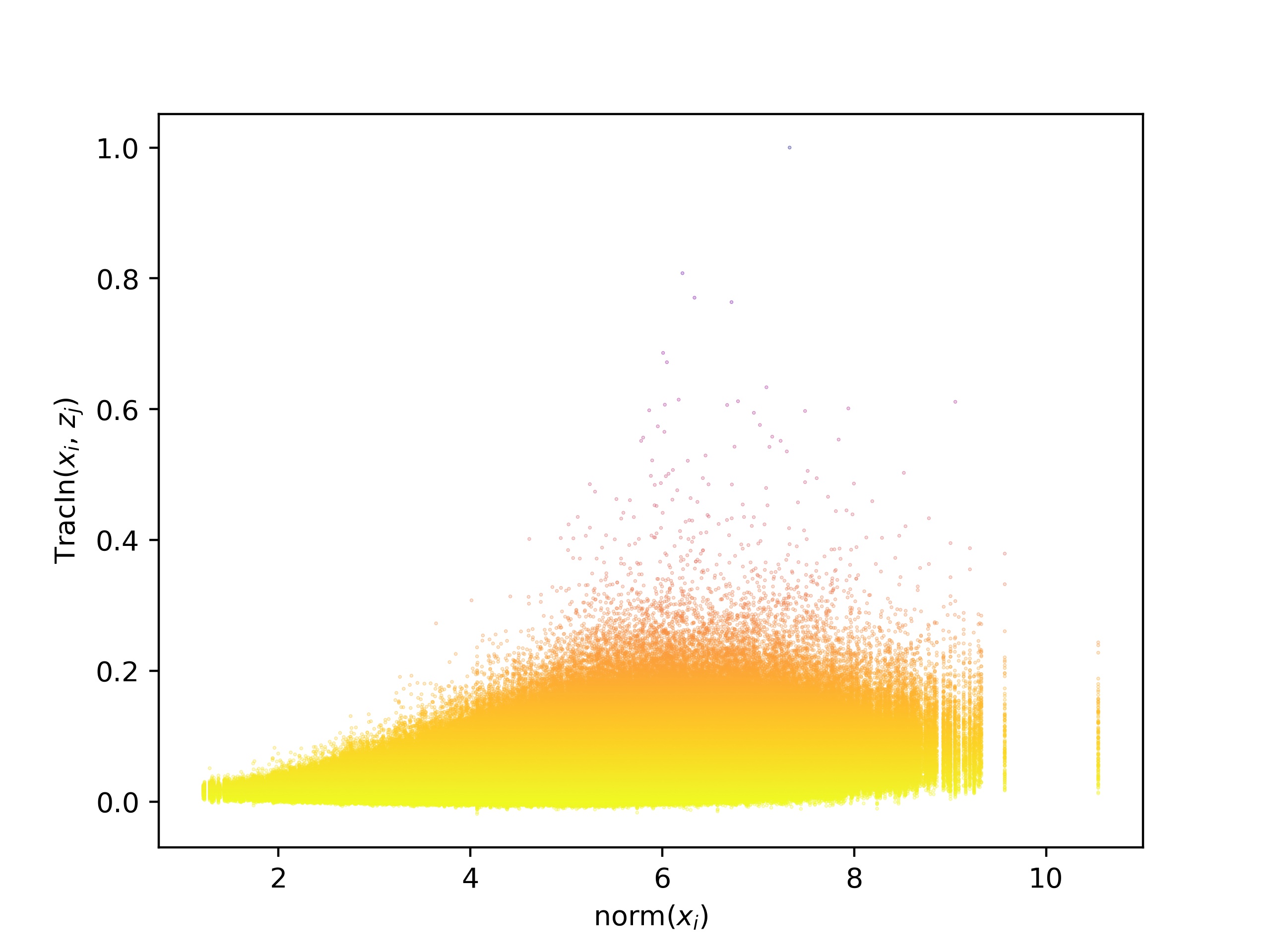}
    }
    \subfloat[][$\dlatent=128$]{ 
    	\includegraphics[trim=10 0 30 30, clip, width=0.3\textwidth]{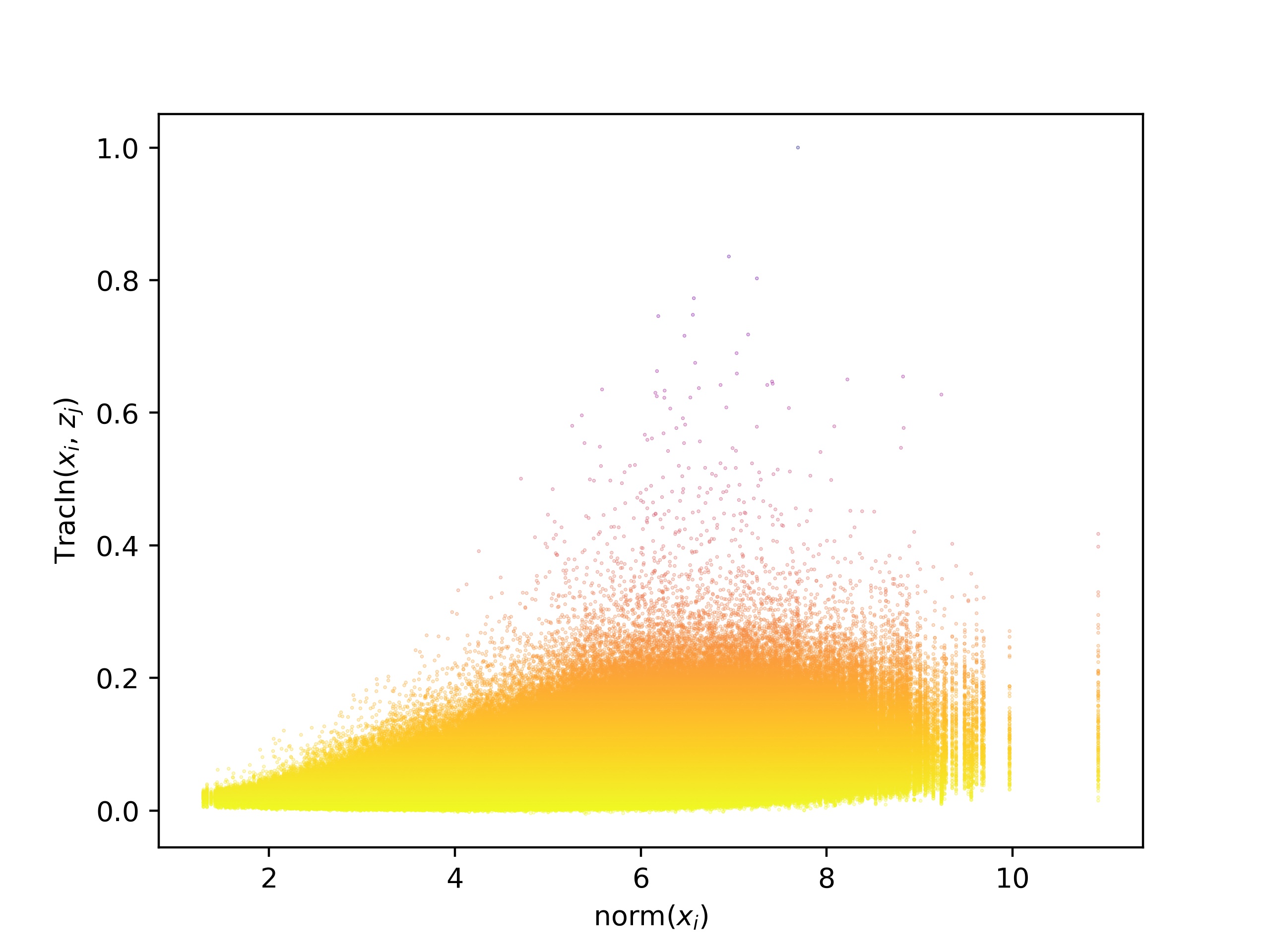}
    }
    \caption{Influences of training samples over test samples (CIFAR) versus norms of training samples in the latent space. It is shown that strong proponents have large norms.}
    \label{fig: cifar test-inf by latent norm appendix}
\end{figure}

\begin{figure}[!h]
    \centering
    \subfloat[][CIFAR$_0$]{ 
    	\includegraphics[trim=10 0 30 30, clip, width=0.19\textwidth]{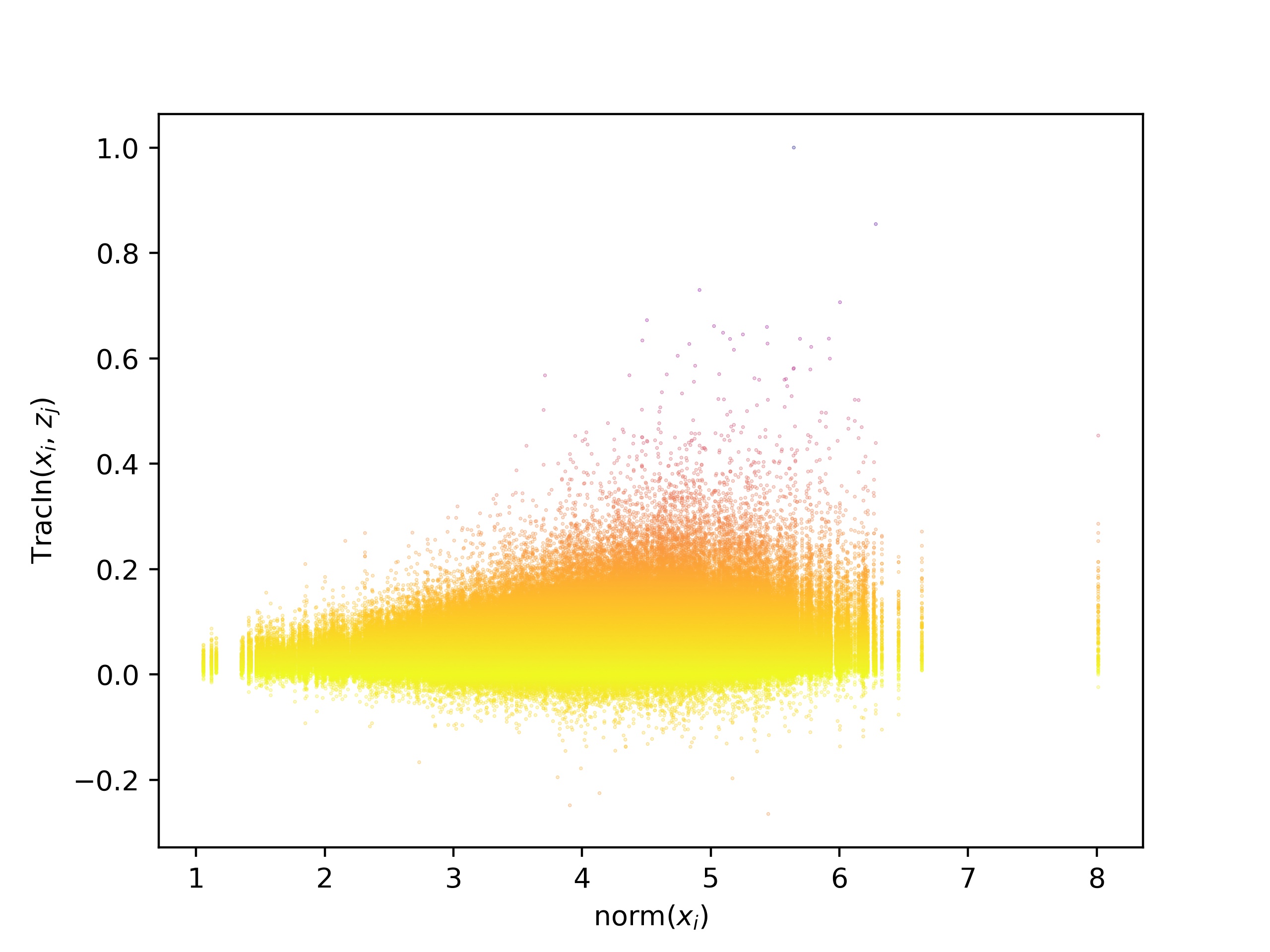}
    }
    \subfloat[][CIFAR$_1$]{ 
    	\includegraphics[trim=10 0 30 30, clip, width=0.19\textwidth]{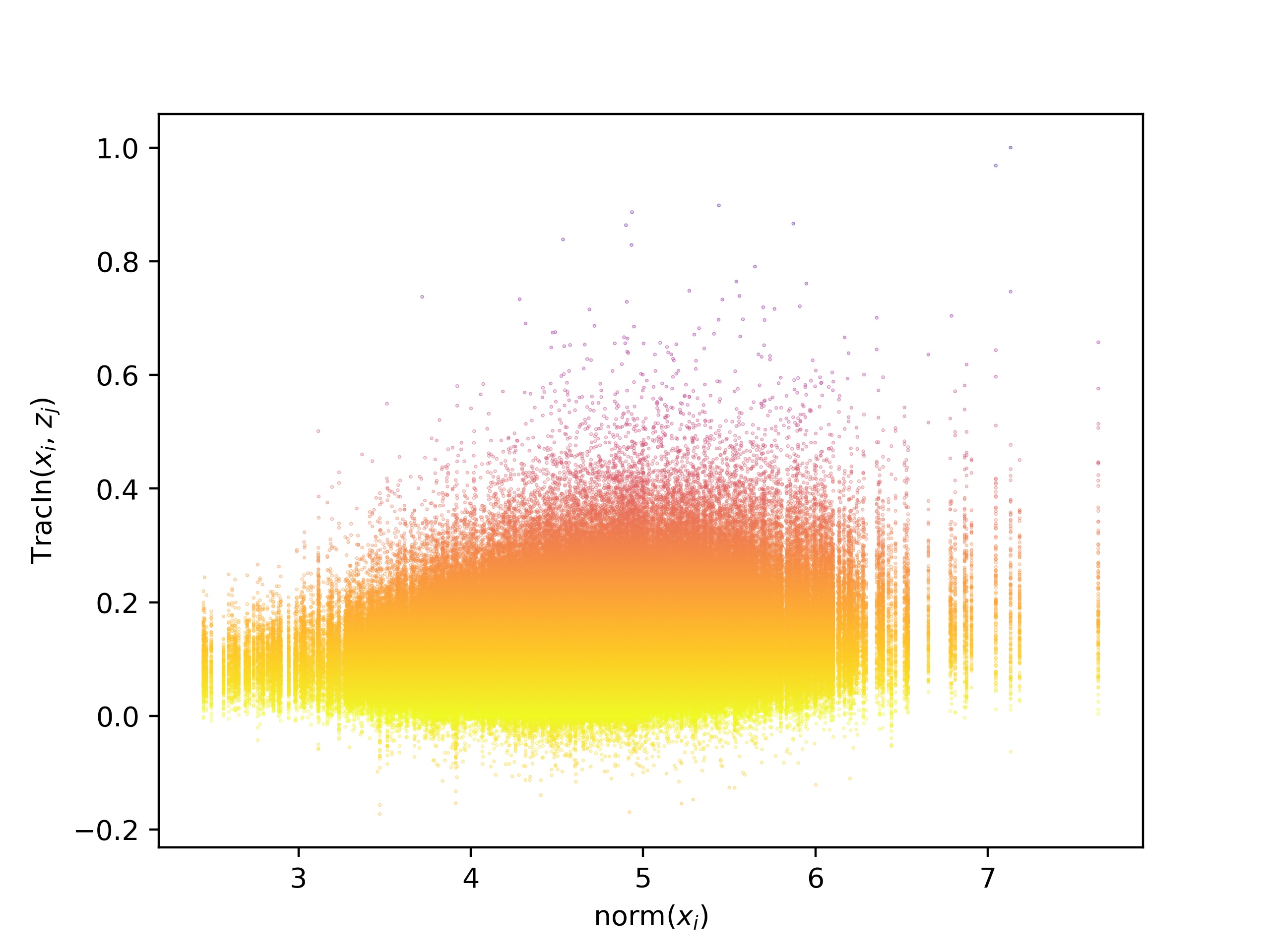}
    }
    \subfloat[][CIFAR$_2$]{ 
    	\includegraphics[trim=10 0 30 30, clip, width=0.19\textwidth]{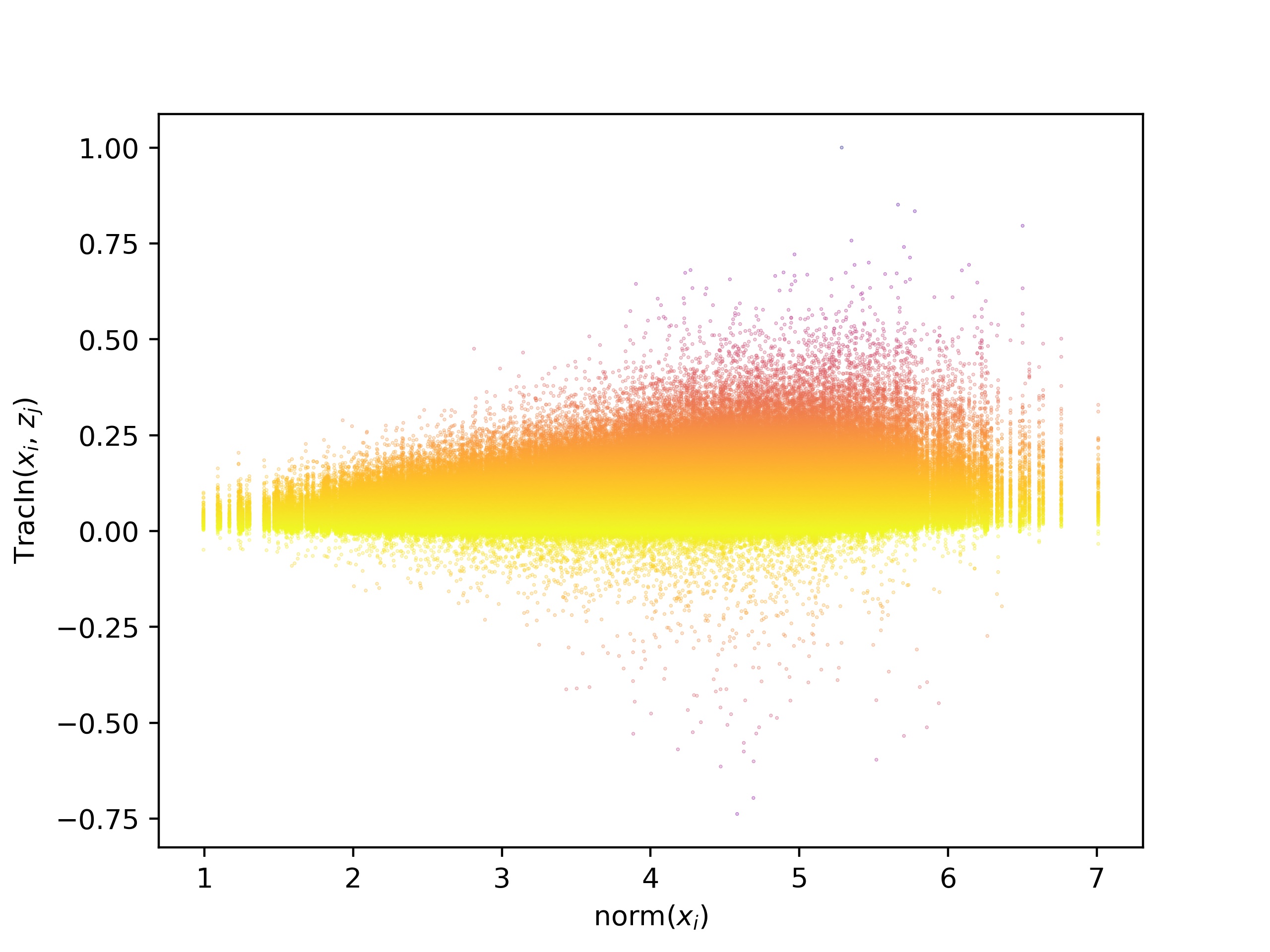}
    }
    \subfloat[][CIFAR$_3$]{ 
    	\includegraphics[trim=10 0 30 30, clip, width=0.19\textwidth]{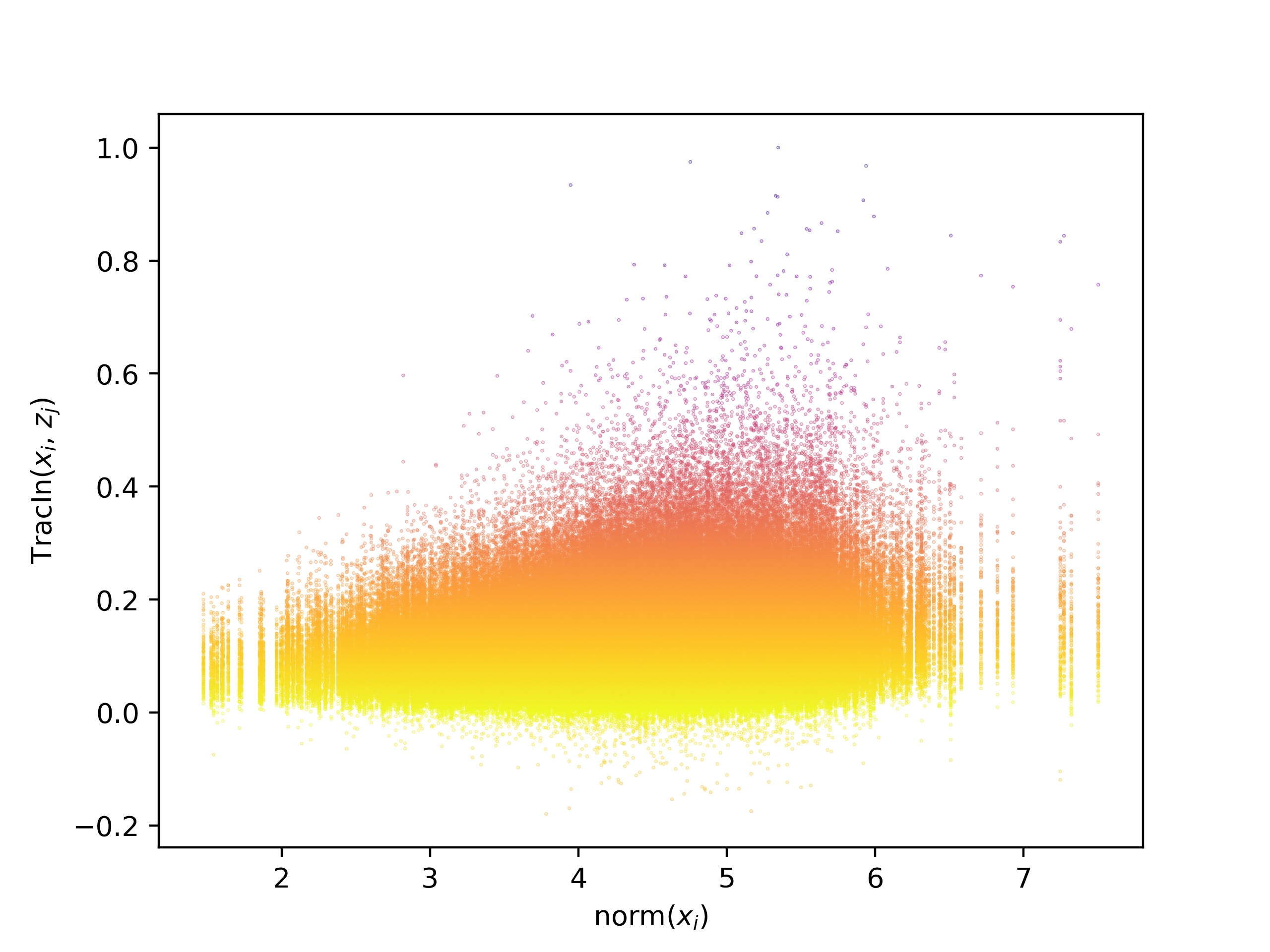}
    }
    \subfloat[][CIFAR$_4$]{ 
    	\includegraphics[trim=10 0 30 30, clip, width=0.19\textwidth]{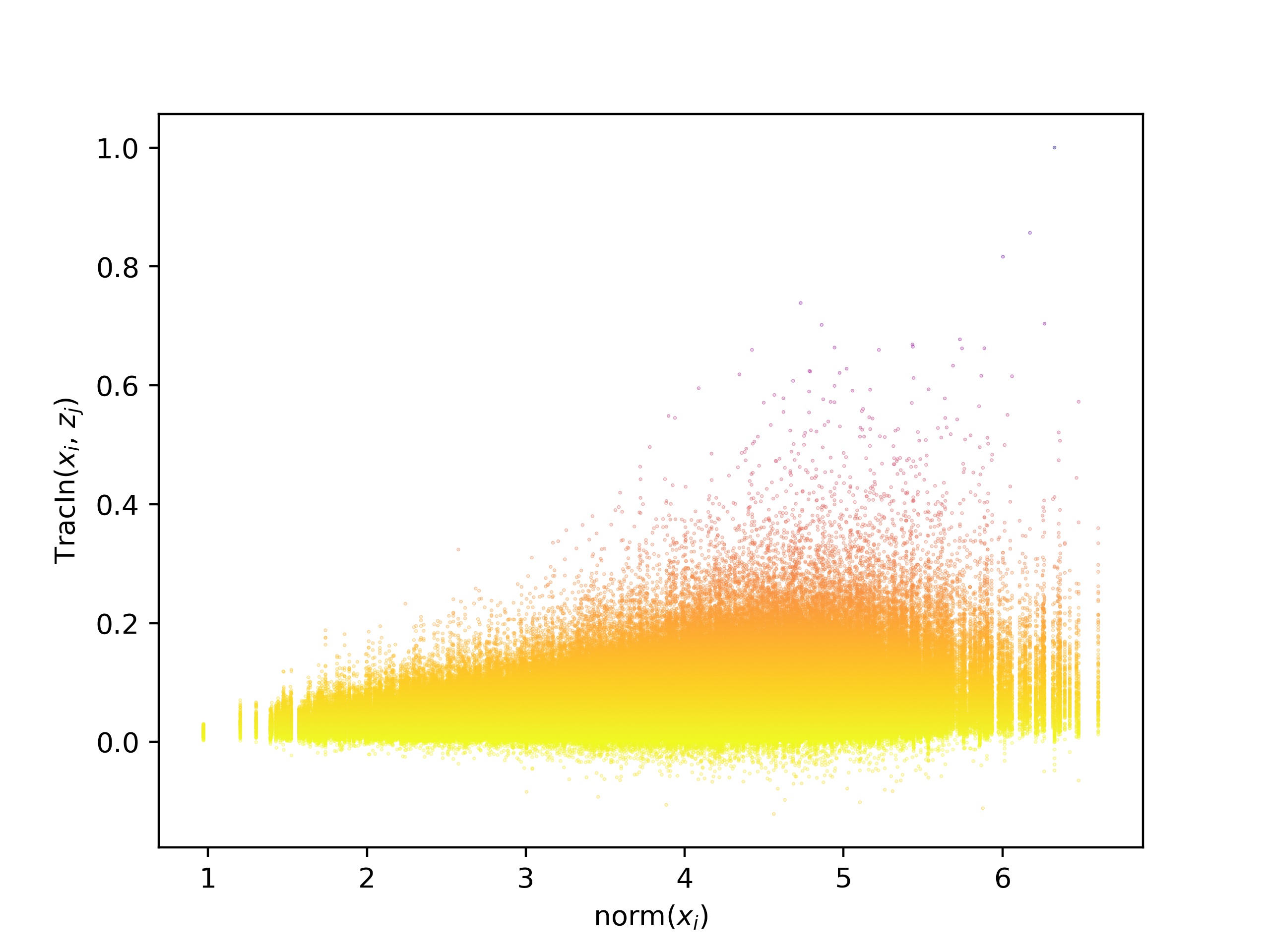}
    }\\
    \subfloat[][CIFAR$_5$]{ 
    	\includegraphics[trim=10 0 30 30, clip, width=0.19\textwidth]{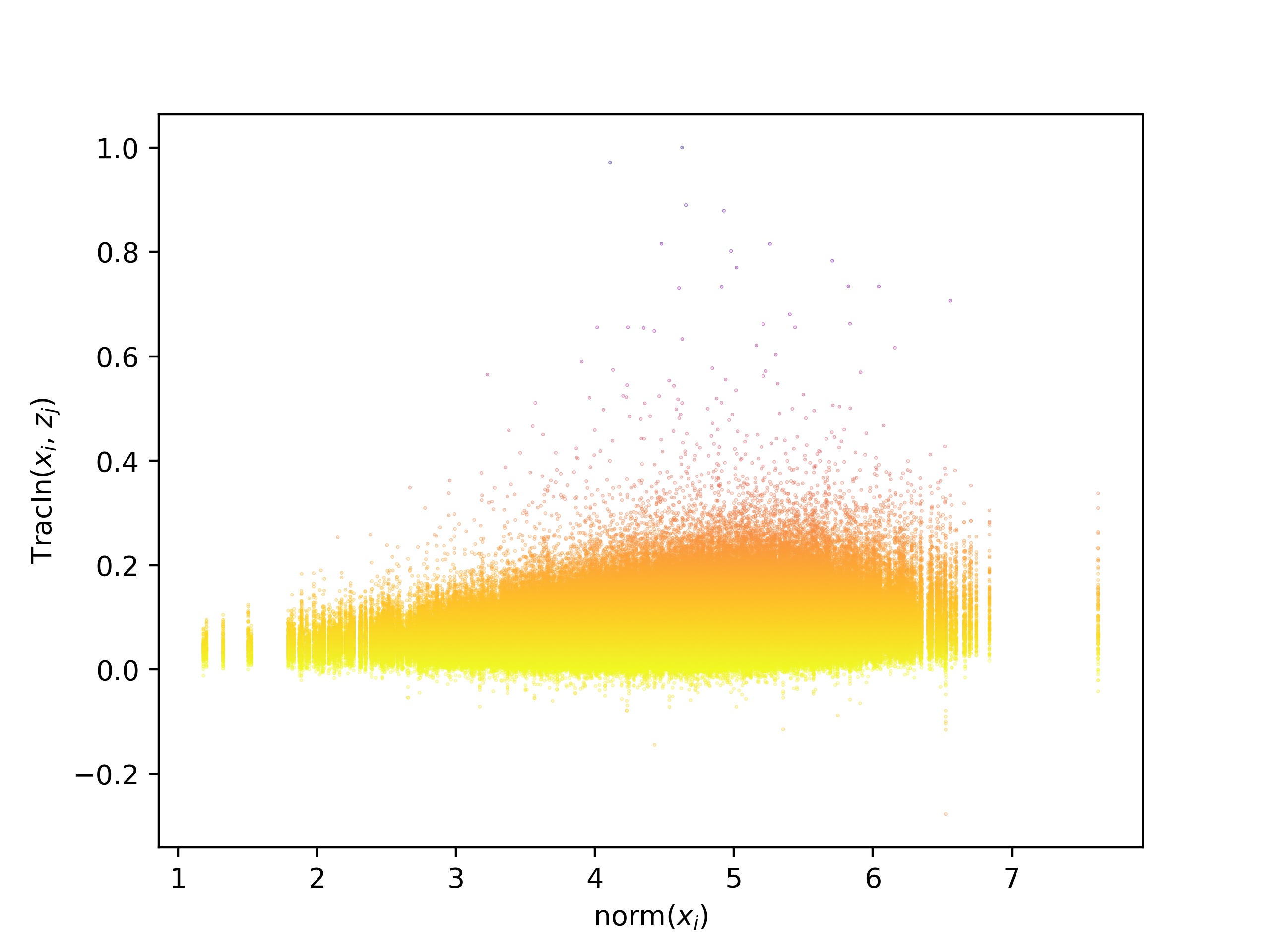}
    }
    \subfloat[][CIFAR$_6$]{ 
    	\includegraphics[trim=10 0 30 30, clip, width=0.19\textwidth]{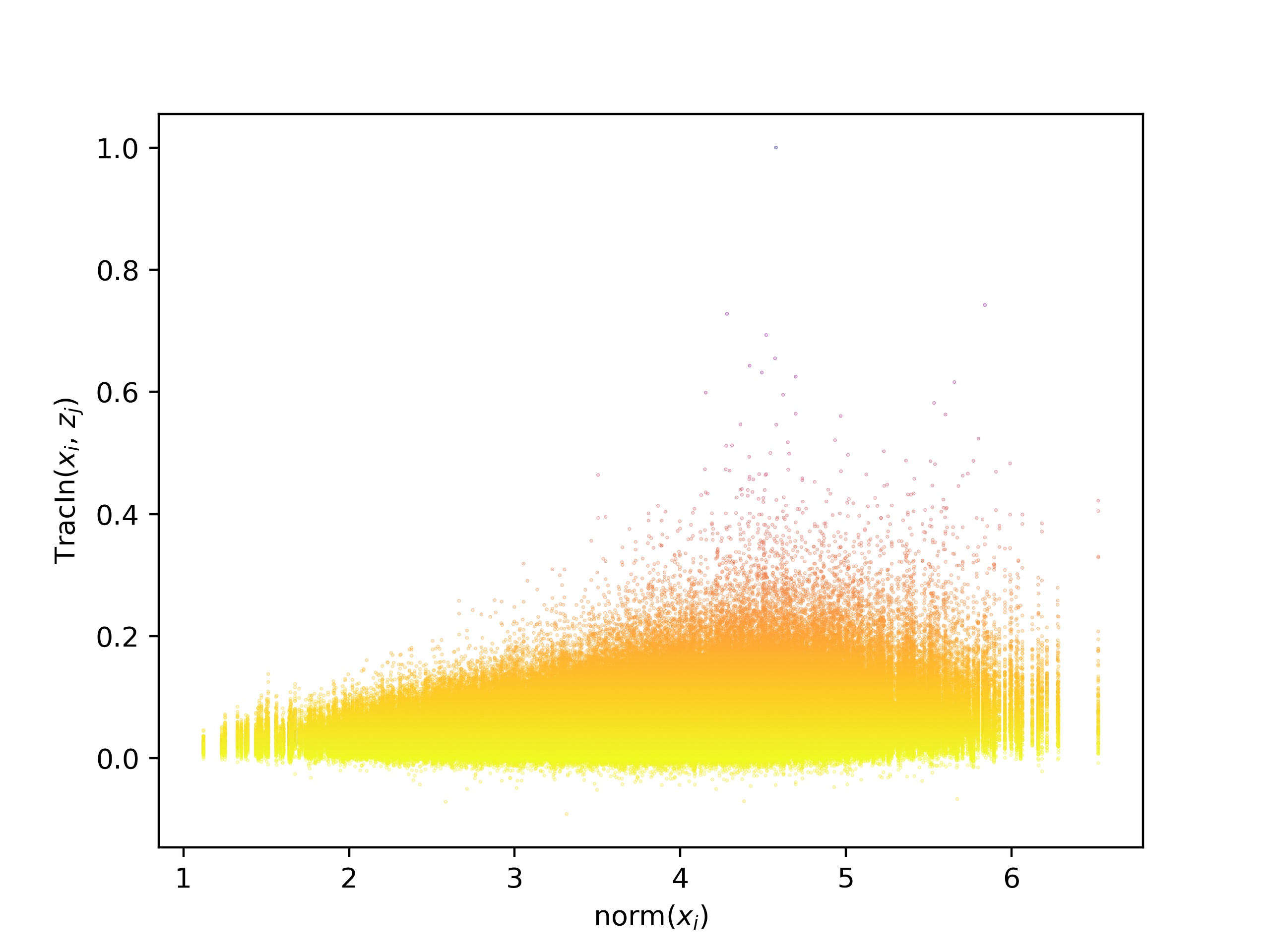}
    }
    \subfloat[][CIFAR$_7$]{ 
    	\includegraphics[trim=10 0 30 30, clip, width=0.19\textwidth]{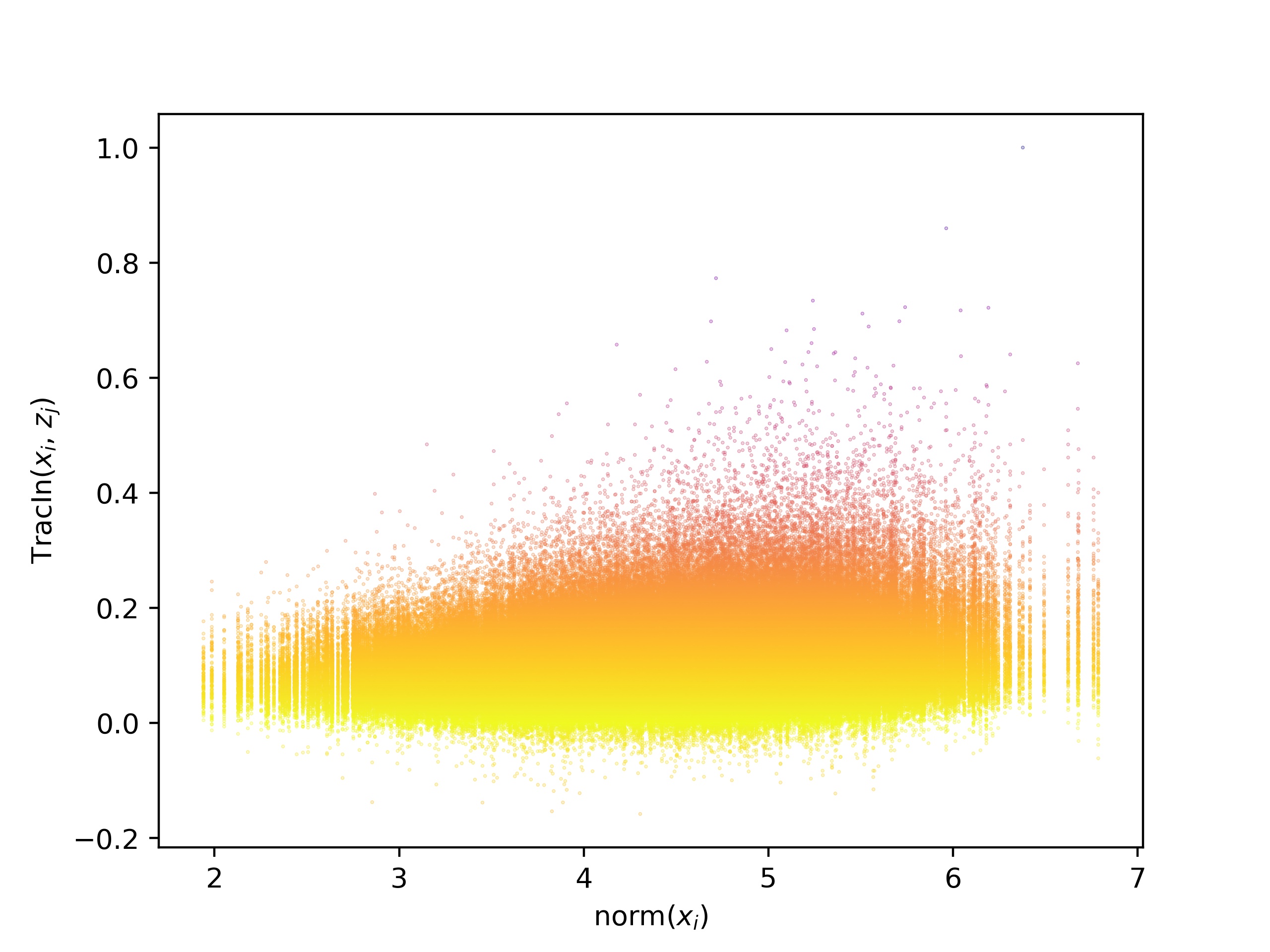}
    }
    \subfloat[][CIFAR$_8$]{ 
    	\includegraphics[trim=10 0 30 30, clip, width=0.19\textwidth]{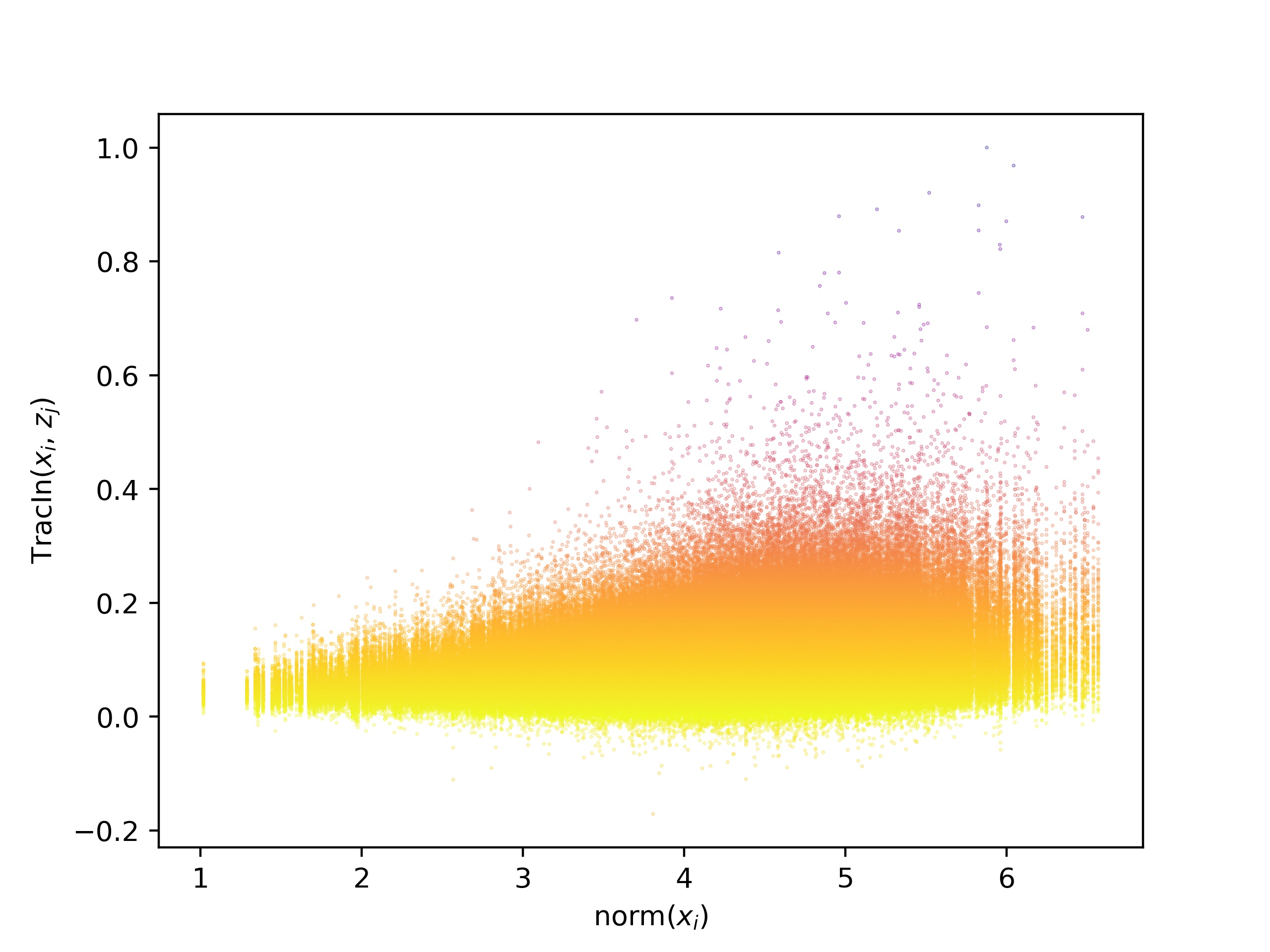}
    }
    \subfloat[][CIFAR$_9$]{ 
    	\includegraphics[trim=10 0 30 30, clip, width=0.19\textwidth]{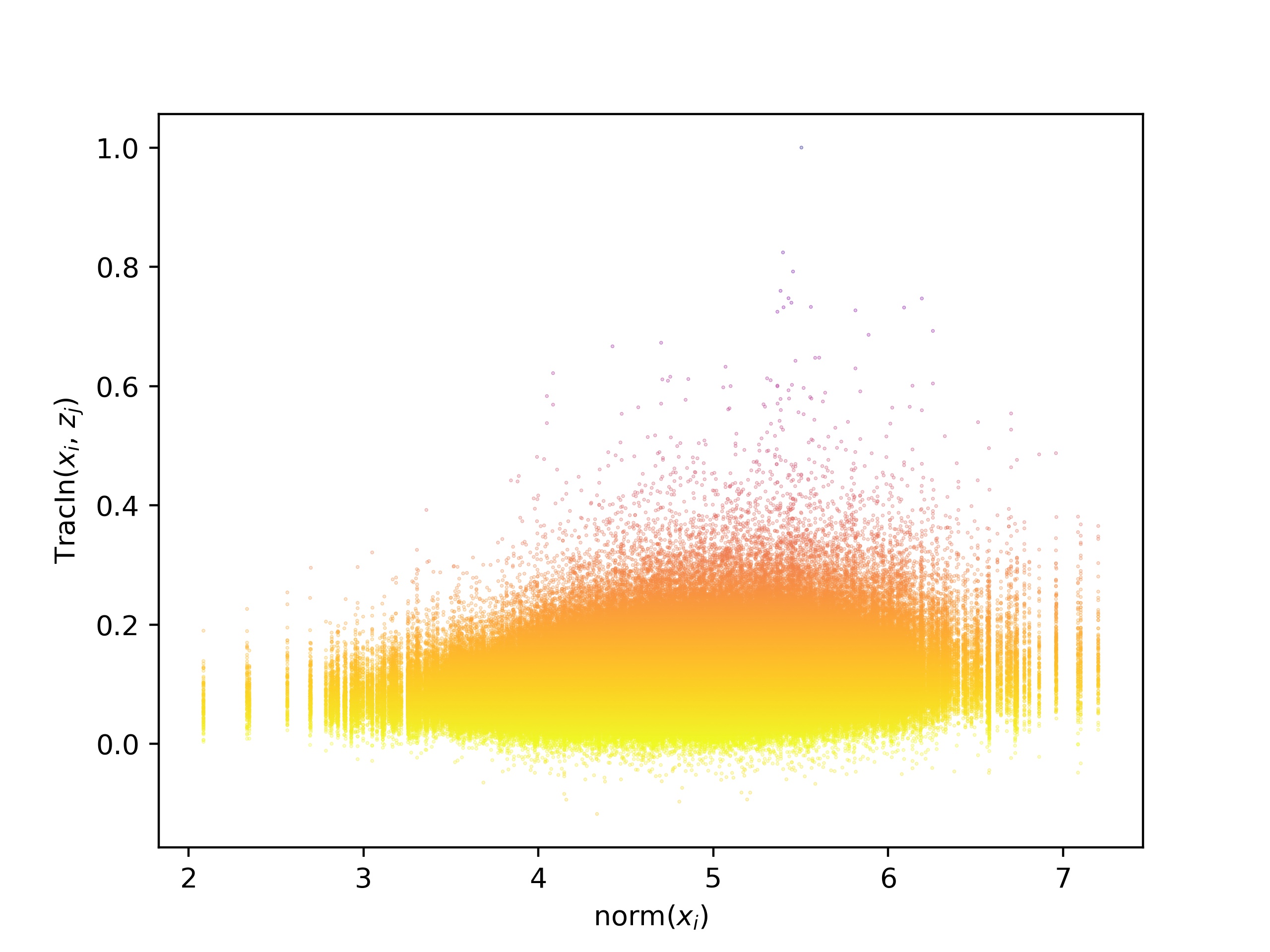}
    }
    \caption{Influences of training samples over test samples (CIFAR subclass) versus norms of training samples in the latent space. It is shown that strong proponents have large norms.}
    \label{fig: cifar subset test-inf by latent norm appendix}
\end{figure}

We compare the influences of training over test samples to the norms of training samples in the latent space. Results for CIFAR are shown in Figure \ref{fig: cifar test-inf by latent norm appendix} and results for CIFAR subclasses are shown in Figure \ref{fig: cifar subset test-inf by latent norm appendix}. We observe that strong proponents tend to have very large norms. This indicates they are high-contrast or very bright samples. This phenomenon occurs to CIFAR and all CIFAR subclasses. 

For 128 test samples in each CIFAR subclass, we report the statistics of the latent space norms of their strongest proponents, strongest opponents, and all training samples in Table \ref{tab: cifar subset test inf stat appendix}.

\begin{table}[!h]
        \caption{The means $\pm$ standard errors of latent space norms of training samples in CIFAR subclasses. Strong proponents tend to have large norms.}
        \vspace{0.5em}
        \label{tab: cifar subset test inf stat appendix}
	\centering
        \begin{tabular}{c|ccc}
           \hline
           Dataset & top-$0.1\%$ strong proponents & top-$0.1\%$ strong opponents & all training samples \\ \hline
           CIFAR$_0$ & $4.73\pm0.78$ & $4.26\pm0.91$ & $4.07\pm0.83$ \\
           CIFAR$_1$ & $5.30\pm0.71$ & $4.54\pm0.65$ & $4.65\pm0.64$ \\
           CIFAR$_2$ & $4.89\pm0.78$ & $4.18\pm0.88$ & $4.09\pm0.93$ \\
           CIFAR$_3$ & $5.09\pm0.75$ & $4.47\pm0.78$ & $4.42\pm0.79$ \\
           CIFAR$_4$ & $5.06\pm0.72$ & $3.96\pm1.00$ & $4.01\pm0.89$ \\
           CIFAR$_5$ & $5.25\pm0.75$ & $4.33\pm0.94$ & $4.54\pm0.81$ \\
           CIFAR$_6$ & $4.73\pm0.66$ & $3.99\pm0.80$ & $3.95\pm0.82$ \\
           CIFAR$_7$ & $5.11\pm0.76$ & $4.42\pm0.73$ & $4.43\pm0.74$ \\
           CIFAR$_8$ & $5.10\pm0.72$ & $4.19\pm0.88$ & $4.22\pm0.84$ \\
           CIFAR$_9$ & $5.48\pm0.62$ & $4.62\pm0.69$ & $4.79\pm0.64$ \\ \hline
        \end{tabular}
\end{table}

\begin{figure}[!h]
    \centering
    \subfloat[][$z$]{ 
    	\includegraphics[height=3cm]{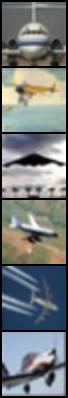}
	} \hspace{2em}
    \subfloat[][Strongest proponents]{ 
    	\includegraphics[height=3cm]{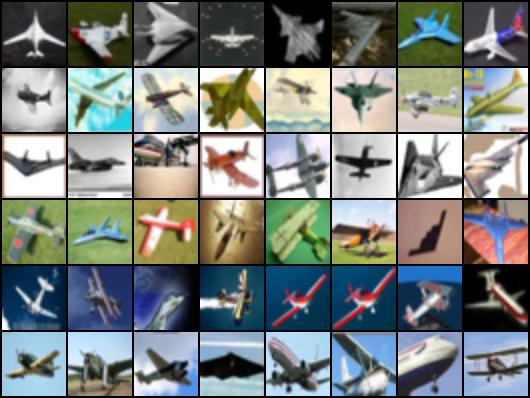}
	}
    \subfloat[][Strongest opponents]{ 
    	\includegraphics[height=3cm]{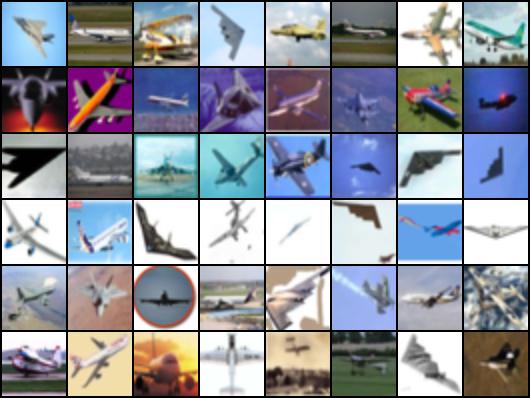}
	}
    \caption{Test samples $z$ from CIFAR$_0$, their strongest proponents, and their strongest opponents.}
    \label{fig: cifar 0 test inf visualization appendix}
\end{figure}

\begin{figure}[!h]
    \centering
    \subfloat[][$z$]{ 
    	\includegraphics[height=3cm]{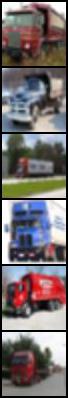}
	} \hspace{2em}
    \subfloat[][Strongest proponents]{ 
    	\includegraphics[height=3cm]{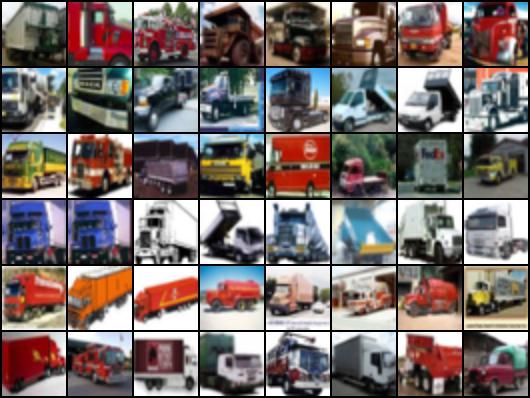}
	}
    \subfloat[][Strongest opponents]{ 
    	\includegraphics[height=3cm]{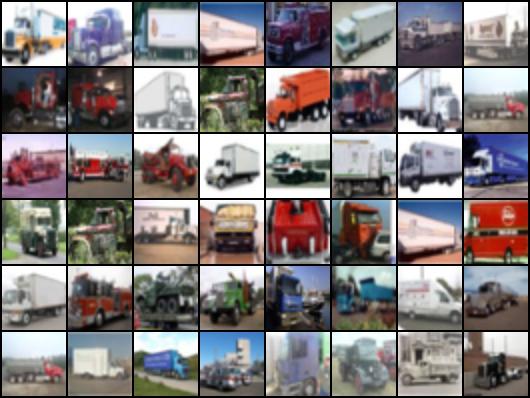}
	}
    \caption{Test samples $z$ from CIFAR$_9$, their strongest proponents, and their strongest opponents.}
    \label{fig: cifar 9 test inf visualization appendix}
\end{figure}

\newpage
For each CIFAR subclass, we display test samples, their strongest proponents, and their strongest opponents in Figures \ref{fig: cifar 0 test inf visualization appendix} $\sim$ \ref{fig: cifar 9 test inf visualization appendix}. The strongest proponents seem to match the color of test samples in terms of background and the object. In addition, they tend to have the same but brighter colors.

\begin{figure}[!h]
    \centering
    \subfloat[][$z$]{ 
    	\includegraphics[height=3cm]{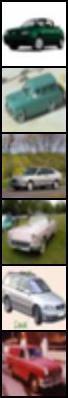}
	} \hspace{2em}
    \subfloat[][Strongest proponents]{ 
    	\includegraphics[height=3cm]{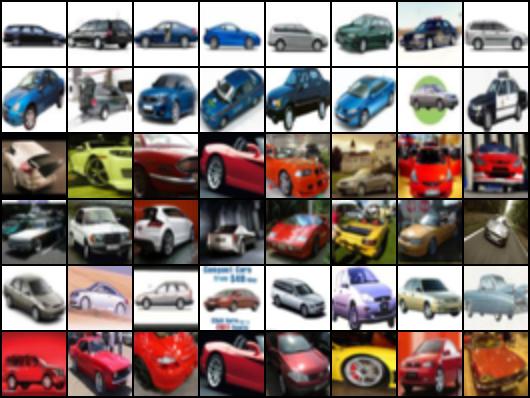}
	}
    \subfloat[][Strongest opponents]{ 
    	\includegraphics[height=3cm]{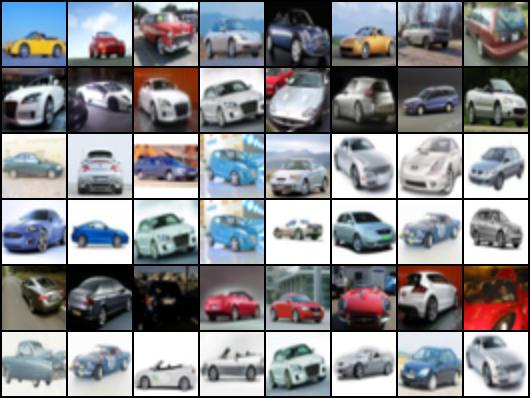}
	}
    \caption{Test samples $z$ from CIFAR$_1$, their strongest proponents, and their strongest opponents.}
    \label{fig: cifar 1 test inf visualization appendix}
\end{figure}

\begin{figure}[!h]
    \centering
    \subfloat[][$z$]{ 
    	\includegraphics[height=3cm]{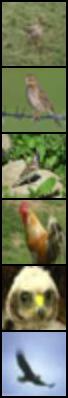}
	} \hspace{2em}
    \subfloat[][Strongest proponents]{ 
    	\includegraphics[height=3cm]{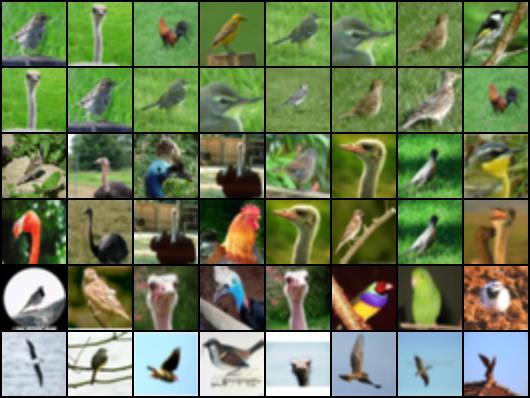}
	}
    \subfloat[][Strongest opponents]{ 
    	\includegraphics[height=3cm]{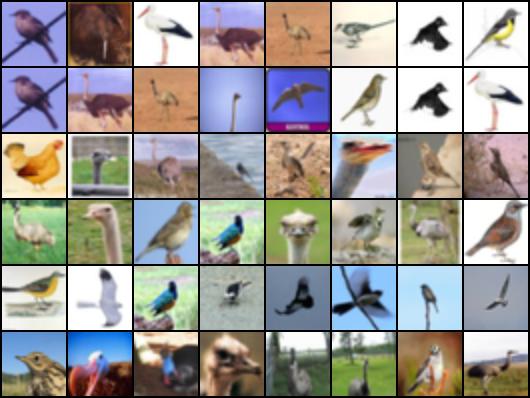}
	}
    \caption{Test samples $z$ from CIFAR$_2$, their strongest proponents, and their strongest opponents.}
    \label{fig: cifar 2 test inf visualization appendix}
\end{figure}

\begin{figure}[!h]
    \centering
    \subfloat[][$z$]{ 
    	\includegraphics[height=3cm]{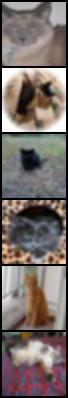}
	} \hspace{2em}
    \subfloat[][Strongest proponents]{ 
    	\includegraphics[height=3cm]{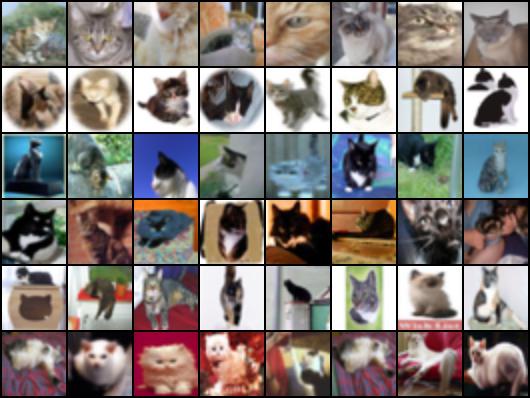}
	}
    \subfloat[][Strongest opponents]{ 
    	\includegraphics[height=3cm]{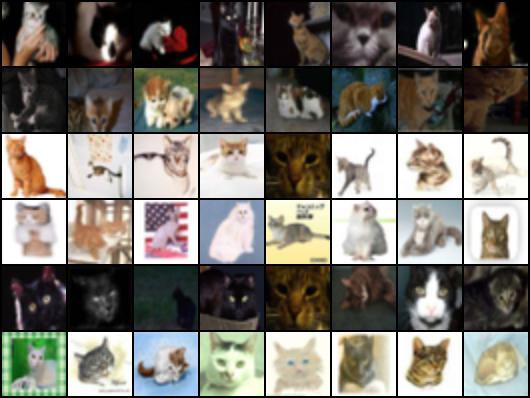}
	}
    \caption{Test samples $z$ from CIFAR$_3$, their strongest proponents, and their strongest opponents.}
    \label{fig: cifar 3 test inf visualization appendix}
\end{figure}

\begin{figure}[!h]
    \centering
    \subfloat[][$z$]{ 
    	\includegraphics[height=3cm]{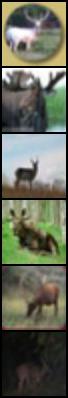}
	} \hspace{2em}
    \subfloat[][Strongest proponents]{ 
    	\includegraphics[height=3cm]{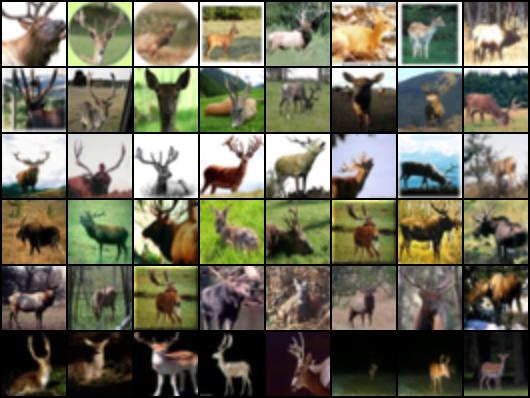}
	}
    \subfloat[][Strongest opponents]{ 
    	\includegraphics[height=3cm]{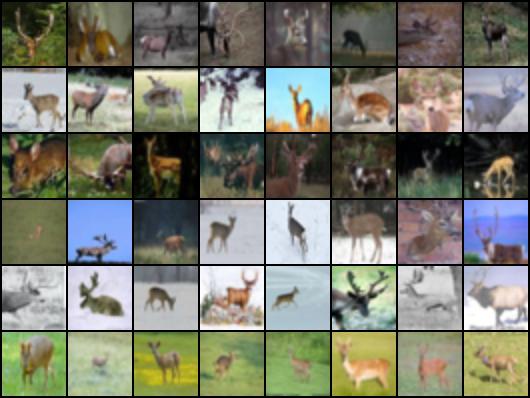}
	}
    \caption{Test samples $z$ from CIFAR$_4$, their strongest proponents, and their strongest opponents.}
    \label{fig: cifar 4 test inf visualization appendix}
\end{figure}

\begin{figure}[!h]
    \centering
    \subfloat[][$z$]{ 
    	\includegraphics[height=3cm]{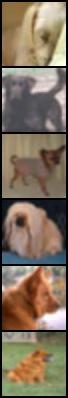}
	} \hspace{2em}
    \subfloat[][Strongest proponents]{ 
    	\includegraphics[height=3cm]{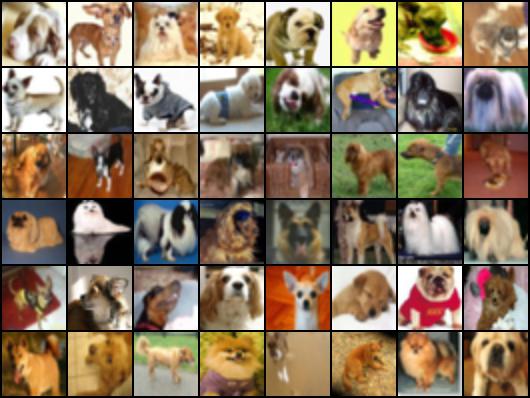}
	}
    \subfloat[][Strongest opponents]{ 
    	\includegraphics[height=3cm]{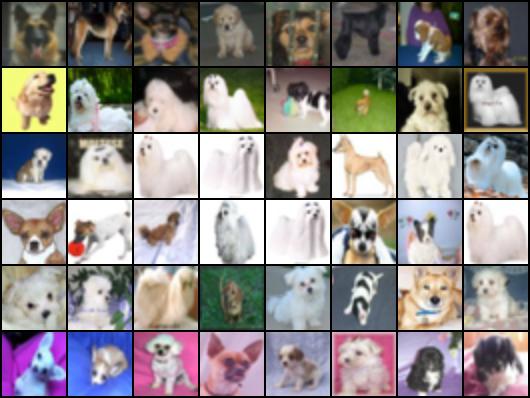}
	}
    \caption{Test samples $z$ from CIFAR$_5$, their strongest proponents, and their strongest opponents.}
    \label{fig: cifar 5 test inf visualization appendix}
\end{figure}

\begin{figure}[!h]
    \centering
    \subfloat[][$z$]{ 
    	\includegraphics[height=3cm]{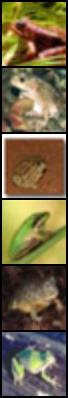}
	} \hspace{2em}
    \subfloat[][Strongest proponents]{ 
    	\includegraphics[height=3cm]{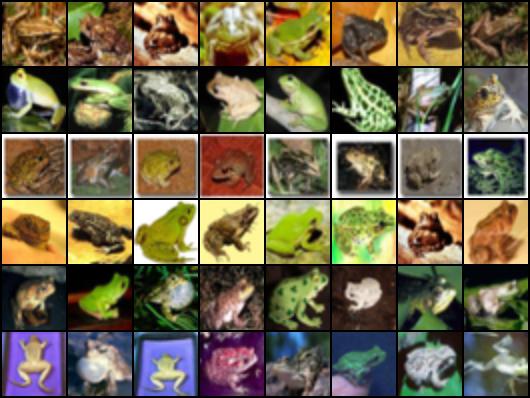}
	}
    \subfloat[][Strongest opponents]{ 
    	\includegraphics[height=3cm]{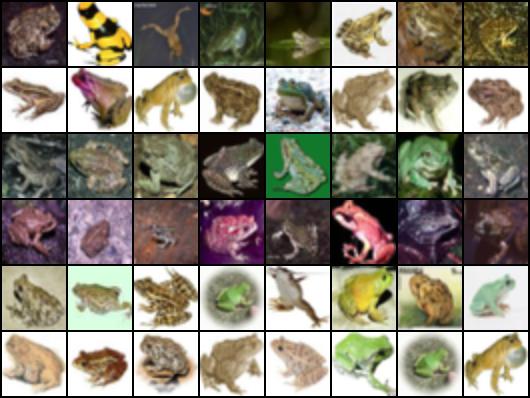}
	}
    \caption{Test samples $z$ from CIFAR$_6$, their strongest proponents, and their strongest opponents.}
    \label{fig: cifar 6 test inf visualization appendix}
\end{figure}

\begin{figure}[!h]
    \centering
    \subfloat[][$z$]{ 
    	\includegraphics[height=3cm]{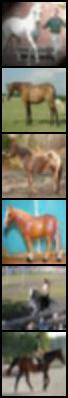}
	} \hspace{2em}
    \subfloat[][Strongest proponents]{ 
    	\includegraphics[height=3cm]{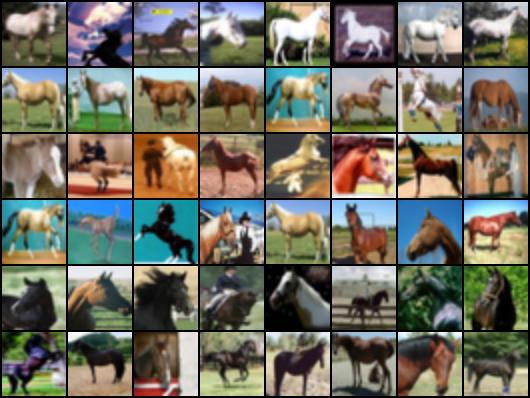}
	}
    \subfloat[][Strongest opponents]{ 
    	\includegraphics[height=3cm]{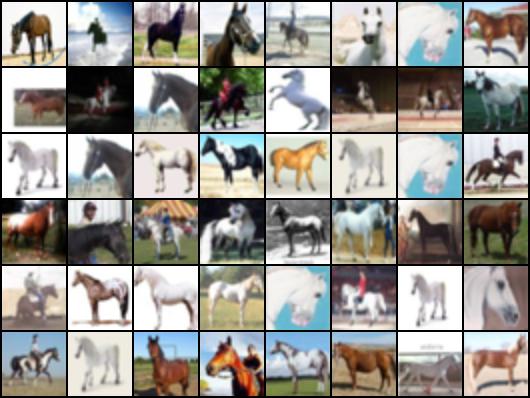}
	}
    \caption{Test samples $z$ from CIFAR$_7$, their strongest proponents, and their strongest opponents.}
    \label{fig: cifar 7 test inf visualization appendix}
\end{figure}

\begin{figure}[!h]
    \centering
    \subfloat[][$z$]{ 
    	\includegraphics[height=3cm]{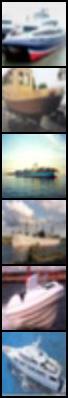}
	} \hspace{2em}
    \subfloat[][Strongest proponents]{ 
    	\includegraphics[height=3cm]{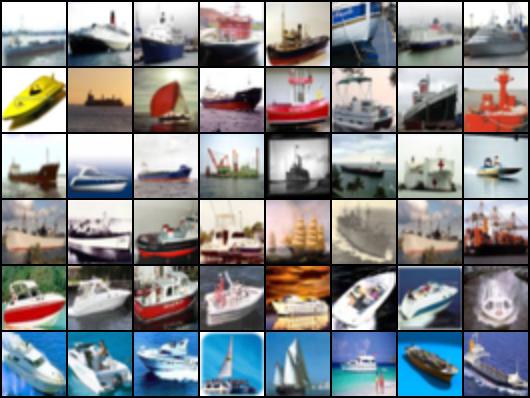}
	}
    \subfloat[][Strongest opponents]{ 
    	\includegraphics[height=3cm]{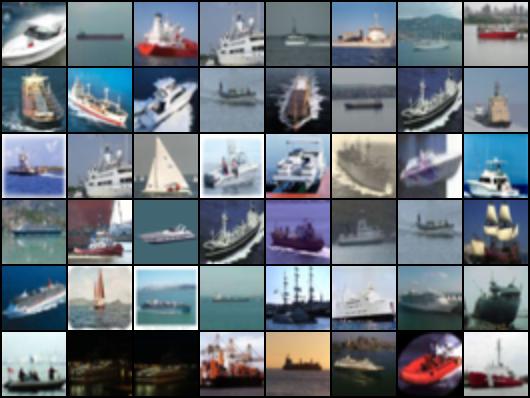}
	}
    \caption{Test samples $z$ from CIFAR$_8$, their strongest proponents, and their strongest opponents.}
    \label{fig: cifar 8 test inf visualization appendix}
\end{figure}

\end{document}